\documentclass[lettersize,journal]{IEEEtran}
\usepackage{amsmath,amsfonts,amssymb}
\usepackage{algorithmic}
\usepackage{algorithm}
\usepackage{array}
\newcolumntype{P}[1]{>{\centering\arraybackslash}p{#1}}
\usepackage[caption=false,font=normalsize,labelfont=sf,textfont=sf]{subfig}
\usepackage{textcomp}
\usepackage{stfloats}
\usepackage{url}
\usepackage{verbatim}
\usepackage{graphicx}
\usepackage{cite}
\def\BibTeX{{\rm B\kern-.05em{\sc i\kern-.025em b}\kern-.08em
    T\kern-.1667em\lower.7ex\hbox{E}\kern-.125emX}}
\usepackage{booktabs}
\usepackage{multirow}
\usepackage{siunitx}
\usepackage[table]{xcolor}
\usepackage[table]{xcolor}
\usepackage{color,soul}
\sethlcolor{yellow}
    
\usepackage[pdfstartview=XYZ,
bookmarks=true,
colorlinks=true,
linkcolor=black,
urlcolor=black,
citecolor=black,
bookmarks=true,
linktocpage=true, 
hyperindex=true
]{hyperref}

\usepackage{orcidlink}

\newtheorem{defka}{Definition}
\newtheorem{thm}{Theorem}
\newtheorem{assumption}{Assumption}
\newtheorem{corollary}{Corollary}
\newtheorem{lem}{Lemma}
\newtheorem{rmk}{Remark}
\newtheorem{prop}{Property}
\newtheorem{pf}{Proof}
\renewcommand{\figurename}{Fig.} 

\renewcommand{\arraystretch}{1.2}
\begin{document}

\title{Orchestrated Robust Controller for Precision Control of Heavy-duty Hydraulic Manipulators}

\author{Mahdi Hejrati \orcidlink{0000-0002-8017-4355}, Jouni Mattila \orcidlink{0000-0003-1799-4323}
\thanks{This work is supported by Business Finland partnership project "Future all-electric rough terrain autonomous mobile manipulators" (Grant 2334/31/222).}
\thanks{M. Hejrati, corresponding author,  is with the Department of Engineering and Natural Science, Tampere University, 7320 Tampere, Finland (e-mail: mahdi.hejrati@tuni.fi).}
\thanks{J. Mattila is with  the Department of Engineering and Natural Science, Tampere University, 7320 Tampere, Finland (e-mail: jouni.mattila@tuni.fi).}}

\markboth{IEEE Transactions,~Vol.~, No.~, December~2023}%
{Shell \MakeLowercase{\textit{et al.}}: A Sample Article Using IEEEtran.cls for IEEE Journals}

© 2025 IEEE. Personal use of this material is permitted.
Permission from IEEE must be obtained for all other uses,
including reprinting/republishing this material for advertising
or promotional purposes, collecting new collected works
for resale or redistribution to servers or lists, or reuse of
any copyrighted component of this work in other works.
This work has been submitted to the IEEE for possible
publication. Copyright may be transferred without notice,
after which this version may no longer be accessible.

\maketitle

\begin{abstract}
	Vast industrial investment along with increased academic research on heavy-duty hydraulic manipulators has unavoidably paved the way for their automatization, necessitating the design of robust and high-precision controllers. In this study, an orchestrated robust controller is designed to address the mentioned issue for generic manipulators with an anthropomorphic arm and spherical wrist. Thanks to virtual decomposition control (VDC), the entire robotic system is decomposed into subsystems, and a robust controller is designed at each local subsystem by considering unknown model uncertainties, unknown disturbances, and compound input nonlinearities. As such, radial basic function neural networks (RBFNNs) are incorporated into VDC to tackle unknown disturbances and uncertainties, resulting in novel decentralized RBFNNs. All robust local controllers designed at each local subsystem, then, are orchestrated to accomplish high-precision control. In the end, for the first time in the context of VDC, a semi-globally uniformly ultimate boundedness is achieved under the designed controller. The validity of the theoretical results is verified by performing extensive simulations and experiments on a 6-degrees-of-freedom industrial manipulator with a nominal lifting capacity of $600\, kg$ at $5$ meters reach. Comparing the simulation result to the state-of-the-art controller along with provided experimental results, demonstrates that proposed method established all promises and performed excellently.
 
\end{abstract}

\begin{NtP}
    Heavy-duty hydraulic manipulators (HHMs) play a pivotal role in various industrial applications, where their automation is critical for enhancing productivity, precision, and operational efficiency. However, automating HHMs presents significant challenges due to their complex dynamics, highly nonlinear behavior, and presence of unknown real-world uncertainties, such as unknown disturbances, non-ideal devices, unmodeled dynamics, and limited computational resources. These challenges underscore the need for control algorithms that are robust to such uncertainties, computationally efficient for real-time deployment, and capable of achieving high accuracy for reliable execution of free-motion tasks. This study designs a control scheme for generic 6-degree-of-freedom (DoF) HHMs, which ensures robustness and accuracy. Theoretical analysis along with experimental validations on a 6-DoF industrial HHM perfectly display stability of the method, robustness to real-world unknown uncertainties, efficiency in real-time implementation, and excellent accuracy in task execution. These results highlight scheme's reliability and universality for industrial HHM applications, making it a significant contribution to advancing automation in this field. Additionally, detailed performance index provided with results enhances replicability and measurability, paving the way for meaningful comparisons in the future.
\end{NtP}

\begin{IEEEkeywords}
Hydraulic manipulators,  neuro-adaptive control, input nonlinearities, nonlinear model-based control
\end{IEEEkeywords}
\newpage
\section*{List of Abbreviations}

\begin{table}[h!]
    \centering
    \small
    \begin{tabular}{p{0.2\textwidth}p{0.25\textwidth}}
        \textbf{HHM} &  Heavy-duty hydraulic manipulator \tabularnewline
        \textbf{PWR} &  Power-to-weight ratio \tabularnewline
        \textbf{DoF} &  Degree of freedom \tabularnewline
        \textbf{VD}C &  Virtual decomposition control \tabularnewline
        \textbf{ORC} &  Orchestrated robust controller \tabularnewline
        \textbf{DRBFNN} &  Decentralized radial basis function neural network \tabularnewline
        \textbf{UUB} &  Uniformely ultimate boundedness \tabularnewline
        \textbf{SGUUB} &  semi-global UUB \tabularnewline
        \textbf{VPF} &  Virtual power flow \tabularnewline
        \textbf{VCP} &  Virtual cutting point \tabularnewline
        \textbf{RHA} &  Rotary hydraulic actuator \tabularnewline
        \textbf{NMBC} &  nonlinear model-based controller \tabularnewline
        \end{tabular}
\end{table}

\section*{List of Symbols}
\begin{table}[h!]
    \centering
    \small
    \begin{tabular}{p{0.2\textwidth}p{0.25\textwidth}}
        $\xi_i , \zeta _i \in \mathbb{R} $ &  Joint angles \tabularnewline
        \{A\}, \{B\} &  Body frame of a rigid body \tabularnewline
        $^AV \in \mathbb{R}^{6}$ &  Spatial velocity vector \tabularnewline
        $^Av \in \mathbb{R}^{3}$ & linear velocity vector \tabularnewline
        $^{A}\omega\in \mathbb{R}^3$ & angular velocity vector \tabularnewline
        $^AF \in \mathbb{R}^{6}$ &  Spatial force vector \tabularnewline
        $^{A}f\in \mathbb{R}^3$ & force vector \tabularnewline
        $^{A}\tau \in \mathbb{R}^3$ & moment vector \tabularnewline
        $^AF^* \in \mathbb{R}^{6}$ &  Net spatial force vector \tabularnewline
        $^AV_r \in \mathbb{R}^{6}$ &  Required spatial velocity vector \tabularnewline
        $^AF_r \in \mathbb{R}^{6}$ &  Required spatial force vector \tabularnewline
        $^AF^*_r \in \mathbb{R}^{6}$ &  Net required spatial force vector \tabularnewline
        $^AU_B \in \mathbb{R}^{6 \times 6} $ &  Transformation matrix \tabularnewline
        $^AR_B \in \mathbb{R}^{3 \times 3} $ &  Rotation matrix \tabularnewline
        $^{A}r_{{A}{B}} \in \mathbb{R}^3 $ &  Distance between \{A\} and \{B\} \tabularnewline
        $M_A \in \mathbb{R}^{6 \times 6}$ &  Mass matrix \tabularnewline
        $C_A \in \mathbb{R}^{6 \times 6}$ &  Coriolis and centrifugal matrix \tabularnewline
        $G_A \in \mathbb{R}^{6}$ &  Gravitational forces vector \tabularnewline
        $Y_A \in \mathbb{R}^{6 \times 10}$ &  Regressor matrix based on \(^AV_r\) \tabularnewline
        $\phi_A \in \mathbb{R}^{10}$ &  Inertial parameter vector \tabularnewline
        $ p_A \in \mathbb{R}$ &  Virtual power flow (VPF) \tabularnewline
        $ \nu (t) $ &  Non-negative accompanying function \tabularnewline
        $ \Dot{\nu} (t) $ &  Time derivative of \(\nu (t)\) \tabularnewline
        $ \delta_i,\, i=1,2,3 $ &  Unknown positive contants \tabularnewline
        $ \Delta_R(t)  \in \mathbb{R}^{6}$ &  Rigid body model uncertainty \tabularnewline
        $ \Delta_a(t)  \in \mathbb{R}$ &  Actuator model uncertainty \tabularnewline
        $ D(t) \in \mathbb{R}^{6}$ &  Unknown disturbance \tabularnewline
        $ \Psi(.)$ &  Radial basis function (RBF) \tabularnewline
        \end{tabular}
\end{table}

\begin{table}[h!]
    \centering
    \small
    \begin{tabular}{p{0.2\textwidth}p{0.25\textwidth}}
        $\chi $ &  Input vector of the activation function  \tabularnewline
        $W^* $ &  Optimal weights of neural networks (NNs) \tabularnewline
        $ Z(.) $ & Output of RBFNNs  \tabularnewline
        $ \mathcal{L}_{A} \in \mathbb{R}^{4\times4}$ &  Pseudo-inertia matrix \tabularnewline
        $ h \in \mathbb{R}^{3}$ &  first mass moment \tabularnewline
        $ \Bar{I} \in \mathbb{R}^{6\times6}$ & Rotational inertia matrix  \tabularnewline
        $ m \in \mathbb{R} $ &  mass \tabularnewline
        $  x_{(.)}(t), \Dot{x}_{(.)}(t) $ & position and velocity of the piston \tabularnewline
        $  r_p , r_{wi} $ & gear ratio\tabularnewline
        $ q_j(t), q_{j1}(t), q_{j2}(t) $ & closed-chain angles \tabularnewline
        $ \Dot{q}_j(t), \Dot{q}_{j1}(t), \Dot{q}_{j2}(t)  $ & closed-chain velocities \tabularnewline
        $ L_j, L_{j1}, L_{j2}  $ & Link lengths \tabularnewline
        $ x_{j0}  $ & actuator effective length \tabularnewline
        $ f_{(.)}(t)  $ & Linear hydraulic actuator force \tabularnewline
        $ J(t) \in \mathbb{R}^{6 \times 6} $ & Jacobian matrix \tabularnewline
        $ \Dot{X}_d(t), \Dot{Y}_d(t), \Dot{Z}_d(t)   $ & End-effector desired linear velocity \tabularnewline
        $\Dot{\varpi}_{xd}(t),\,\Dot{\varpi}_{yd}(t),\,\Dot{\varpi}_{zd}(t)   $ & End-effector desired angular velocity \tabularnewline
        $ \Dot{\Pi}_d(t) \in \mathbb{R}^{6} $ & End-effector desired velocity vector \tabularnewline
        $ \Dot{\Theta}_d(t) \in \mathbb{R}^{6}  $ & Joint desired velocity vector \tabularnewline
        $ \lambda, \Bar{\lambda}_{xj}, \sigma_i  $ & Positive constants \tabularnewline
        $  K_A , \Gamma $ & Positive-definite constants \tabularnewline
        $ u  $ & Spool valve voltage  \tabularnewline
        $DZ(.)$ &  Deadzone nonlienarity \tabularnewline
        $BS(.)$ &  Backlash nonlinearity \tabularnewline
        $DB(.)$ &  Deadzone-backlash nonlinearity \tabularnewline
        $  m_d>0, b_r>0, b_l<0 $ & Deadzone parameters \tabularnewline
        $ v^*  $ & Deadzone input  \tabularnewline
        $ u^*,u^*_d  $ & Backlash and desired backlash outputs, respectively \tabularnewline
        $ k_b>0, B_r>0, B_l<0  $ & Backlash parameters \tabularnewline
        $  \Bar{\hat{w}}(t) $ & Estimation of discontinuous function \tabularnewline
        $  \theta \in \mathbb{R}^{5} $ & Deadzone-backlash parameter vector \tabularnewline
        $  \eta \in \mathbb{R}^{5} $ &  Deadzone-backlash regression vector\tabularnewline
        $  \eta_0  $ & Deadzone-backlash mismatch error \tabularnewline
        $ Y_f  \in \mathbb{R}^{1\times 7} $ & Friction regressor matrix \tabularnewline
        $  \phi_f \in \mathbb{R}^{7} $ & Friction parameter vector \tabularnewline
        $  A_a, A_b $ & Cross-sectional area \tabularnewline
        $  p_a(t), p_b(t) $ & Chamber pressures \tabularnewline
        $ p_s  $ & Supply pressure \tabularnewline
        $ p_r  $ & Oil tank pressure \tabularnewline
        $ c_{p1},c_{p2},c_{n1},c_{n2}  $ & Flow coefficients \tabularnewline
        $  Q_a(t), Q_b(t) $ & Flow rate \tabularnewline
        $ s  $ & Maximum stroke of piston \tabularnewline
        $ c_l  $ & Leakage coefficient \tabularnewline
        $  \beta $ & oil bulk modulus \tabularnewline
        $ \theta_v, \theta_d  $ & Actuator parameter vectors \tabularnewline
        $ Y_v(t), Y_d(t)  $ & Regressor matrices \tabularnewline
        $  k_f, k_x $ & Positive constants \tabularnewline
        $ \gamma, \pi, \gamma_{(.)},\gamma_{(.)}  $ & Positive constants \tabularnewline
        $ \hat{\square}  $ & Estimated signals \tabularnewline
        $ \Tilde{\square}  $ & Estimation error (\(\square-\hat{\square}\))\tabularnewline
        \end{tabular}
\end{table}
 
%

\section{Introduction}

\subsection{Heavy-duty Hydraulic Manipulators}

\begin{figure}[t]
      \centering
      \subfloat[]{\includegraphics[width = 0.4\textwidth]{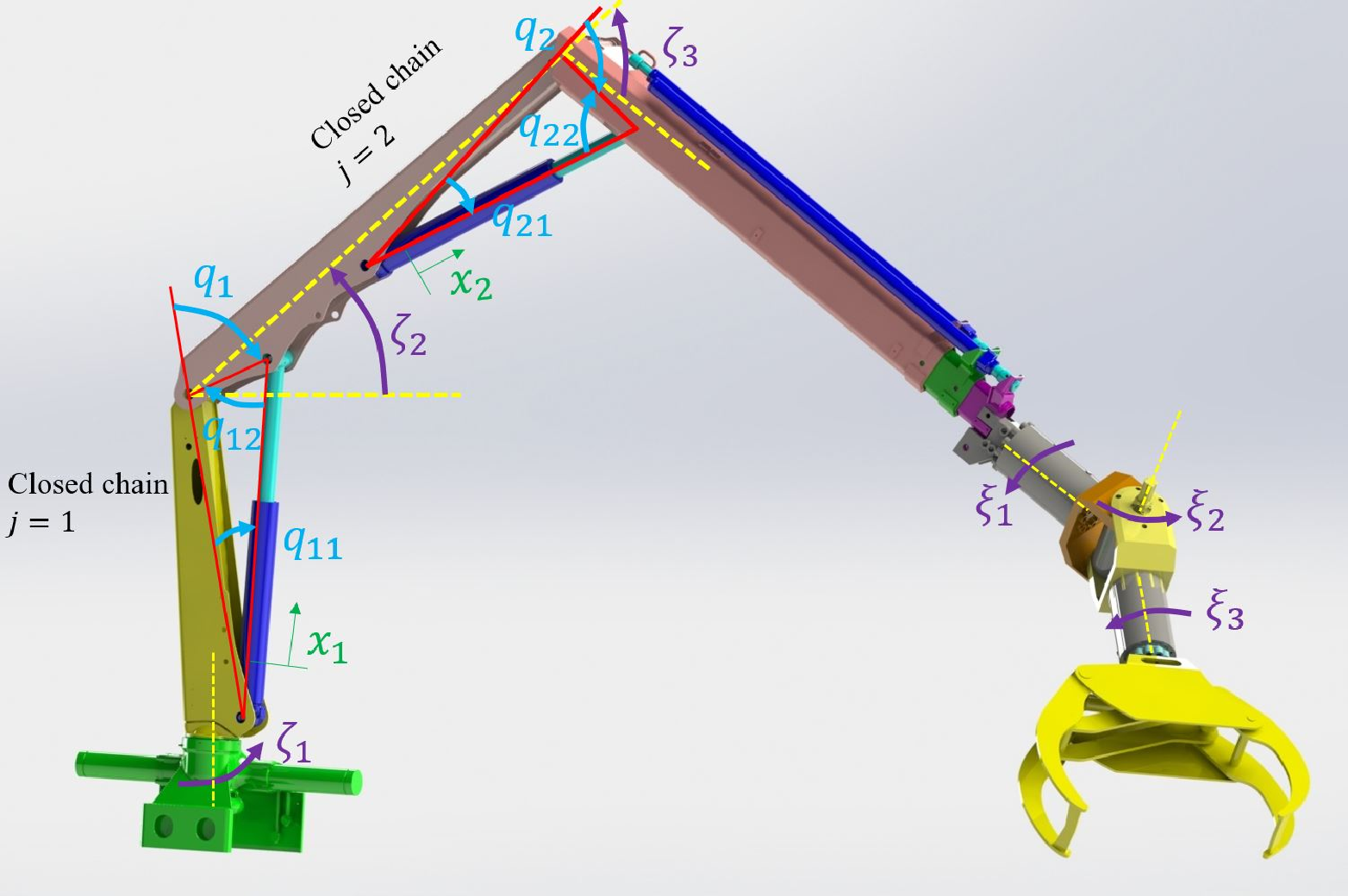}
      \centering
      \label{fIg1a o}}
      \hfil
      \subfloat[]{\includegraphics[width = 0.4\textwidth]{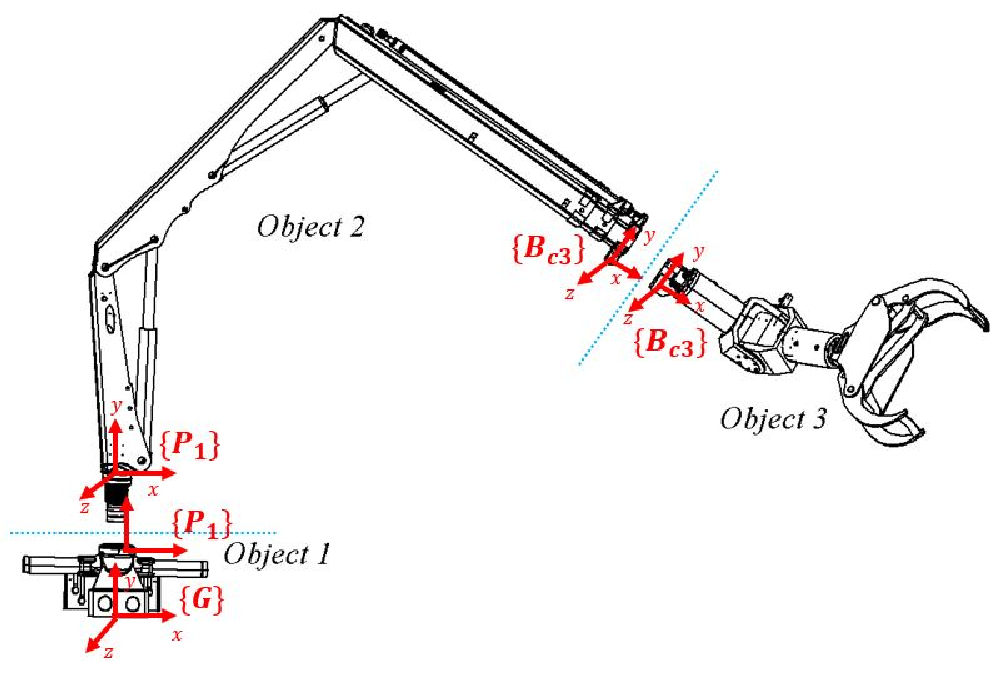}
      \centering
      \label{fIg1b o}}
      \caption{ a) Heavy-duty hydraulic manipulator schematic with kinematic detail, b) Decomposition of the robot into objects: object 1 contains a base joint with hydraulic rack and pinion mechanism, object 2 includes two parallel mechanisms, and object 3 encompasses a spherical hydraulic wrist.}
      \label{Fig Hiab objects}
   \end{figure}

\IEEEPARstart{B}{ecause} of their higher robustness and remarkably larger payload-to-weight ratio (PWR) \cite{chen2023accurate}, heavy-duty hydraulic manipulators (HHMs) have been widely utilized in various fields, including construction, forestry, and agriculture. For example, the commercial manipulator shown in Fig. \ref{Fig Hiab objects} has a lifting capacity of 600 kg at 5 meters reach, with a total weight of 445 kg (PWR almost 1:1). In contrast, the ABB robot \cite{ABB8700} has a lifting capacity of 550 kg at 4.2 meters with the robot weight of 4,600 kg (PWR of 1:8). Such structural rigidity of the electrical manipulators is important to achieve higher accuracy in the manufacturing industry. However, in the field robotic applications, lightweight HHMs are particularly suited for installation on mobile machines, addressing the critical requirement for traversability. {One prominent application of HHMs is in the mining industry, where their exceptional power is crucial for heavy-duty operations in challenging environments such as rock drilling \cite{Sandvik}. Additionally, HHMs have offshore applications, where they are used for heavy-object manipulation in complex conditions. Consequently, it has become crucial to design a controller for HHMs that preserves robustness and precision in various industrial applications.} The current vast amount of industrial investments with increased academic interests will revolutionize the HHMs industry \cite{li2024energy,truong2023backstepping, Sandvik, USExcavator, saidi2016robotics,mattila2017survey}.

Despite their advantages, HHMs suffer from fundamental challenges, such as model nonlinearities and complexities emerging from structural complicatedness and fluid dynamics\cite{mattila2017survey}. In contrast to electric actuators, whose model complexity can be neglected, hydraulic actuators are governed by fluid dynamics and may be subjected to nonsmooth and discontinuous nonlinearities due to friction between cylinder and piston, compound input nonlinearities, and valve under/overlap. Opposite to the electrically actuated open-chain manipulators, HHMs mostly comprise parallel-serial structures \cite{petrovic2022mathematical} that convert linear piston motion to rotational motion. Meanwhile, in rotary hydraulic actuators, the helical gears are employed to convert the piston motion to rotation, the same mechanism in the wrist actuators of the manipulator in this study (Fig. \ref{Fig Hiab objects}), each of which can deliver $2,000 \, Nm$ of torque. The mentioned nonlinearities and complexities render the stability analysis and control design of HHMs significantly challenging, emphasizing the need for robust and high-performance control schemes, not only in theory but also in real-world applications.

\subsection{State-of-the-Art Review}
As explained above, HHMs are characterized by a variety of nonlinearities. Therefore, NMBC can obtain a better performance than linear controllers\cite{bonchis2002experimental}. In the following, the proposed schemes for HHMs control objectives are elaborated on.

\subsubsection{Model-based controller}
In \cite{bech2013experimental}, different NMBC methods, such as sliding mode control (SMC), adaptive inverse dynamics controller (AIDC), and model-reference adaptive controllers with velocity measurement (MRACV), have been designed and compared. The results unveil the better performance of MRACV in comparison to AIDC and SMC. In \cite{bu2000observer}, an observer-based adaptive robust controller (ARC) has been designed to control a hydraulic manipulator, in which the backstepping approach has been employed to derive the control law. However, the experimental results in \cite{bu2001desired} with desired compensation ARC (DCARC) have shown that DCARC yields lower tracking error than ARC. Further, in \cite{zhu2005adaptive} an adaptive output force tracking controller has been designed to control a 6-degrees-of-freedom (DoF) hydraulic manipulator by employing the virtual decomposition control (VDC) scheme, which was first introduced in \cite{zhu2010virtual}. A VDC control, also, in \cite{koivumaki2013automation} has been designed for the precision control of a hydraulic crane. Additionally, in \cite{koivumaki2019energy}, a VDC-based separate meter-in separate meter-out controller has been presented to not only control the 3-DoF HHM but also improve the energy efficiency of the hydraulic actuator, resulting in a 47\% reduction in energy consumption. In \cite{liang2023adaptive}, an adaptive controller with gravity and friction identification has been developed for n-link hydraulic manipulators, and experimental results have been performed for 5-DoF HHM. Moreover, the combination of the time delay control with terminal SMC has been utilized to address the control problems of excavators in \cite{kim2019discrete}. In both \cite{yang2022neural} and \cite{deng2022neural} the radial basic function neural networks (RBFNNs) are employed to estimate uncertainties. Additionally, in \cite{liang2024adaptive} and \cite{yao2024model} finite time tracking control and model-data hybrid control are proposed to control hydraulic manipulators, respectively.

Based on \cite{mattila2017survey}, an appropriate way of evaluating and comparing the proposed methods in the field of HHMs is using the following performance index \cite{zhu2005adaptive}:
\begin{equation}
    \rho = \frac{|\boldsymbol{e}|_{max}}{|\Dot{\boldsymbol{x}}|_{max}}
    \label{rho index}
\end{equation}
with $|\boldsymbol{e}|_{max}$ and $|\Dot{\boldsymbol{x}}_{max}|$ denoting the absolute value of the maximum tracking error and velocity, respectively. A smaller value of $\rho$ demonstrates a better performance for the controller. The $\rho$ value for the controllers in \cite{bech2013experimental}, \cite{bu2001desired}, \cite{zhu2005adaptive},\cite{koivumaki2013automation}, \cite{kim2019discrete}, and \cite{koivumaki2019energy} are 0.0044, 0.005, 0.005, 0.0039, 0.058, and 0.003, respectively. Thus, such a comparison demonstrates that the VDC approach has shown significant performance in comparison to DCARC, SMC, TSMC, and backstepping. Consequently, the VDC scheme has been employed as the baseline controller in the current study.
   
\subsubsection{Input nonlinearities}
By incorporating input nonlinearities, control systems can operate reliably and prevent undesirable outcomes by respecting the physical limitations of the controlled system or device. The adaptive backlash inverse method introduced in \cite{tao1995continuous} and the adaptive deadzone inverse method introduced in \cite{recker1991adaptive} are widely utilized to tackle unknown backlash or deadzone in electric robots \cite{zhao2022deterministic,zhao2023neural}. However, a limited amount of research has been conducted to address input constraint in the field of hydraulic manipulators \cite{li2021valve,lampinen2019model,kang2020almost}. As stated in \cite{li2021valve}, backlash and deadzone existing in the valve and gears can result in performance deterioration and even instability. In \cite{mustalahti2018nonlinear} and \cite{mustalahti2019nonlinear}, the backlash in hydraulic rotary actuators with both rack and pinion mechanism and helical gear has been examined, and a better performance in experiments has been achieved. However, none of the studies considered both backlash and deadzone as compound input nonlinearities in their control method, which is especially crucial for hydraulic actuators because they are more prone to such a compound constraint. In \cite{linan2012controller}, a compound constraint consisting of saturation and backlash/deadzone is converted to an equivalent saturation function using the right inverse of the backlash or deadzone function. However, in real-world applications, the actual parameter of the backlash or deadzone is unavailable, and such an inverse cannot be computed. Therefore, considering the compound input nonlinearities in the control law is an open problem and needs to be addressed.

\subsubsection{Unmodeled uncertainties}
Due to their universality and perfect capabilities in function approximation, RBFNNs are widely used in robotics and control applications to estimate model uncertainty and disturbances \cite{yecsildirek1995feedback,xu2024dynamic}. In \cite{yang2022neural}, a neural adaptive dynamic surface method has been proposed to control the n-link hydraulic manipulator, similar to the traditional backstepping method with the asymptotic convergence of errors. A prescribed performance-based adaptive neural controller has been designed in \cite{deng2022neural} to achieve a good performance for HHM. In \cite{liu2018adaptive}, uncertainties of hydraulic actuators are estimated using RBFNNs and UUB is proved with asymptotic tracking of the desired trajectory, highlighting the importance of achieving UUB in the presence of uncertainties. In the mentioned works, RBFNN is used to estimate disturbances and uncertainties in a rigid body-actuator model, while separately performing the estimations in actuator and rigid body parts will increase the accuracy. This is especially important in HHMs where the governing dynamics of the actuator and rigid body are completely different (the actuator is governed by fluid dynamics), and separate RBFNNs with corresponding input should be designed to learn the nature of the subsystem. It is stated in \cite{sunderhauf2018limits} that learning for complex systems (like uncertainty and disturbance estimation in 6-DoF HHM) should be solved using the decomposition and re-composition methods. Consequently, having a decentralized RBFNNs to deal with actuator and rigid body dynamics can boost the control performance and is of high importance.

\subsubsection{VDC scheme}
In addition to its better performance, the VDC scheme as a baseline controller offers other benefits, as follows: 1) It decomposes the entire complex system into subsystems where the decentralized controller and stability analysis can be performed; 2) the dynamics of each subsystem remains relatively simple, with fixed dynamic structures invariant to the target system; and 3) all uncertainties can be addressed at local subsystems. Moreover, the VDC controller designed for the electric manipulator can be applied to the hydraulic manipulator by only changing the subsystem related to the actuator while the rigid body part remains the same \cite{zhu2010virtual}. This is unlike other methods, in which applying a controller designed for electric robots to hydraulic robots requires redesigning the entire control scheme. Additionally, VDC has displayed a great performance in various fields, such as bilateral teleoperation \cite{lampinen2021force}, physical human-robot interaction \cite{10210088}, and impedance control \cite{koivumaki2015stability}. All the mentioned benefits of VDC express the motivation of this paper to employ it as a baseline controller in this study.

Despite all its benefits, VDC suffers from some drawbacks as well. As expressed in \cite{deng2022neural} and \cite{yang2022neural}, strong stability in the presence of model uncertainty and disturbance for hydraulic manipulators is both critical and in high demand. However, VDC ensures the $L_2$ boundedness for the states of the system which is not strong in the presence of uncertainties, especially when there is compound input nonlinearities, such as deadzone in servo valves and backlash in the gears. Consequently, to have a robust VDC, it is essential to consider unknown model uncertainty, unknown disturbance, and input nonlinearities in the control design while ensuring strong boundedness and stability. These are the open problems in the context of the VDC scheme. Since VDC has shown a better performance than other controllers in the field of HHM, and there is a high demand for HHM in the industry, having a robust VDC would be significantly helpful for paving the way for industrial growth. The contributions of this study properly address the mentioned issues of the VDC.

\subsection{Aim and Contributions}
\subsubsection{Aim}
The aim of the current paper is to achieve a robust and high precision control law for generic 6-DoF HHMs consisting of anthropomorphic arm and spherical wrist. For doing so, the ORC scheme is proposed to consider compound input nonlinearities along with unmodeled uncertainties to achieve a robust and accurate tracking performance.
\subsubsection{Contributions}
Considering all the above-mentioned issues, compound input nonlinearities, unknown model uncertainties, and unknown disturbances are incorporated into control law to ensure the robustness and perfect performance of the controller. Altogether, this has given rise to the ORC scheme: orchestrating all the local controllers at the rigid body and actuator subsystems that are equipped to achieve high-precision control. As such, the contributions of this study can be stated with their significance as follows:
\begin{itemize}
    \item  The compound input nonlinearities model is considered in the modeling part and handled by designing a novel adaptive deadzone-backlash inverse controller. The parametrizable part of the deadzone-backlash model is used to handle parameter uncertainties while the non-parametrizable part, which contains all the nonlinearities, is tackled by novel DRBFNNs. In addition, DRBFNNs are employed to estimate unknown model uncertainties and unknown disturbances in actuator and rigid body subsystems.
    \item Taking the presented ORC into account along with the complexity of the system, semi-global uniformly ultimate boundedness (SGUUB) is achieved for the first time for VDC. Establishing such stability in the context of VDC is mathematically challenging; however, it is a crucial step in ensuring the robustness of the proposed method for real-world applications, marking another significant contribution.
    \item The results are experimentally validated using a 6-DoF commercial hydraulic manipulator, shown in Fig. \ref{Fig Hiab objects}. This is another significance of the presented paper, which showcases the reliability and universality of the proposed method in real-world industrial application.
\end{itemize}

The rest of the paper is organized as follows. Section II expresses the fundamental mathematics of the VDC approach along with the essential lemmas and definitions utilized in this paper. Section III describes the modeling of the rigid body subsystem, whereas in Section IV, the details of the proposed controller are expressed. The low-level control design procedure for the hydraulic actuator is provided in Section V. The stability of the entire system under the proposed approach is proved in Section VI. The experimental and simulation results are provided in Section VII, and the validity of the stability and control results are verified. Finally, Section VIII concludes this study.
   
\section{Mathematical preliminaries}
In this section, the basic mathematical foundation of the VDC approach along with important lemmas and assumptions are presented, which are essential for control design and stability analysis. Fig. \ref{Fig Hiab objects} demonstrates the schematics of the generic 6-DoF HHMs with anothromorphic arm and spherical wrist. Fig. \ref{fIg1a o} shows the kinematic details with base angle ($\zeta_1$), lift angle ($\zeta_2$), tilt angle ($\zeta_3$), and spherical wrist angles of ($\xi_1$),($\xi_2$), and ($\xi_3$). Further, Fig. \ref{fIg1b o} displays the system that decomposed into three objects, each of which will be analyzed separately, by exploiting the virtual cutting point (VCP) concept of VDC.

In the VDC context, spatial velocities and forces/moments are properly integrated within a 6D vector to facilitate the transformation of velocities and forces/moments between different frames. Spatial 6D vectors, which are based on Plücker coordinates, allow the complete representation of both linear and angular motion of multibody mechanisms in a compact manner.

Consider \(\{A\}\) and \(\{B\}\) as frames that are attached to the \(i^{th}\) rigid body. Then, the spatial velocity and force vectors can be expressed as follows \cite{zhu2010virtual}:
\begin{equation*}
^{A}V = [\,^{A}v,\,^{A}\omega]^T,\quad ^{A}F = [\,^{A}f,\,^{A}\tau]^T
\end{equation*}
The transformation matrix that transforms spatial force and velocity vectors between frames \(\{A\}\) and \(\{B\}\) is \cite{zhu2010virtual},
\begin{equation}
^{A}U_{B} = \begin{bmatrix}
^{A}R_{B} & \textbf{0}_{3\times3} \\
(^{A}r_{{A}{B}}\times)\, ^{A}R_{B} & ^{A}R_{B}
\end{bmatrix}
\label{eqn: A_U_B}
\end{equation}
where (\(\times\)) is a skew-symmetric operator defined in \cite{zhu2010virtual}. Based on (\ref{eqn: A_U_B}), the spatial force and velocity vectors can be transformed between frames as \cite{zhu2010virtual},
\begin{equation}
^{B}V =\, ^{A}U_{B}^T\,^{A}V,\quad ^{A}F =\, ^{A}U_{B}\, ^{B}F.
\label{eqn: BV trans}
\end{equation}
The net spatial force vector of the rigid body in frame $\left\lbrace A \right\rbrace$ is \cite{hejrati2022decentralized}:
\begin{equation}
M_{A}\frac{d}{dt}(^{A}V)+C_{A}(^{A}\omega)^{A}V+G_{A}=\, ^{A}F^*.
\label{eqn: net force}
\end{equation}

\begin{prop}
    \cite{hejrati2022decentralized}. The equation (\ref{eqn: net force}) can be written in linear-in-parameter form as below,
    \begin{equation}
    M_{ A} \frac{d}{dt} \left( {^{ A}V} \right) + C_{ A} \left( {^{ A}{\omega}}  \right) {^{ A}V} + G_{ A} = \,\Bar{Y}_{A}\phi_{ A}.
    \label{equ: lin-in-parm}
\end{equation}
\end{prop}

The design variable in the VDC approach is the required velocity that can be designed for either motion control \cite{hejrati2022decentralized} or compliance control \cite{10210088}. As this study examines only the free motion task, the required joint velocity is defined as:
\begin{equation}
\dot{p}_{r} = \dot{p}_{d} + \lambda \, \left(p_{d} - p \right).
\label{required ang vel}
\end{equation}
with $\dot{p}_{d}$ being desired joint velocity, $p_{d}$ being desired joint trajectory, $p$ being measured joint variable, and $\lambda>0$ being a positive constant. The variable $p$ depends on the type of joint; if the joint is revolute, $p$ represents angular measurements, and if the joint is prismatic, $p$ denotes linear motion. Then, the spatial velocity vector, indicated as ${^{ A} V_r}$, can be computed based on $\dot{p}_{r}$ by performing kinematic computation, for which the thorough details will be provided later. Thus, the required spatial force vector can be defined as:
\begin{equation}
{^{ A} F_r^*} = \,Y_{A}\hat{\phi}_{ A} +\, ^{A}F_c .
\label{eqn: tot req force}
\end{equation}
where $ Y_{A}$ is in the sense of (\ref{equ: lin-in-parm}) and $^{A}F_c$ is the regulating control term. The required spatial force vector in (\ref{eqn: tot req force}) demonstrates the amount of spatial force that must be applied to the rigid body subsystem in order to achieve the control objectives at the subsystem level. In the context of VDC, each subsystem (rigid body and actuator) has a corresponding local required control action that accomplishes the local control goals. Then, by orchestrating all these local controllers, the primary goal of the system can be established. Such decomposition into subsystems and orchestration to achieve objectives are due to the VCP and VPF properties of the VDC.
\begin{defka}
    \cite{zhu2010virtual}. A VCP is a directed separation interface that conceptually cuts through a rigid body. At the cutting point, the two parts resulting from the virtual cut maintain an equal position and orientation.
    \label{Def: VCP}
\end{defka}

\begin{defka}
    \cite{zhu2010virtual}. Given the frame $\left\lbrace  A \right\rbrace$, the VPF can be defined as,
    \begin{equation*}
        p_{ A} = ( {^{ A} V_r} -  {^{ A} V} )^T \, ( {^{ A} F_r} -  {^{ A} F} ).
    \end{equation*}
    \label{Def: VPF}
\end{defka}

\begin{defka}
    \cite{zhu2010virtual}. A non-negative accompanying function $\nu(t) \in {R}$ is a piecewise differentiable function that encompasses the following properties:
    i) $\nu(t) \geq 0$ for $t > 0$, and
    ii) $\Dot{\nu}(t)$ exists almost everywhere.
    \label{Def: 4}
\end{defka}
\begin{defka}
    \cite{zhu2010virtual}. A decomposed subsystem is called virtually stable if for the given $\nu(t)$, the $\dot{\nu}(t)\leq -\|B(t)\|+VPFs$ can be established, with $B(t)$ being the positive function.
    \label{Def: 41}
\end{defka}
\begin{defka}
    The following unit vectors are used in the VDC context: ${x}_{f} = (1,0,0,0,0,0)^T$, ${y}_{f} = (0,1,0,0,0,0)^T$, ${z}_{f} = (0,0,1,0,0,0)^T$, ${x}_{\tau} = (0,0,0,1,0,0)^T$, ${y}_{\tau} = (0,0,0,0,1,0)^T$, ${z}_{\tau} = (0,0,0,0,0,1)^T$.
    \label{unit vectors}
\end{defka}

\begin{assumption}
    For the unknown signals, we have,
    \begin{equation*}
        |D(t)|\leq \delta_1, \qquad |\Delta_{R}|\leq \delta_2, \quad |\Delta_{{a}}|\leq \delta_3,
    \end{equation*}
    with \(\delta_1,\, \delta_2,\, \delta_3 \geq 0\) being unknown constants, indicating the upper bounds.
    \label{Assumption: 1}
\end{assumption}

\begin{assumption}
Friction moments in all the rotating joints are neglected.
\label{ass: no friction in joints}
\end{assumption}

\begin{defka}
    \cite{chen2010robust}. RBFNNs can be utilized to estimate an unknown continuous function \(Z(\chi): \mathbb{R}^{\textit{m}} \to \mathbb{R}\) with the approximation of,
    
\begin{equation*}
    Z(\chi) = \hat{W}^T\Psi(\chi) + \hat{\varepsilon}
\end{equation*}
The optimal weight vector \(W^*\) can be expressed by,
\begin{equation*}
    W^* = \arg \min_{\hat{W} \in\, \Xi_N}  \lbrace \sup_{\chi \in\, \Xi_T}|\hat{Z}(\chi|\hat{W})-Z(\chi)| \rbrace
\end{equation*}
where \(\Xi_N = \lbrace \hat{W}|\lVert \hat{W} \rVert \leq \kappa \rbrace\) is a valid set of vectors with \(\kappa\) being a design value, \(\Xi_T\) is an allowable set of the state vectors, and \(\hat{Z}(\chi|\hat{W}) = \hat{W}^T\,\Psi(\chi)\). Additionally, \(\hat{\varepsilon}\), being the estimation error, is a function of input data.
\label{Lemma: 1}
\end{defka}
\begin{rmk}
    While deep neural networks (DNNs) can model complex patterns by stacking multiple layers, this often requires significant computational resources and careful tuning. Since in real-time systems where low latency and fast processing are critical, RBFNNs are often a better choice than DNNs.
\end{rmk}

\begin{defka}
    \cite{hejrati2022decentralized,lee2018natural}. For any inertial parameter vector \(\phi_{A}\), there is a one-to-one linear map \(\mathcal{T}:\mathbb{R}^{10} \rightarrow S(4)\) such that,
\begin{equation*}
    \mathcal{T}(\phi_{A})= \mathcal{L}_{A} = \begin{bmatrix}
        0.5tr(\Bar{I}).\textbf{1}-\Bar{I} & h \\
        h^T & m
        \end{bmatrix}
\end{equation*}
\begin{equation*}
    \mathcal{T}^{-1}(\mathcal{L}_{A}) = \phi_{A}(m,h,tr(\Sigma).\textbf{1}-\Sigma)
\end{equation*}
where \(\Sigma = 0.5tr(\Bar{I})-\Bar{I}\) and \(tr(.)\) is the Trace operator of a matrix. 
\label{Lemma: 2}
\end{defka}

\begin{lem}
    \cite{lee2018natural}. For \(\mathcal{L}_{A}\) defined in Definition (\ref{Lemma: 2}), Bregman divergence with the log-det function can be defined as,
\begin{equation*}
    \mathcal{D}_F(\mathcal{L}_{A}\rVert \hat{\mathcal{L}}_{A}) = log\frac{|\hat{\mathcal{L}}_{A}|}{|\mathcal{L}_{A}|}+tr(\hat{\mathcal{L}}_{A}^{-1}\mathcal{L}_{A})-4,
\end{equation*}
with the time derivative of,
\begin{equation*}
    \Dot{\mathcal{D}}_F(\mathcal{L}_{A}\rVert \hat{\mathcal{L}}_{A}) = tr([\hat{\mathcal{L}}_{A}^{-1}\Dot{\hat{\mathcal{L}}}_{A}\,\hat{\mathcal{L}}_{A}^{-1}]\,\Tilde{\mathcal{L}}_{A}).
\end{equation*}
The Bregman divergence denotes the distance between the actual value $\mathcal{L}_{A}$ and its estimation $\hat{\mathcal{L}}_{A}$ over the manifold $\mathcal{M} \simeq \lbrace \mathcal{L}_{A} \in \mathcal{S}(4): \mathcal{L}_{A} \succ 0 \rbrace = \mathcal{P}(4)$, with $\mathcal{S}(4)$ being space of $4\times4$ real-symmetric matrices and $\mathcal{P}(4)$ being space of $4\times4$ real-symmetric, positive-definite matrices.
\label{Lemma: 3}
\end{lem}

\section{Modeling the Rigid body subsystem}
In this section, equations of motion for the generic 6-DoF HHM are derived. As demonstrated in Fig. \ref{fIg1b o}, the entire complex system is divided into three objects, which are analyzed separately. Such a decomposition may help ease the understanding of the modeling and control design, as the HHM in this study is a complex system consisting of a rack and pinion joint, serial-parallel joints, and hydraulic rotary joints. By employing VDC, each subsystem, representing a rigid body or actuator, is treated individually.

\subsection{Kinematics of the Manipulator}
In this part, kinematic analyses are performed separately for each object, representing a critical step in the VDC context that paves the way to compute total forces applied to rigid bodies.
   
\subsubsection{Kinematics of Object 1}
\begin{figure}[b]
\centering
\includegraphics[width=.5\textwidth]{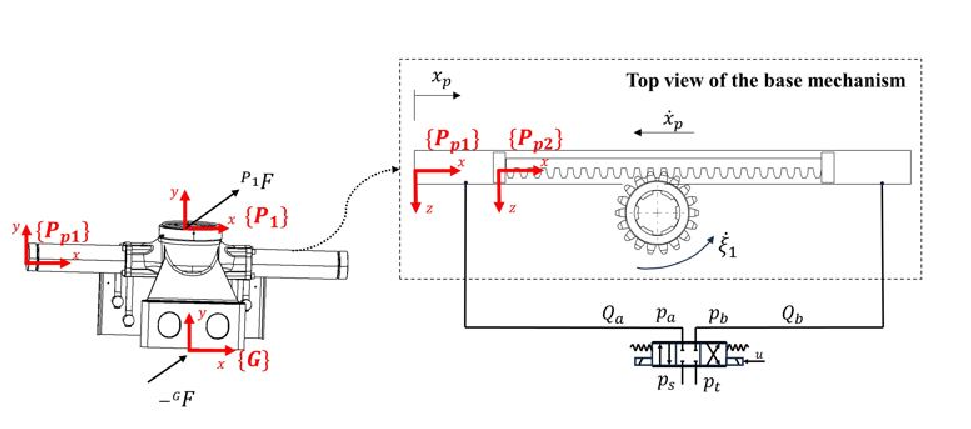}
\caption{VDC frames of object 1} \label{fig: object 1}
\end{figure}

Consider ${^{ G} V}$ to be the spatial velocity vector of the ground. The pillar velocity ${^{ P_1} V}$ can be computed according to Fig. \ref{fig: object 1}:
\begin{equation}
    {^{ P_1} V} = {^{ G}{U_{P_1}^T}}\, {^{ G} V} + y_\tau \Dot{\zeta}_1
    \label{eqn: P1_V}
\end{equation}
with ${^{ G}{U_{P_1}}}$ being in the sense of (\ref{eqn: A_U_B}), and \(\Dot{\zeta}_1\) is generated by the piston's linear motion through the rack and pinion mechanism. According to Fig. \ref{fig: object 1}, the piston spatial velocity can be obtained as:

\begin{equation}
    {^{ P_{p2}} V} = {^{ P_{p1}}{U_{P_{p2}}^T}}\, {^{ P_{p1}} V} + x_f \Dot{x}_p,
    \label{eqn: P_{p1}_V}
\end{equation}
indicating the following relation between the angular velocity of the pillar and the linear velocity of the piston:
\begin{equation}
    \Dot{x}_p = r_p\,\Dot{\zeta}_1
    \label{Dxp}
\end{equation}

\subsubsection{Kinematics of Object 2}

\begin{figure}[t]
\centering
\subfloat[]{%
    \includegraphics[width=0.5\textwidth]{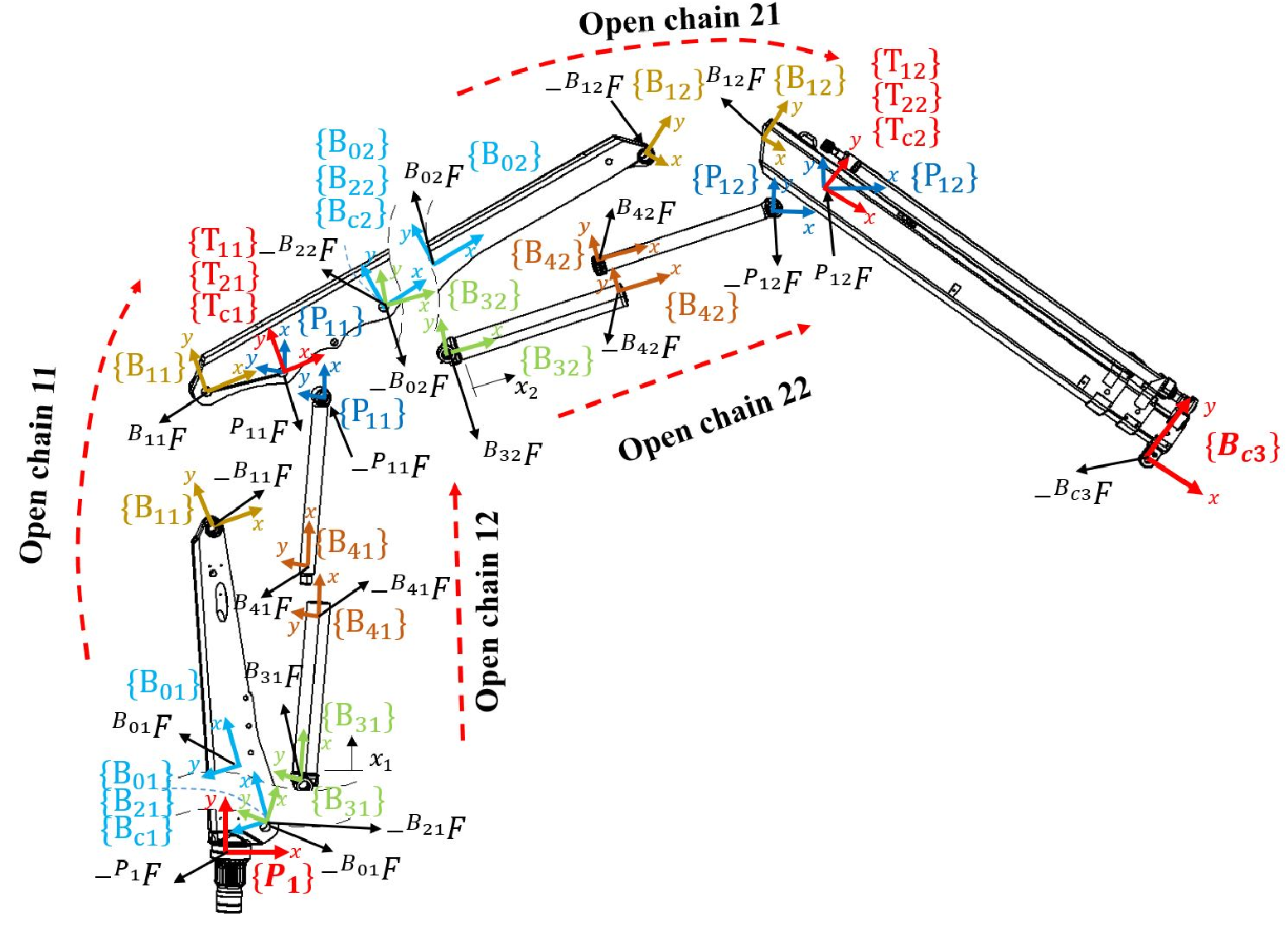}
    \label{fig:object2a}
}%
\hfill
\subfloat[]{%
    \includegraphics[width=0.35\textwidth]{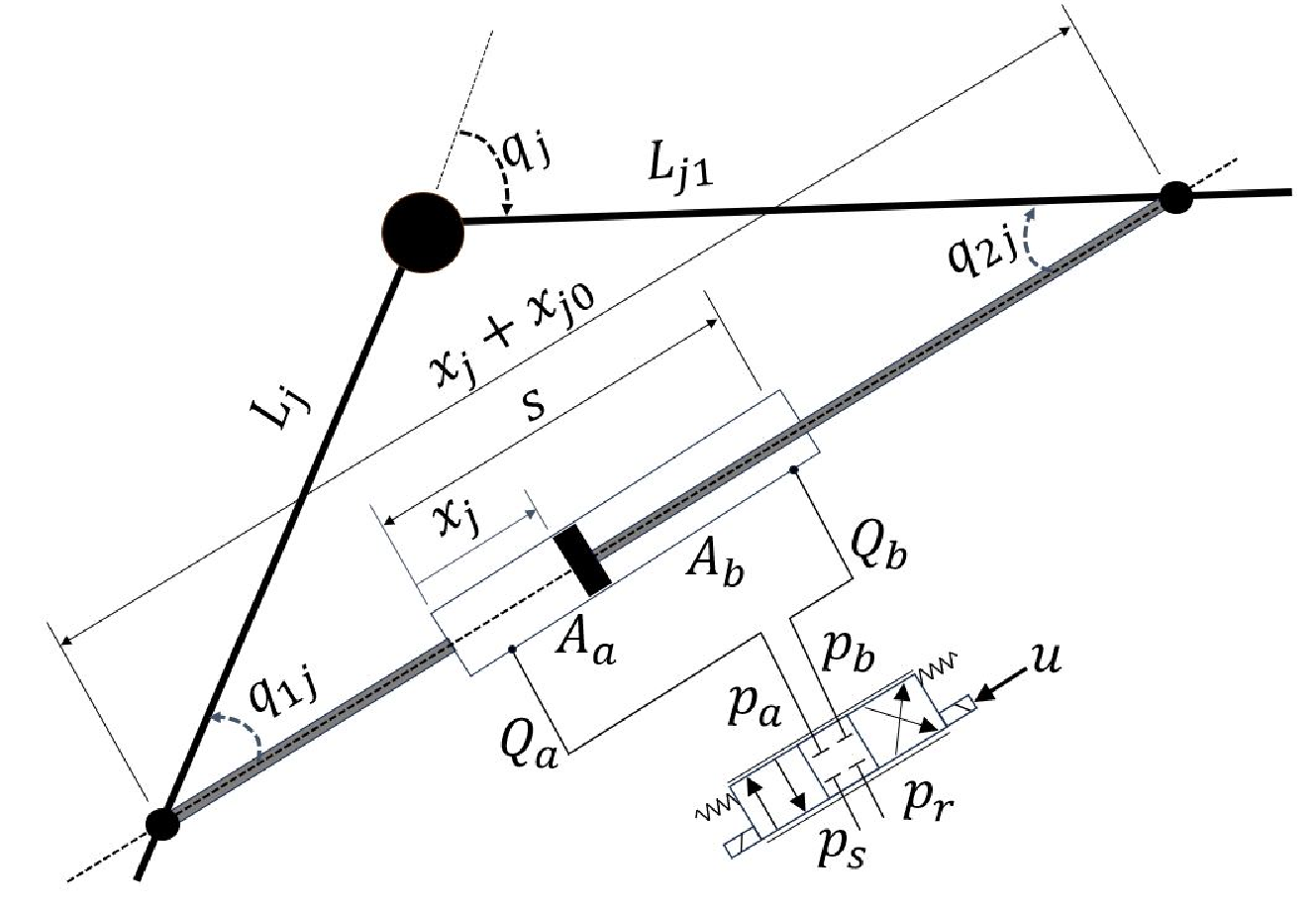}
    \label{fig: object 2b}
}%
\caption{a) VDC frames of object 2, b) Closed chain mechanism with hydraulic actuator}
\label{fig: object 2}
\end{figure}

As demonstrated in Fig. \ref{fIg1b o}, object 2 is composed of two closed-chain kinematics, each of which has important geometric relations that represent constraints in the chain. In the following, the mentioned geometric relations are first expressed, and then spatial velocity vectors are derived. In the notation, $j = 1,2$ represents the first and second closed chains, respectively. Both chains have a passive revolute joint driven by a linear hydraulic actuator, as shown in Fig. \ref{fig: object 2b}. Hence, we have:
\begin{equation}
    q_1 = -2.0736-\zeta_2, \qquad q_2 = \zeta_3-0.4116,
    \label{q1 and q2}
\end{equation}
\begin{equation}
    x_j = \sqrt{L_j^2 + L_{j1}^2 + 2 \, L_j \, L_{j1} \, \cos q_j } - x_{j0},
    \label{eqn: xj}
\end{equation}
\begin{equation}
q_{j1} = -\arccos \left( \dfrac{L_{j1}^2 - (x_j + x_{j0})^2 - L_{j}^2}{-2 \, (x_j + x_{j0}) \, L_j} \right),
\label{eqn: qj1}
\end{equation}
\begin{equation}
q_{j2} = -\arccos \left( \dfrac{L_{j}^2 - (x_j + x_{j0})^2 - L_{j1}^2}{-2 \, (x_j + x_{j0}) \, L_{j1}} \right).
\label{eqn: qj2}
\end{equation}
Differentiating (\ref{eqn: xj})–(\ref{eqn: qj2}) and appropriately transforming them result in:
\begin{equation}
\dot{x}_j = - \dfrac{L_j \, L_{j1} \, \sin q_{j}}{x_j + x_{j0}} \,  \dot{q}_j,
\label{eqn: dxj}
\end{equation}
\begin{equation}
\dot{q}_{j1} = - \dfrac{(x_j + x_{j0}) - L_j \, \cos q_{j1}}{(x_j + x_{j0}) \, L_j \, \sin q_{j1}} \, \dot{x}_j,
\label{eqn: dqj1}
\end{equation}
\begin{equation}
\dot{q}_{j2} = - \dfrac{(x_j + x_{j0}) - L_{j1} \, \cos q_{j2}}{(x_j + x_{j0}) \, L_{j1} \, \sin q_{j2}} \, \dot{x}_j.
\label{eqn: dqj2}
\end{equation}
with $\dot{q}_1 = \dot{\zeta}_2$ and $\dot{q}_2 = \dot{\zeta}_3$. The equations in (\ref{q1 and q2})-(\ref{eqn: dqj2}) reveal the relations between angular motion of passive joints and corresponding piston motion.
Exploiting VCP, closed chains in object 2 are decomposed into subsystems (shown in Fig. \ref{fig:object2a}) through open chains, such that the spatial velocities in the open kinematic chain $1j$ and $2j$ can be written as:
\begin{equation}
{^{ B_{1j}}{V}} = {z}_{\tau} \, \dot{q}_j + {^{ B_{0j}}{U}_{{B_{1j}}}^T} \, {^{ B_{0j}}V},
\label{eqn: B1jV}
\end{equation}
\begin{equation}
{^{ T_{1j}}V} = {^{ B_{1j}}{U}_{{T_{1j}}}^T} \, {^{ B_{1j}}V},
\end{equation}
\begin{equation}
{^{ B_{3j}}V} = {z}_\tau \, \dot{q}_{j1} + {^{ B_{2j}}{U}_{{B_{3j}}}^T} \, {^{ B_{2j}}V},
\label{eqn: B3jV}
\end{equation}
\begin{equation}
{^{ B_{4j}}V} = {x}_f \, \dot{x}_{j} + {^{ B_{3j}}{U}_{{B_{4j}}}^T} \, {^{ B_{3j}}V},  
\label{eqn: B4jV}
\end{equation}
\begin{equation}
{^{ T_{2j}}V} = {z}_\tau \, \dot{q}_{j2} +  {^{ B_{4j}}{U}_{{T_{2j}}}^T} \, {^{ B_{4j}}V}.
\end{equation}
Considering Fig. \ref{fig:object2a}, spatial velocities at the driving VCP of the closed kinematic chains are:
\begin{equation}
{^{ T_{cj}}V}  = {^{ T_{1j}}V}  = {^{ T_{2j}}V}.
\end{equation} 
Note that, from Fig. \ref{fig:object2a}, we have:
\begin{equation}
{^{ B_{c1}}V} = {^{ P_{1}}{U}_{{B_{c1}}}^T} \, {^{ P_1}V},
\label{eqn: Bc1V}
\end{equation}
\begin{equation}
{^{ B_{c2}}V} = {^{ T_{c1}}{U}_{{B_{c2}}}^T} \, {^{ T_{c1}}V},
\label{eqn: Bc2V}
\end{equation}
\begin{equation}
{^{ B_{cj}}V} = {^{ B_{0j}}V} = {^{ B_{2j}}V}.
\label{eqn: BV}
\end{equation}
Finally, spatial velocities measured and expressed in frame $\left\lbrace  B_{c3} \right\rbrace$ of object 2 can be calculated as:
\begin{equation}
{^{ B_{c3}}V} = {^{ T_{c2}}{U}_{{B_{c3}}}^T} \, {^{ T_{c2}}V}.
\label{eqn: E1jV}
\end{equation}
Therefore, all the spatial velocity vectors of object 2 are computed.

\subsubsection{Kinematics of Object 3}
\begin{figure}
\centering
\includegraphics[width=.5\textwidth]{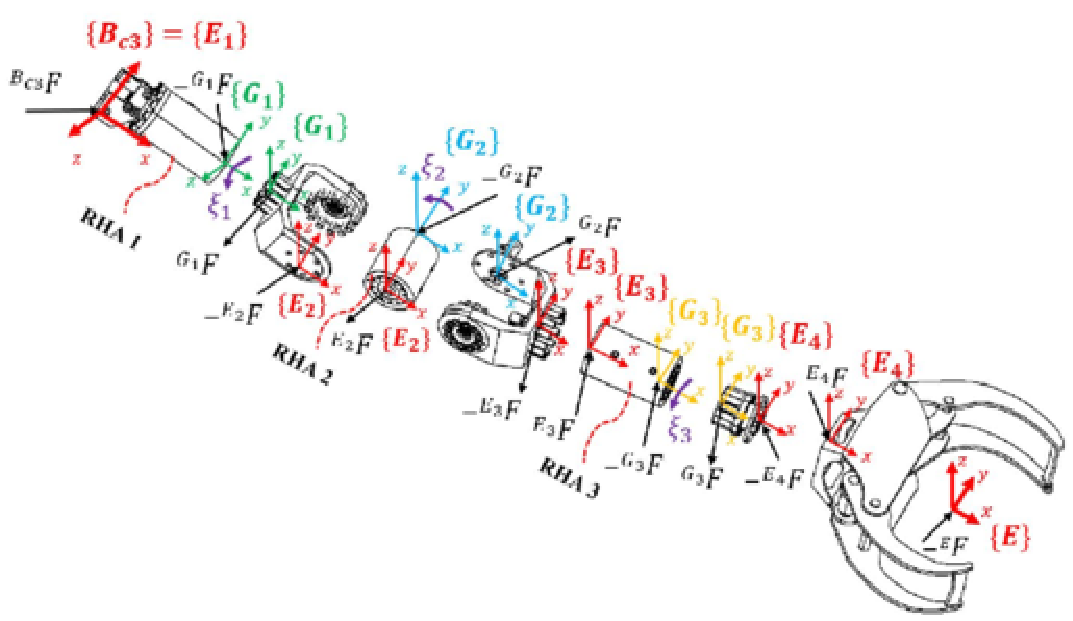}
\caption{VDC frames of object 3. RHA is the abbreviation for rotary hydraulic actuator.} \label{fig: object 3}
\end{figure}

According to Fig. \ref{fig: object 3}, we have:
\begin{equation}
{^{ G_{i}}{V}} = \varrho_i \, \dot{\xi}_i + {^{ E_{i}}{U}_{{G_{i}}}^T} \, {^{ E_{i}}V},
\label{eqn: G_i V}
\end{equation}
\begin{equation}
{^{ E_{i+1}}{V}} = {^{ {G_{i}}}{U}_{ E_{i+1}}^T} \, {^{ G_{i}}V}.
\label{eqn: E_i V}
\end{equation}
with $i=1, 2, 3$, ${^{ E_{1}}V} = {^{ B_{c3}}V}$, and $\varrho_{1,3} = {x}_{\tau}$ and $\varrho_{2} = {y}_{\tau}$. Moreover, $ ^{E}V = {^{E_4}{U}_{ E}^T}\, {^ {E_4}}V$ is the spatial velocity of the end-effector. Additionally, the following relation between angular motion and piston displacement holds:
\begin{equation}
    \Dot{x}_{wi} = r_{wi}\Dot{\xi_i}.
    \label{Dxw}
\end{equation}

\subsection{Dynamics of the Manipulator}
In this section, the total spatial force vectors exerted on each rigid body in each object are derived. The net forces in (\ref{eqn: net force}) can be computed for each rigid body by replacing $ \left\lbrace A \right\rbrace$ with the corresponding frame. In order to consider the unknown disturbance and unknown model uncertainty, we can rewrite (\ref{eqn: net force}) as:
\begin{equation}
M_{ A} \dfrac{\rm d}{\mathrm{d}t} \left( {^{ A}V} \right) + C_{ A} \left( {^{ A}{\omega}}  \right) {^{ A}V} + G_{ A} +\, {^{ A}}\Delta_R(t) = {^{ A} F^*}-\,{^{ A}}D(t).
\label{eqn: tot force 2}
\end{equation}

\subsubsection{Dynamics of Object 3}
The net spatial force vector of object 3 in Fig. \ref{fig: object 3} can be expressed as:
\begin{equation}
{^{ G_{i}} F^*} = {^{ G_{i}} F} - {^{ {G_{i}}}{U}_{ E_{i+1}}} {^{ E_{i+1}} F}.
\label{eqn: net G_i F*}
\end{equation}
which enables us to compute ${^{ G_{i}} F}$:
\begin{equation}
 {^{ G_{i}} F} = {^{ G_{i}} F^*}+{^{ {G_{i}}}{U}_{ E_{i+1}}} {^{ E_{i+1}} F}.
\label{eqn: net G_i F}
\end{equation}
In the same way, one can write:
\begin{equation}
 {^{ E_{i}} F} = {^{ E_{i}} F^*}+{^{ {E_{i}}}{U}_{ G_{i}}} {^{ G_{i}} F}.
\label{eqn: net E_i F}
\end{equation}
for $i = 3,2,1$. The ${^{ E_{4}} F} = {^{ {E_{4}}}{U}_{ E}}\, {^{ E} F}$ with ${^{ E} F}$ is the spatial force sensed from the environment. Consequently, the linear piston force in RHAs can be computed as:
\begin{equation}
f_{cw{i}} = \frac{1}{r_{w {i}}}\varrho_i^T \, {^{ G_{{i}}} F} + \varrho_{i}^T\, {^{ G_{p{i}}} F^*}.
\label{wrist piston force}
\end{equation}
In (\ref{wrist piston force}), the first term indicates the piston force that must overcome rigid body forces, and the second term computes the inertial effect of the piston body.

\subsubsection{Dynamics of Object 2}

The dynamics of the open kinematic chain $2j$ can be represented as:
\begin{equation}
{^{ B_{4j}} F} = {^{ B_{4j}} F^*} + {^{ B_{4j}}{U_{P_{1j}}}}	\, {^{ P_{1j}} F},
\label{eqn: B4jF}
\end{equation}
\begin{equation}
{^{ B_{3j}} F} = {^{ B_{3j}} F^*} + {^{ B_{3j}}{U_{B_{4j}}}}	\, {^{ B_{4j}} F},
\label{eqn: B3jF}
\end{equation}
\begin{equation}
{^{ B_{2j}} F} =  {^{ B_{2j}}{U_{B_{3j}}}}	\, {^{ B_{3j}} F}.
\label{eqn: B2jF}
\end{equation}
The dynamics of  open chain $1j$ can be expressed in the same way as:
\begin{equation}
{^{ B_{1j}} F} = {^{ B_{1j}} F^*} + {^{ B_{1j}}{U_{B_{c{j+1}}}}}	\, {^{ {B_{c{j+1}}} } F}	- {^{ B_{1j}}{U_{P_{1j}}}}	\, {^{ P_{1j}} F}.
\label{eqn: B1jF}
\end{equation}
\begin{equation}
{^{ B_{0j}} F} = {^{ B_{0j}} F^*} +   {^{ B_{0j}}{U_{B_{1j}}}}	\, {^{ B_{1j}} F}.
\label{eqn: B0jF}
\end{equation}
where ${^{ {B_{c{3}}} } F} = {^{ {E_1} } F}$. Considering (\ref{eqn: B4jF})-(\ref{eqn: B0jF}) and provoking theorems 3 and 4 of \cite{petrovic2022mathematical}, the spatial force vector at the driven VCP along with linear actuator force of the parallel-serial joints in object 2, can be derived as:
\begin{equation}
	\begin{array}{ll}
		{^{ B_{cj}} F} = & {^{ B_{0j}} F^*} + {^{ B_{0j}}{U_{B_{1j}}}}  {^{ B_{1j}} F^*}  + {^{ B_{2j}}{U_{B_{3j}}}}  {^{ B_{3j}} F^*} + \\ & {^{ B_{2j}}{U_{B_{3j}}}}   {^{ B_{3j}}{U_{B_{4j}}}} {^{ B_{4j}} F^*} + {^{ B_{0j}}{U_{B_{1j}}}} {^{ B_{1j}}{U_{E_{1j}}}}	\, {^{ E_{1j}} F}
	\end{array}
	\label{eqn: driven cc force}
\end{equation}
\begin{equation}
	\begin{array}{ll}
		f_{cj} =  &  {x}_f^T \, {^{{B_{4j}}}F^*}     - \dfrac{{z}_{\tau}^T \left( {^{ B_{1j}}F^*} + {^{ B_{1j}}{U_{E_{1j}}}}	\, {^{ E_{1j}}F} \right)}{L_{j1} \, \sin q_{j2}}   \\
		& -\dfrac{{z}_{\tau}^T \, ({^{{B_{3j}}}F^*}) + {z}_{\tau}^T \, ({^{{B_{4j}}}F^*}) }{(x_j + x_{j0} ) \, \tan q_{j2}}\\
        &- \dfrac{ {y}_{f}^T \left( {^{{B_{4j}}}F^*} \right) \, (x_j + x_{j0}  - l_{cj})}{(x_j + x_{j0} ) \, \tan q_{j2}}.
	\end{array}
	\label{eqn: piston force}
\end{equation}
Now, with having ${^{ B_{c1}} F}$, the dynamics of the object 1 can be derived.

\begin{corollary}
The following expressions can be written and assumed to be valid:
\begin{equation}
	{z}_{\tau}^T \, {^{ B_{1j}} F} =\, {z}_{\tau}^T \, {^{ P_{1j}} F}= \, {z}_{\tau}^T \, {^{ B_{3j}} F} = 0.
	\label{eqn: fric qj}
\end{equation}	
\end{corollary}

\subsubsection{Dynamics of Object 1}
Having ${^{ B_{c1}} F}$ from object 2 and considering Fig. \ref{fig: object 2}, one can obtain the spatial force vector of the pillar as:
\begin{equation}
 {^{ P_{1}} F} =\, {^{ P_{1}} F^*} +  {^{ {P_{1}}}{U}_{ B_{c1}}} {^{ B_{c1}} F}.
\label{eqn: Pillar Force}
\end{equation}
which results in linear piston force in object 1 as below:
\begin{equation}
f_{cp} = \frac{1}{r_{p}} y_\tau^T \, {^{ P_{1}} F} + x_f^T {^{ P_{p2}} F^*}.
\label{base piston force}
\end{equation}
${^{ P_{p2}} F^*}$ is the net spatial force vector of the piston body.

\begin{rmk}
    {The equations in \mbox{(\ref{eqn: P1_V})-(\ref{Dxw})} represent the kinematics of the entire system, which is decomposed into subsystems, as shown in Figs. \mbox{\ref{fig: object 1}-\ref{fig: object 3}}. These equations are required for force analysis in \mbox{(\ref{eqn: net G_i F})-(\ref{base piston force})}, which are the representation of the system dynamics. The velocities and forces of each subsequent frame are computed through the transformation matrix defined in \mbox{(\ref{eqn: A_U_B})}. This approach ensures that the analysis is grounded in precise, systematic computations, facilitating the accurate modeling of the manipulator's behavior.}
\end{rmk}

\section{Control Design of the rigid body subsystem}
In this section, for the modeled 6-DoF HHM, the procedure of local controller design is expressed, where the main objective of the system can be achieved by orchestrating all the local controllers.
\subsection{Desired Trajectory}
The kinematic relation between the Cartesian space and joint space for the 6-DoF manipulator can be written as:
\begin{equation}
     \Dot{\Pi}_d =  J \Dot{\Theta}_d
    \label{TS pose vel}
\end{equation}
which results in:
\begin{equation}
    \Dot{\Theta}_d =  J^{-1}  \Dot{\Pi}_d.
    \label{JS ang vel}
\end{equation}
It can be seen from (\ref{JS ang vel}) that the desired trajectory of working space can be projected onto a joint space trajectory. It means that precise joint control will end up with precise end-effector pose control. This study aims to design a precise controller to achieve a considerably low tracking error in joint space.

\subsection{Required Velocity Vectors}
By solving the inverse kinematic problem in (\ref{JS ang vel}), the desired angular velocity and desired joint angle can be computed. Then, the required angular velocity in the sense of (\ref{required ang vel}) can be derived. As expressed in (\ref{Dxp}) and (\ref{Dxw}), the linear motion is converted to angular motion through a constant parameter. 

However, the expression in (\ref{eqn: dxj}) is complex and depends on the kinematics of the closed chain. Consequently, we define the required velocity of objects 1 and 3 based on the angular motion, and for object 2 based on piston motion. It is straightforward to compute the desired piston velocity and position for object 2 using (\ref{JS ang vel}), (\ref{q1 and q2}), (\ref{eqn: xj}), and (\ref{eqn: dxj}) by replacing the actual signals with desired ones. Therefore, the required joint velocities can be defined as:
\begin{eqnarray}
\dot{\zeta}_{1r} = \dot{\zeta}_{1d} + \lambda \, (\zeta_{1d} - \zeta_1),
\label{eqn: zeta1r}
\end{eqnarray}
\begin{eqnarray}
\dot{x}_{jr} = \dot{x}_{jd} + \Bar{\lambda}_{xj} \, (x_{jd} - x_j),
\label{eqn: xjr}
\end{eqnarray}
\begin{eqnarray}
\dot{\xi}_{ir} = \dot{\xi}_{id} + \sigma_{i} \, (\xi_{1d} - \xi_i),
\label{eqn: xir}
\end{eqnarray}
with $j = 1,2$, $i = 1,2,3$. By replacing (\ref{eqn: zeta1r})-(\ref{eqn: xir}) in the kinematic part of each object and reusing (\ref{Dxp}), (\ref{eqn: dxj})-(\ref{eqn: dqj2}), and (\ref{Dxw}), the required spatial velocities can be achieved as:
\begin{equation}
    {^{ P_1} V_r} = {^{ G}{U_{P_1}^T}}\, {^{ G} V_r} + y_\tau \Dot{\zeta}_{1r}
    \label{eqn: P1_Vr}
\end{equation}
\begin{equation}
    {^{ P_{p2}} V_r} = {^{ P_{p1}}{U_{P_{p2}}^T}}\, {^{ P_{p1}} V_r} + x_f \Dot{x}_{pr}
    \label{eqn: E2_Vr}
\end{equation}
\begin{equation}
{^{ B_{1j}}{V}_r} = {z}_{\tau} \, \dot{q}_{jr} + {^{ B_{0j}}{U}_{{B_{1j}}}^T} \, {^{ B_{0j}}V_r},
\label{eqn: B1jVr}
\end{equation}
\begin{equation}
{^{ T_{1j}}V_r} = {^{ B_{1j}}{U}_{{T_{1j}}}^T} \, {^{ B_{1j}}V_r},
\end{equation}
\begin{equation}
{^{ B_{3j}}V_r} = {z}_\tau \, \dot{q}_{j1r} + {^{ B_{2j}}{U}_{{B_{3j}}}^T} \, {^{ B_{2j}}V},
\label{eqn: B3jVr}
\end{equation}
\begin{equation}
{^{ B_{4j}}V_r} = {x}_f \, \dot{x}_{jr} + {^{ B_{3j}}{U}_{{B_{4j}}}^T} \, {^{ B_{3j}}V_r},  
\label{eqn: B4jVr}
\end{equation}
\begin{equation}
{^{ T_{2j}}V_r} = {z}_\tau \, \dot{q}_{j2r} +  {^{ B_{4j}}{U}_{{T_{2j}}}^T} \, {^{ B_{4j}}V_r}.
\end{equation}
\begin{equation}
{^{ B_{c1}}V_r} = {^{ P_{1}}{U}_{{B_{c1}}}^T} \, {^{ P_1}V_r},
\label{eqn: Bc1Vr}
\end{equation}
\begin{equation}
{^{ B_{c2}}V_r} = {^{ T_{c1}}{U}_{{B_{c2}}}^T} \, {^{ T_{c1}}V_r},
\label{eqn: Bc2Vr}
\end{equation}
\begin{equation}
{^{ B_{c3}}V_r} = {^{ T_{c2}}{U}_{{B_{c3}}}^T} \, {^{ T_{c2}}V_r}.
\label{eqn: E1jVr}
\end{equation}
\begin{equation}
{^{ G_{i}}{V}_r} = \varrho_i \, \dot{\xi}_{ir} + {^{ E_{i}}{U}_{{G_{i}}}^T} \, {^{ E_{i}}V_r},
\label{eqn: G_i Vr}
\end{equation}
\begin{equation}
{^{ E_{i+1}}{V}_r} = {^{ {G_{i}}}{U}_{ E_{i+1}}^T} \, {^{ G_{i}}V_r}.
\label{eqn: E_i Vr}
\end{equation}
The required joint velocities in (\ref{eqn: zeta1r})-(\ref{eqn: xir}) show the required trajectory that each joint must follow in order to achieve the local control objectives. On the other side, the required spatial velocity vectors in (\ref{eqn: P1_Vr})-(\ref{eqn: E_i Vr}) display the velocity that the rigid body should acquire to track the desired motion. In the following, the local controller will be designed in a way that provides the required effort to establish the desired motion.

\subsection{Required Force Vectors}
Considering the presented model in (\ref{eqn: tot force 2}), the required net spatial force vector can be designed in the sense of (\ref{eqn: tot req force}) by taking advantage of (\ref{equ: lin-in-parm}) as:
\begin{equation}
    {^{ A} F_r^*} = Y_{A}\hat{\phi}_{ A} + K_{ A} \, \left({^{ A} V_r} - {^{ A} V} \right) +\, {^{ A}} \Delta_R +\, {^{ A}} D(t).
    \label{Fr des}
\end{equation}
However, the model uncertainty ${^{ A}} \Delta_R$ is unknown in real-world applications. Additionally, the magnitude and frequency of the external disturbances are unknown for many applications, especially for industrial manipulators. In this study, adaptive control and DRBFNN approaches are utilized to address the mentioned issues.  According to Definition \ref{Lemma: 1}, we can define:
\begin{equation}
    {^{ A}} \Delta_R = {^{ A} }W^T\Psi(\chi_{ A}) +\, {^{ A}}\varepsilon^*
    \label{Delta def}
\end{equation}
where ${^{ A} }W \in \mathbb{R}^{6\times \Bar{n}_A}$ is the RBFNNs weight, $\Psi(.)$ is Gaussian activation function, $\chi_{ A} = [{^{ A} V}^T, {^{ A} V_r^T}, {^{ A} \Dot{V}_r^T}]^T \in \mathbb{R}^{18}$, ${^{ A}}\varepsilon^* \in \mathbb{R}^{6}$ is the RBFNNs approximation error, and $\Bar{n}_A$ is the number of neurons in rigid body subsystem $\lbrace A \rbrace$. By changing $\Bar{n}_A$, one can set different numbers of nodes for different rigid body subsystems. Since the actual values for ${^{ A} }W$ and  ${^{ A}}\varepsilon^*$ are not available, their estimations, denoted ${^{ A} }\hat{W}$ and ${^{ A}}\hat{\varepsilon}^*$ are utilized in the control design. By defining ${^{ A}}\varepsilon = {^{ A}}\varepsilon^* + {^{ A}} D(t)$, we can rewrite (\ref{Fr des}) as:
\begin{equation}
    {^{ A} F_r^*} = Y_{A}\hat{\phi}_{ A} + K_{ A} \, \left({^{ A} V_r} - {^{ A} V} \right) +\, {^{ A} }\hat{W}^T\Psi(\chi_{ A}) +\, {^{ A}}\hat{\varepsilon}
    \label{Fr des final}
\end{equation}
with ${^{ A}}\hat{\varepsilon}$ being the estimation of ${^{ A}}\varepsilon$. In (\ref{Fr des final}), the term ${^{ A}}\hat{\varepsilon}$ estimates both the bias of the RBFNNs ${^{ A}}\varepsilon^*$ and the external disturbance $D(t)$. Then, the required spatial force vectors can be computed by evoking spatial force vectors and using (\ref{Fr des final}) as:
\begin{equation}
 {^{ G_{i}} F_r} = {^{ G_{i}} F_r^*}+{^{ {G_{i}}}{U}_{ E_{i+1}}} {^{ E_{i+1}} F_r}
\label{eqn: net G_i Fr}
\end{equation}
\begin{equation}
 {^{ E_{i}} F_r} = {^{ E_{i}} F_r^*}+ {^{ {E_{i}}}{U}_{ G_{i}}} {^{ G_{i}} F_r}
\label{eqn: net E_i Fr}
\end{equation}
\begin{equation}
{^{ B_{1j}} F_r} = {^{ B_{1j}} F_r^*} + {^{ B_{1j}}{U_{B_{c,{j+1}}}}}	\, {^{ {B_{c,{j+1}}} } F_r}	- {^{ B_{1j}}{U_{P_{1j}}}}	\, {^{ P_{1j}} F_r}.
\label{eqn: B1jFr}
\end{equation}
\begin{equation}
{^{ B_{0j}} F_r} = {^{ B_{0j}} F_r^*} +   {^{ B_{0j}}{U_{B_{1j}}}}	\, {^{ B_{1j}} F_r}.
\label{eqn: B0jFr}
\end{equation}
\begin{equation}
	\begin{array}{ll}
		{^{ B_{cj}} F_r} = & {^{ B_{0j}} F_r^*} + {^{ B_{0j}}{U_{B_{1j}}}}  {^{ B_{1j}} F_r^*}  + {^{ B_{2j}}{U_{B_{3j}}}}  {^{ B_{3j}} F_r^*}\\
        &+ {^{ B_{2j}}{U_{B_{3j}}}}   {^{ B_{3j}}{U_{B_{4j}}}} {^{ B_{4j}} F_r^*} \\ 
        &+ {^{ B_{0j}}{U_{B_{1j}}}} {^{ B_{1j}}{U_{E_{1j}}}}	\, {^{ E_{1j}} F_r}
	\end{array}
	\label{eqn: driven cc force req}
\end{equation}
\begin{equation}
 {^{ P_{1}} F_r} = {^{ P_{1}} F_r^*} + {^{ {P_{1}}}{U}_{ B_{c1}}} {^{ B_{c1}} F_r}
\label{eqn: Pillar Force req}
\end{equation}
with $i = 3,2,1$ and $j = 2,1$. Finally, the required piston forces that must be applied at the actuator level in order to accomplish the control objective can be derived as:
\begin{equation}
f_{cwir} = \frac{1}{r_{wi}}\varrho_i^T \, {^{ G_{i}} F_r} + \varrho_i^T {^{ G_{pi}} F_r^*}
\label{wrist piston force req}
\end{equation}
\begin{equation}
	\begin{array}{ll}
		f_{cjr} =  &  {x}_f^T \, {^{{B_{4j}}}F_r^*}     - \dfrac{{z}_{\tau}^T \left( {^{ B_{1j}}F_r^*} + {^{ B_{1j}}{U_{E_{1j}}}}	\, {^{ E_{1j}}F_r} \right)}{L_{j1} \, \sin q_{j2}}   \\
		& -\dfrac{{z}_{\tau}^T \, ({^{{B_{3j}}}F_r^*}) + {z}_{\tau}^T \, ({^{{B_{4j}}}F_r^*})}{(x_j + x_{j0} ) \, \tan q_{j2}}\\
        &+\dfrac{ {y}_{f}^T \left( {^{{B_{4j}}}F_r^*} \right) \, (x_j + x_{j0}  - l_{cj})}{(x_j + x_{j0} ) \, \tan q_{j2}}
	\end{array}
	\label{eqn: piston force req}
\end{equation}
\begin{equation}
f_{cpr} = \frac{1}{r_{p}} y_\tau^T \, {^{ P_{1}} F_r} + x_f^T\, {^{ P_{p1}} F_r^*}.
\label{base piston force req}
\end{equation}

\begin{rmk}
    {Building upon the system's kinematics and dynamics analyses presented in Section III, this section proposes a robust control algorithm designed to achieve the system's objectives effectively. The equations in \mbox{(\ref{eqn: zeta1r})-(\ref{eqn: E_i Vr})} describe the expected kinematic behavior of the system when following the desired commands, laying the foundation for our control strategy. Leveraging this model-informed data, we then calculate the necessary forces to enforce the desired system behavior as in \mbox{(\ref{eqn: net G_i Fr})-(\ref{base piston force req})}. Furthermore, the proposed algorithm accounts for unknown model uncertainties by incorporating RBFNNs as detailed in \mbox{(\ref{Fr des final})}, ensuring robustness in the face of unpredictable real-world conditions.}

\end{rmk}

\begin{rmk}
    The control laws in (\ref{wrist piston force req})-(\ref{base piston force req}) have three local goals: i) canceling the inertial effect of the rigid body, ii) compensating for unknown disturbances and model uncertainty, and iii) having a precise trajectory tracking. As demonstrated, the required action for the rigid body part and actuator part are computed separately, avoiding coupling nonlinearities. However, the required actuator force in (\ref{wrist piston force req})-(\ref{base piston force req}) encompasses the rigid body-joint coupling nonlinearities, a significant feature of VDC that deals with each subsystem separately.    
\end{rmk}

In the following section, by taking into account the complex and nonlinear model of the hydraulic actuator, the low-level controller is designed to ensure the required forces in (\ref{wrist piston force req})-(\ref{base piston force req}) will be generated at the actuator level. Since the procedure of modeling for all the hydraulic actuators in this study is the same, we derived the required voltage for a given linear piston force $f_c$, similar to (\ref{wrist piston force}), (\ref{eqn: piston force}), and (\ref{base piston force}), and required piston force $f_{cr}$ as in (\ref{wrist piston force req})-(\ref{base piston force req}). 

\section {Modeling and control of actuator subsystem}

In this section, first, for a given desired voltage $u_d^*$ and voltage applied to the system in the presence of deadzone-backlash nonlinearity $u^*$, the error $u^*-u_d^*$ is derived which later will be incorporated into stability analysis to ensure robustness and to handle parameter uncertainties. Then, the required voltage is computed for a given required piston force.

\subsection{Inverse Deadzone-Backlash}
Suppose that $u_d^*$ is a control signal that may accomplish control objectives in the absence of input nonlinearities, while $v^*$ is the control signal with deadzone-backlash compensation. The deadzone function can be defined as,
\begin{equation}
    v_1 = DZ(v^*)= \left\lbrace \begin{array}{cc}
	 m_d(v^*-b_r), & v^*\geq b_r,\\
	 0, & b_l<v^*<b_r \\
	 m_d(v^*-b_l), & v^*\leq b_l
\end{array} \right.
\label{eqn: deadzone}
\end{equation}
Backlash function can be defined as,
\begin{equation}
    u^* = BS(v_1)= \left\lbrace \begin{array}{cc}
	 k_b(v_1-B_r), & \Dot{v}_1> 0,\\
	 k_b(v_1-B_l), & \Dot{v}_1< 0 \\
	 u^*(t_-), & otherwise
\end{array} \right.
\label{eqn: backlash}
\end{equation}
By substituting (\ref{eqn: deadzone}) in (\ref{eqn: backlash}), one can obtain,
\begin{equation}
    u^* = DB(v^*)= \left\lbrace \begin{array}{cc}
	 c(v^*-b_r)-d_r, & \Dot{v}_1> 0\, \& \,v^*\geq b_r\\
	 -d_r, & \Dot{v}_1> 0\, \& \,b_l<v^*<b_r\\
	 c(v^*-b_l)-d_r, & \Dot{v}_1> 0\, \& \,v^*\leq b_l\\
      c(v^*-b_r)-d_l, & \Dot{v}_1< 0\, \& \,v^*\geq b_r,\\
	 -d_l, & \Dot{v}_1< 0\, \& \,b_l<v^*<b_r \\
	 c(v^*-b_l)-d_l, & \Dot{v}_1< 0\, \& \,v^*\leq b_l\\
      u^*(t_-), & otherwise
\end{array} \right.
\label{eqn: backlash deadzone}
\end{equation}
After designing the control voltage in the absence of compound input nonlinearities (\ref{eqn: backlash deadzone}), the control law can be enhanced to cancel the damaging effect of such a constraint, such that $u_d^* = DB(DBI(u_d^*))$ with $DBI(.)$ being the inverse deadzone-backlash function. On the other hand, in real-world applications, the parameters in (\ref{eqn: backlash deadzone}) are unknown, and, it is desirable to estimate them. Consequently, we can write the adaptive inverse deadzone-backlash control term as,
\begin{equation}
    v^* = \frac{1}{\hat{c}}u_d^*+\Bar{\hat{w}}(t)
    \label{v BDI}
\end{equation}
with 
\begin{equation}
    \hat{w}(t) = \left\lbrace \begin{array}{cc}
	 \frac{\hat{d}_r}{\hat{c}}+\hat{b}_r, & \Dot{u}^*_d> 0\, \& \,u^*_d> 0,\\
	 \frac{\hat{d}_r}{\hat{c}}+\hat{b}_l, & \Dot{u}^*_d> 0\, \& \,u^*_d< 0\\
      \frac{\hat{d}_l}{\hat{c}}+\hat{b}_r, & \Dot{u}^*_d< 0\, \& \,u^*_d> 0,\\
	 \frac{\hat{d}_l}{\hat{c}}+\hat{b}_l, & \Dot{u}^*_d< 0\, \& \,u^*_d< 0,\\
       w(t_-)            & otherwise.
\end{array} \right.
\label{eqn: w}
\end{equation}

As the signal $\hat{w}(t)$ in (\ref{eqn: w}) is discontinuous when $\Dot{u}_d^*$ and $u_d^*$ changes the sign, we utilize the following differential equation as a continuous approximation of $\hat{w}(t)$\cite{tao1995continuous}, which is used in (\ref{v BDI}):
\begin{equation}
    \Dot{\Bar{\hat{w}}}(t) = \alpha(-\Bar{\hat{w}}(t)+\hat{w}(t))
    \label{w_dot}
\end{equation}
with $\alpha>0$ and $\Bar{\hat{w}}(0) \in [\frac{d_l}{c}+b_l, \frac{d_r}{c}+b_r]$, and $\hat{c}$, $\hat{d}_r$,$\hat{d}_l$,$\hat{b}_r$, and $\hat{b}_l$ being the estimation of $c$, $d_r$,$d_l$,$b_r$, and $b_l$, respectively. 
Further, to derive an adaptation law for the estimation of unknown parameters we need to parameterize the control error $u^*-u_d^*$ utilizing (\ref{eqn: backlash deadzone}) and (\ref{v BDI}). To do so, (\ref{v BDI}) can be rewritten as,
\begin{equation}
    v^*(t) = \frac{1}{\hat{c}}(u_d^*+\varphi(\Dot{u}_d^*)\hat{d}_r+\varphi(-\Dot{u}_d^*)\hat{d}_l)+\varphi(u_d^*)\hat{b}_r+\varphi(-u_d^*)\hat{b}_l
    \label{v expand}
\end{equation}
where the smooth function $\varphi(\kappa) = (\tanh([\kappa-\kappa_0]/x_0)+1)/2$, with $\kappa_0$ and $x_0$ being small constants, is utilized to compensate for the chattering resulting from discontinuity in (\ref{v BDI}). Considering $\hat{\theta} = [\hat{c},\widehat{cb}_r,\widehat{cb}_l,\hat{d}_r,\hat{d}_l]^T$ with $\hat{b}_r = \widehat{cb}_r/\widehat{c}$, $\hat{b}_l = \widehat{cb}_l/\hat{c}$, $\eta = [-v,\varphi(\Dot{u}_d^*),\varphi(-\Dot{u}_d^*),\varphi(u_d^*),\varphi(-u_d^*)]^T$, (\ref{v expand}) can be written in parameterized way:
\begin{equation}
    u_d^*(t) = -\hat{\theta}^T\eta.
\end{equation}
Subsequently, the adaptive deadzone-backlash inverse error can be written as,
\begin{equation}
    u^*(t)-u_d^*(t) = -(\theta-\hat{\theta})^T\eta + \eta_0 = -\Tilde{\theta}^T\eta +\eta_0.
    \label{DB error}
\end{equation}

\begin{rmk}
    {As demonstrated, the complex and highly nonlinear nature of the compound input nonlinearities, as described in \mbox{(\ref{eqn: backlash deadzone})}, is transformed into a more manageable error dynamic representation in \mbox{(\ref{DB error})}. This compact formulation enables the incorporation of these challenging nonlinearities directly into the control law, allowing for effective handling of them to ensure both robustness and precision in system performance.} 
\end{rmk}

\begin{figure}[b]
      \centering
      \subfloat[]{\includegraphics[width = 0.5\textwidth]{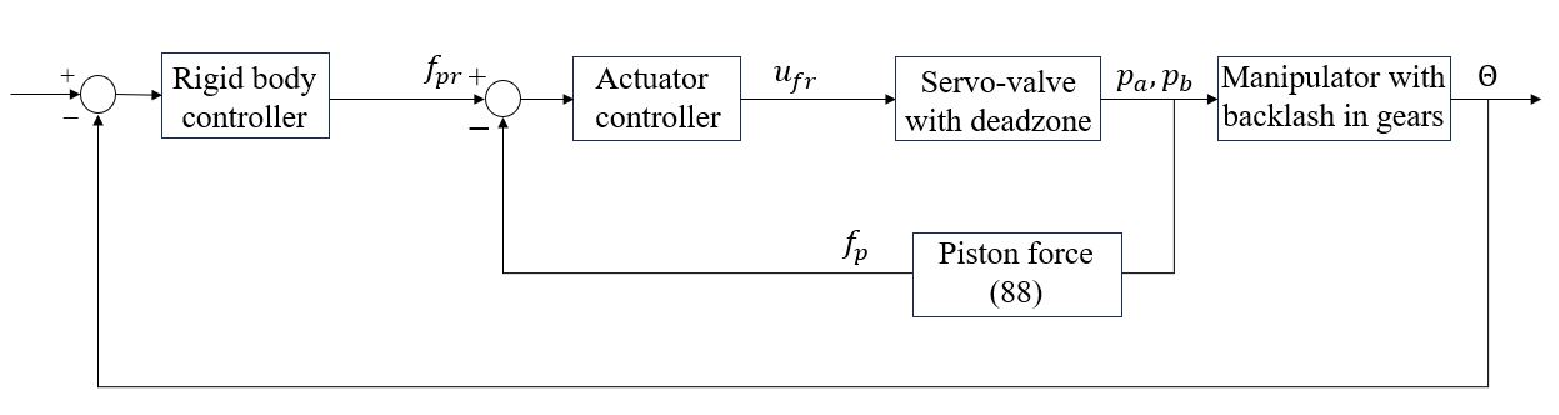}
      \centering
      \label{fIg5a}}
      \hfil
      \subfloat[]{\includegraphics[width = 0.5\textwidth]{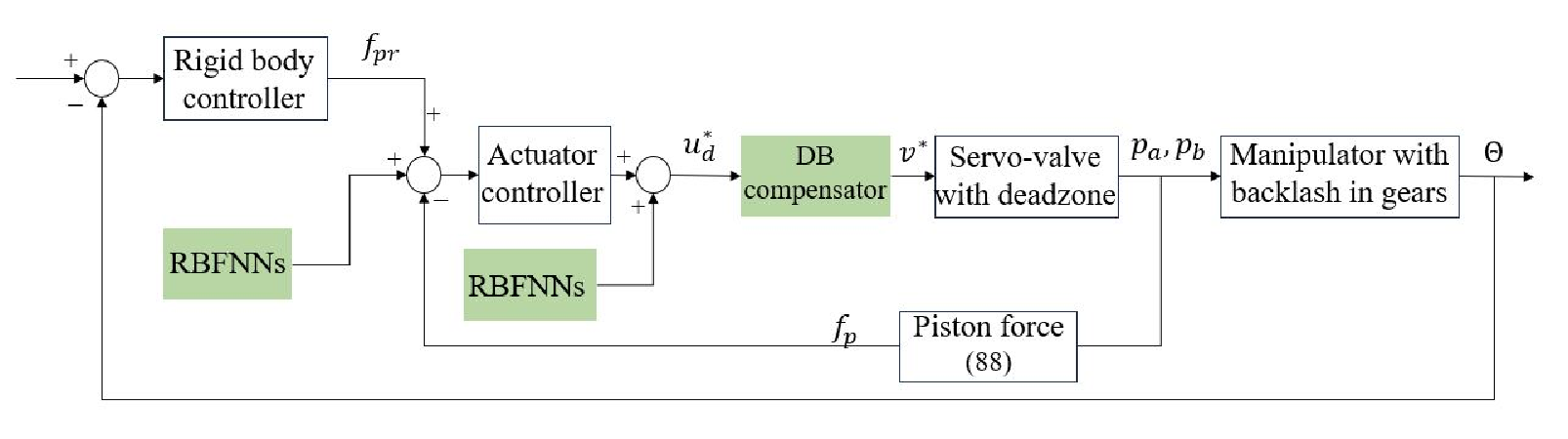}
      \centering
      \label{fIg5b}}
      \caption{ a) The closed-loop scheme of the original VDC in the presence of compound input nonlinearities and unknown model and actuator uncertainties, b) proposed method with deadzone-backlash and uncertainty compensator }
      \label{Fig 5}
   \end{figure}

To guarantee robustness and better performance, the parameter and model uncertainties of the compound input nonlinearities can be tackled by designing an adaptation law for $\hat{\theta}$ and RBFNNs for estimation of $\eta_0$, which will be addressed later. Fig. \ref{Fig 5} displays the closed-loop scheme of the original VDC and the proposed method. As it is shown in Fig. \ref{fIg5b} the deadzone-backlash compensator is incorporated into the controller to assure robustness and better performance in comparison to the original VDC.
\begin{rmk}
    The deadzone-backlash compensation error in (\ref{DB error}) is general for any system that is subjected to such a compound constraint. Therefore, it can be incorporated into any control law at the force/torque and voltage levels.
\end{rmk}

\subsection{Hydraulic actuator dynamics and control}
As mentioned in Assumption \ref{ass: no friction in joints}, only the friction between the piston and the cylinder in the hydraulic actuator is considered, as it significantly impacts actuator performance. In this study, the model proposed in \cite{zhu2005adaptive} is utilized to tackle the friction:
\begin{equation}
    f_f = Y_f\,\phi_f.
    \label{friction lin-in-par}
\end{equation}
The friction model in (\ref{friction lin-in-par}) includes coulomb friction, Stribeck friction, viscous friction, and the average deformation of the seal bristles, as well as a smooth transition between the presliding and sliding motions.

Now, considering the friction force and piston force, we have:
\begin{equation}
    f_p = f_c + f_f.
    \label{fp1}
\end{equation}
On the other hand, the piston force can be computed using the chamber pressure, as:
\begin{equation}
    f_p = A_a\, p_a - A_b\, p_b.
    \label{fp}
\end{equation}
The cylinder chamber pressure can be generated by controlling the fluid flow rate entering the chamber. The equations of these fluid flow rates, denoted as $Q_a$ and $Q_b$ can be written as:
\begin{equation}
Q_{a} = c_{p1} \, \mathcal{Z}(p_s - p_{a}) \, u \, \mathcal{W}(u) + c_{n1} \, \mathcal{Z}(p_{a} - p_{r}) \, u \, \mathcal{W}(-u), 
\label{eqn: Qa}
\end{equation}
\begin{equation}
Q_{b} =  - c_{n2} \, \mathcal{Z}(p_{b} - p_{r}) \, u \, \mathcal{W}(u)-c_{p2} \, \mathcal{Z}(p_s - p_{b}) \, u \, \mathcal{W}(-u),
\label{eqn: Qb}
\end{equation}
where $\mathcal{W}(\mathrm{v})$ is the selection function for the given signal \(\mathrm{v}\), defined as:
\begin{eqnarray}
\mathcal{W}(\mathrm{v}) = \left\lbrace \begin{array}{ll}
	1, & \text{if} \, \quad \mathrm{v} > 0, \\
	0, & \text{if} \, \quad \mathrm{v} \leqslant 0, \\
\end{array} \right.
\end{eqnarray}
and $\mathcal{Z}(\cdot)$ is the function related to drops in pressure:
\begin{eqnarray}
\mathcal{Z}(\cdot) = \text{sign}(\cdot)  \, \sqrt{\left\lvert \cdot \right\rvert}.
\end{eqnarray}
Employing continuity equations for hydraulic actuators, one can express the pressure dynamics in the cylinder chamber as:
\begin{equation}
\dot{p}_{a} = \dfrac{\beta}{V_{0a} +\, A_{a}{x}} \left( Q_{a} - A_{a} \dot{x}-\, Q_l \right),
\label{eqn: dpa}
\end{equation}
and
\begin{equation}
\dot{p}_{b} = \dfrac{\beta}{V_{0b} + \,A_{b}({s-x})} \left( Q_{b} + A_{b} \dot{x}+\, Q_l \right),
\label{eqn: dpb}
\end{equation}
where the laminar leakage flow $Q_l$ between the cylinder chambers is defined as:
\begin{equation}
    Q_l = c_l (p_a-p_b).
    \label{Ql}
\end{equation}
Taking the time derivative of the (\ref{fp}) and recalling (\ref{eqn: Qa}), (\ref{eqn: Qb}), and (\ref{eqn: dpa})-(\ref{Ql}), one can obtain:
\begin{equation}
\begin{split}
    \dot{f}_{p} &= \beta ( u_{f} - \dfrac{A_{a}\dot{x}}{V_{0a}/A_a + x} - \dfrac{A_{b}\dot{x}}{V_{0b}/A_b + (s - x)}\\
    &-\, c_l \dfrac{(pa-p_b)(A_aV_{0b}+A_bV_{0a}+A_aA_bs)}{(V_{0a}+A_Ax)(V_{0b}+A_b(s-x))} + \Delta_a)
\end{split}
\label{eqn: fp with uf}
\end{equation}
with 
\begin{equation}
    u_{f} = -Y_v \theta_v+ \Delta_a
\label{eqn: uf def}
\end{equation}
where $Y_v = [Y_{v1},Y_{v2},Y_{v3},Y_{v4}]$ with $Y_{v1} = -\dfrac{\mathcal{Z}(p_s-p_a)}{V_{0a}/ A_a + x}\mathcal{W}(u)u$, $Y_{v2} = -\dfrac{\mathcal{Z}(p_s-p_b)}{V_{0b}/ A_b + (s-x))}\mathcal{W}(-u)u$, $Y_{v3} = -\dfrac{\mathcal{Z}(p_a-p_r)}{V_{0a}/ A_a + x}\mathcal{W}(-u)u$, and $Y_{v4} = -\dfrac{\mathcal{Z}(p_b-p_r)}{V_{0b}/ A_b + (s-x))}\mathcal{W}(u)u$, and $\theta_v = [c_{p1},c_{p2},c_{n1},c_{n2}]^T $.

\begin{assumption}
Pressures in linear hydraulic actuator chambers are always smaller than the supply pressure, and they are always higher than the return line pressure, which is never zero.
\label{ass: pressures}
\end{assumption}

Assumption \ref{ass: pressures} provides univalence between $u$ and $u_{f}$. It means that for the given $u_f$, a unique spool valve voltage signal $u$ can be obtained as:
\begin{equation}
\begin{split}
    u &= \dfrac{u_f}{c_{p1}\,\dfrac{\mathcal{Z}(p_s-p_a)}{V_{0a}/ A_a + x}+c_{n2}\dfrac{\mathcal{Z}(p_b-p_r)}{V_{0b}/ A_b + (s-x))}}\mathcal{W}(u_f)\, \\
    &+ \dfrac{u_f}{c_{n1}\,\dfrac{\mathcal{Z}(p_a-p_r)}{V_{0a}/ A_a + x}+c_{p2}\dfrac{\mathcal{Z}(p_s-p_b)}{V_{0b}/ A_b + (s-x))}}\mathcal{W}(-u_f).
\end{split}
\label{voltage signal}
\end{equation}
In order to ensure that the required piston force is generated at the actuator level, we need to design the voltage law $u_{fr}$. Considering (\ref{friction lin-in-par}), (\ref{fp1}), (\ref{fp}), and (\ref{eqn: fp with uf}) along with Definition \ref{Lemma: 1}, we have:
\begin{equation}
    f_{pr} = f_{cr} + Y_f\,\hat{\theta}_f,
    \label{fpr}
\end{equation}
\begin{equation}
    u_{fr}^* = Y_d\,\hat{\theta}_d +\, k_f(f_{pr}-f_p)+\, k_x(\Dot{x}_r-\Dot{x})+ \hat{W}_a^T\Psi(\chi_a) +\,\hat{\varepsilon}_a,
    \label{ufr}
\end{equation}
with $\theta_d = [1/\beta, 1,1,c_l]^T $, and $Y_d = [Y_{d1},Y_{d2},Y_{d3},Y_{d4}]$ with $Y_{d1} = \Dot{f}_{pr}$, $Y_{d2} = (A_a\Dot{x})/(V_{0a}/A_a + x)$, $Y_{d3} = (A_b\Dot{x})/(V_{0b}/A_b + (s-x))$, and $Y_{d4} = ((pa-p_b)(A_aV_{0b}+A_bV_{0a}+A_aA_bs))/((V_{0a}+A_Ax)(V_{0b}+A_b(s-x)))$. Additionally, \(\chi_a = [x, \Dot{x}, p_a, p_b]^T \in \mathbb{R}^4\), $\hat{W}_a \in \mathbb{R}^{\Bar{n}_a\times1}$ and $\hat{\varepsilon}_a \in \mathbb{R}$ are the estimation of $W_a $ and $\varepsilon_a = \varepsilon_a^* + \eta_0$ in the sense of Definition \ref{Lemma: 1} to handle the unmodeled dynamics in the hydraulic actuator. $\eta_0$ is defined in (\ref{DB error}) and $\Bar{n}_a$ is the total number of neurons in the actuator subsystem. The signal in (\ref{ufr}) is the required voltage that accomplishes the control goals in the absence of deadzone-backlash, while $v_{fr}$ is equipped with the input nonlinearities compensator in the sense of (\ref{v BDI}) as: 
\begin{equation}
    v_{fr} = \frac{1}{\hat{c}}u_{fr}^*+\Bar{\hat{w}}(t)
    \label{v_{fr}}
\end{equation}
Subsequently, the voltage error can be written in the sense of (\ref{DB error}) as:
\begin{equation}
    u_{fr}-u_{fr}^* = -\Tilde{\theta}^T\,\eta+\eta_0 
    \label{final DB error}
\end{equation}
where $u_{fr} = DB(v_{fr})$. 

Therefore, by recalling (\ref{voltage signal}), the spool valve control signal may written as:
\begin{equation}
\begin{split}
    u &= \dfrac{u_{fr}\mathcal{W}(u_{fr})}{\hat{c}_{p1}\,\dfrac{\mathcal{Z}(p_s-p_a)}{V_{0a}/ A_a + x}+\hat{c}_{n2}\dfrac{\mathcal{Z}(p_b-p_r)}{V_{0b}/ A_b + (s-x))}}\, \\
    &+ \dfrac{u_{fr}\mathcal{W}(-u_{fr})}{\hat{c}_{n1}\,\dfrac{\mathcal{Z}(p_a-p_r)}{V_{0a}/ A_a + x}+\hat{c}_{p2}\dfrac{\mathcal{Z}(p_s-p_b)}{V_{0b}/ A_b + (s-x))}}.
\end{split}
\label{voltage signal req}
\end{equation}
In order to estimate the flow coefficient in (\ref{voltage signal req}), we need to write it in the linear-in-parameter form. Therefore, as in (\ref{eqn: uf def}), we can inversely write (\ref{voltage signal req}) as:
\begin{equation}
    u_{fr} = -Y_v \hat{\theta}_v.
\label{eqn: uf def req}
\end{equation}
The low-level control law designed in (\ref{v_{fr}}) not only compensates for compound input nonlinearities and unknown unmodeled dynamics at the actuator level, but it also makes the actuator produce the required piston forces that tackle unknown model uncertainty and unknown disturbances in rigid body part, resulting in precise trajectory tracking. 

\begin{figure*}
\centering
\includegraphics[width=1\textwidth]{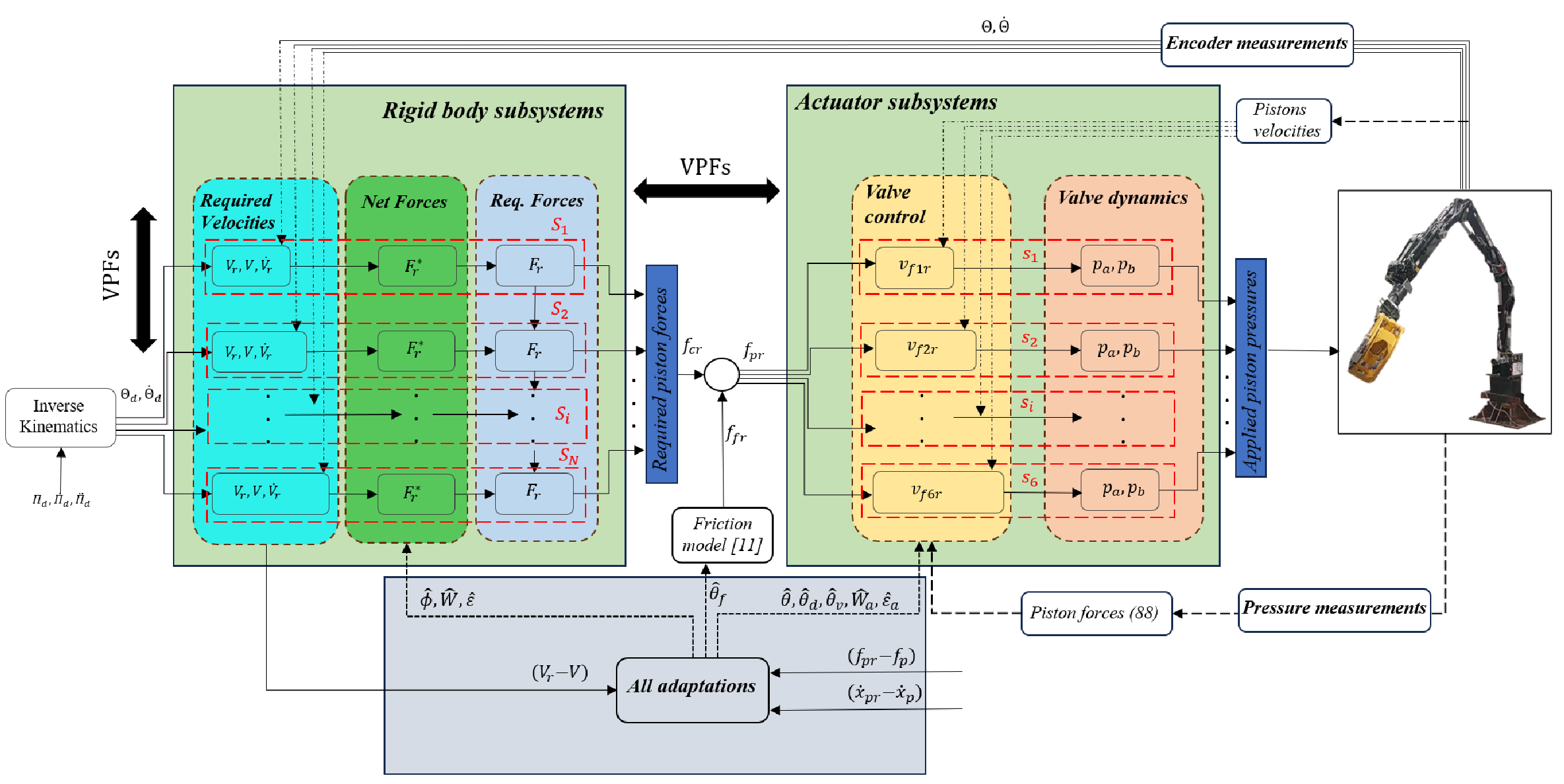}
\caption{General scheme of the orchestrated robust controller. $f_{cr}$ is the required piston force that if applied to the rigid body, that subsystem will have the desired motion. $v_{fr}$ is the required low-level voltage that if applied to the valves, the actuator will produce $f_{cr}$. $S_i$ is subsystems in the rigid body parts with $N$ being the total number of rigid bodies, and $s_i$ is a subsystem in the actuator level which is the same as the number of actuators.} \label{fig: scheme}
\end{figure*}

\section{Stability Analysis}

In the previous section, the modeling and control of the HHM were established. Fig. \ref{fig: scheme} demonstrates the scheme of the proposed approach with details of subsystem-based modeling and control. It is shown that the required force designed in the rigid body subsystem is transmitted to the actuator subsystem to be generated. The connections between rigid body subsystems and actuator subsystems are established by means of VPFs. In this section, the stability of the entire system under the proposed controller is analyzed. The adaptation functions for handling parameter uncertainties in rigid body parts and actuator parts are derived to achieve SGUUB. 

\begin{thm}
    Consider the 6-DoF hydraulic manipulator demonstrated in Fig. \ref{Fig Hiab objects} that decomposed into rigid body and actuator subsystems in Fig. \ref{fig: object 1}-\ref{fig: object 3}. The rigid body dynamics (\ref{eqn: tot force 2}) in the presence of unknown model uncertainty and disturbance under the robust local controller (\ref{Fr des final}) with the adaptation laws of:
    \begin{equation}
        ^{ A}\Dot{\hat{\mathcal{L}}} = \frac{1}{\gamma}\, ^{ A}\hat{\mathcal{L}}\,\left(^{ A}\mathcal{I}-\gamma_0\,^{ A}\hat{\mathcal{L}}\right)\, ^{ A}\hat{\mathcal{L}}
        \label{L adapt}
    \end{equation}
    \begin{equation}
        ^{ A}\Dot{\hat{W}} = {^{ A}\Gamma}\,\left(\Psi(\chi_{ A})\,(^{ A}{ V}_r-\,^{ A}{ V})^T- ^{ A}\tau_0\,{^{ A}\hat{W}}\right)
        \label{W adapt}
    \end{equation}
    \begin{equation}
        {^{ A}}\Dot{\hat{\varepsilon}} = {^{ A}\pi} \left (({^{ A}{ V}_r} - {^{ A}{ V}})- ^{ A}\pi_0 {^{ A}}\hat{\varepsilon}\right),
        \label{eps adapt}
    \end{equation}
is virtually stable in the sense of Definition 4, where $ A \in \Upsilon$ and $  \Upsilon = \left\lbrace  P_1,  P_{p2},  B_{0j},  B_{1j},  B_{3j},  B_{4j}, G_{i}\right\rbrace$ for $i = 1,2,3$ and $j = 1,2$. Moreover, $^{ A}\mathcal{I}$ is a unique symmetric matrix, defined Appendix \ref{Apendix D}. Let the non-negative accompanying function for the rigid body part be chosen in the sense of Definition \ref{Def: 4} as ${\nu}_1$. Then, the time derivative of ${\nu}_1$ is:
\begin{equation}
    \Dot{{\nu}}_1 \leq -\alpha_1 {\nu}_1 + \alpha_{10}+VPFs
\end{equation}
with $\alpha_1$ and $\alpha_{10}$ being positive, and VPFs denoting the sum of driving and driven cutting points in the sense of Definition \ref{Def: VCP} and Definition \ref{Def: VPF}.
\label{thm: rigid body}
\end{thm}
\begin{pf}
    The proof is provided in Appendix A.
\end{pf}

\begin{thm}
    Consider the 6-DoF hydraulic manipulator demonstrated in Fig. \ref{Fig Hiab objects} that decomposed into rigid body and actuator subsystems in Fig. \ref{fig: object 1}-\ref{fig: object 3}. The actuator dynamics under the required piston forces (\ref{wrist piston force req}), (\ref{eqn: piston force req}), and (\ref{base piston force req}) with low-level voltage control signal (\ref{ufr}) and (\ref{v_{fr}}) along with following adaptation functions:
    \begin{equation}
        \Dot{\hat{\theta}}_{f(.)} =  \gamma_{f(.)}\left(\frac{1}{k_{x(.)}}Y_{f(.)}^T(\Dot{x}_{p(.)r}-\Dot{x}_{p(.)})-\gamma_{f(.)0}\hat{\theta}_{f(.)}\right)
        \label{hat theta f}
    \end{equation}
    \begin{equation}
        \Dot{\hat{\theta}}_{v(.)} = \gamma_{v(.)}\left(\frac{1}{k_{x(.)}}Y_{v(.)}^T(f_{p(.)r} - f_{p(.)})-\gamma_{v(.)0}\hat{\theta}_{v(.)}\right)
        \label{hat theta v}
    \end{equation}
    \begin{equation}
        \Dot{\hat{\theta}}_{d(.)} = \gamma_{d(.)}\left(\frac{1}{k_{x(.)}}Y_{d(.)}^T(f_{p(.)r} - f_{p(.)})-\gamma_{d(.)0}\hat{\theta}_{d(.)}\right)
        \label{hat theta d}
    \end{equation}
    \begin{equation}
        \Dot{\hat{\theta}}_{(.)} =  \delta_{(.)}\left(\frac{1}{k_{x(.)}}(f_{p(.)r} - f_{p(.)})-\delta_{(.)0}\hat{\theta}_{(.)}\right)
        \label{hat theta }
    \end{equation}
    \begin{equation}
        \Dot{\hat{W}}_{a(.)} = \delta_{a(.)}\left(\frac{1}{k_{x(.)}}(f_{p(.)r} - f_{p(.)})\Psi(\chi_{a(.)})-\delta_{a(.)0}\hat{W}_{a(.)}\right)
        \label{hat Wa v}
    \end{equation}
    \begin{equation}
        \Dot{\hat{\varepsilon}}_{a(.)} = \Bar{\delta}_{a(.)}\left(\frac{1}{k_{x(.)}}(f_{p(.)r} - f_{p(.)})-\Bar{\delta}_{a(.)0}\hat{\varepsilon}_{a(.)}\right)
        \label{hat vareps d}
    \end{equation}
    is virtually stable in the sense of Definition 4, with $(.)$ denoting all the objects. Defining the non-negative accompanying functions for linear hydraulic actuators as $\nu_{a}$ in the sense of Definition \ref{Def: 4} yields to:
    \begin{equation}
        \Dot{\nu}_{a} \leq -\mu_{a}\nu_{a}+\mu_{a0}+VPFs
    \end{equation}
    with $\mu_{a}$ and $\mu_{a0}$ being positive, and VPFs being the virtual power flow at the actuator level.
    \label{thm: actuator}
\end{thm}
\begin{pf}
    The proof is provided in Appendix B.
\end{pf}

\begin{thm}
    Consider the generic 6-DoF HHM with rigid body and actuator dynamics represented in (\ref{eqn: tot force 2})-(\ref{base piston force}) and (\ref{friction lin-in-par})-(\ref{voltage signal}) with control laws of (\ref{Fr des final}), (\ref{wrist piston force req}), (\ref{eqn: piston force req}), (\ref{base piston force req}), (\ref{ufr}), and (\ref{v_{fr}}) and adaptation functions in (\ref{L adapt})-(\ref{eps adapt}) and (\ref{hat theta f})-(\ref{hat vareps d}). All the signals and errors in the system are SGUUB and will remain within a compact set. For example, velocity tracking error, actuator force tracking error, and RBFNNs weight estimation error will stay within a compact set $\Omega_V$, $\Omega_f$, and $\Omega_W$, respectively, where $\Omega_V = \lbrace e_V \in \mathbb{R}^6|\, ||e_v||\leq \sqrt{\frac{2B}{\lambda_{min}(M)}} \rbrace$, $\Omega_f = \lbrace e_f \in \mathbb{R}|\, ||e_f||\leq \sqrt{2B \beta k_{x}} \rbrace$, and $\Omega_W = \lbrace \Tilde{W}_a \in \mathbb{R}^{n}|\, ||\Tilde{W}_a||\leq \sqrt{2B \delta_{x}} \rbrace$ with $B = \Bar{\mu}_0/\Bar{\mu}$. Similarly, for all the errors, a compact set can be achieved.
    \label{thm: total}
\end{thm}
\begin{pf}
    The proof is provided in Appendix C.
\end{pf}
\begin{rmk}
    The adaptation laws in (\ref{L adapt})-(\ref{eps adapt}) and (\ref{hat theta f})-(\ref{hat vareps d}) are structured to achieve SGUUB. This ensures not only that the system remains bounded and robust but also that the control terms converge reliably under various uncertainties.
\end{rmk}

\section{Simulation Results}

In this section, the simulation results are provided to evaluate the performance of the proposed controller in comparison to other methods. Three different controllers are employed as follow; original VDC, proportional-differential (PD) controller, and the state-of-the-art controller, adaptive dynamic surface controller with funnel control and neural networks (ADSC-FC-NN) proposed in \cite{yang2022neural}. The simulation analysis is performed for the base joint of the HHM, shown in Fig. \ref{Fig Hiab objects}, which is subjected to deadzone in valves and backlash in the rack and pinion mechanism.

\begin{figure}[h]
      \centering
      \subfloat[]{\includegraphics[width = 0.45\textwidth]{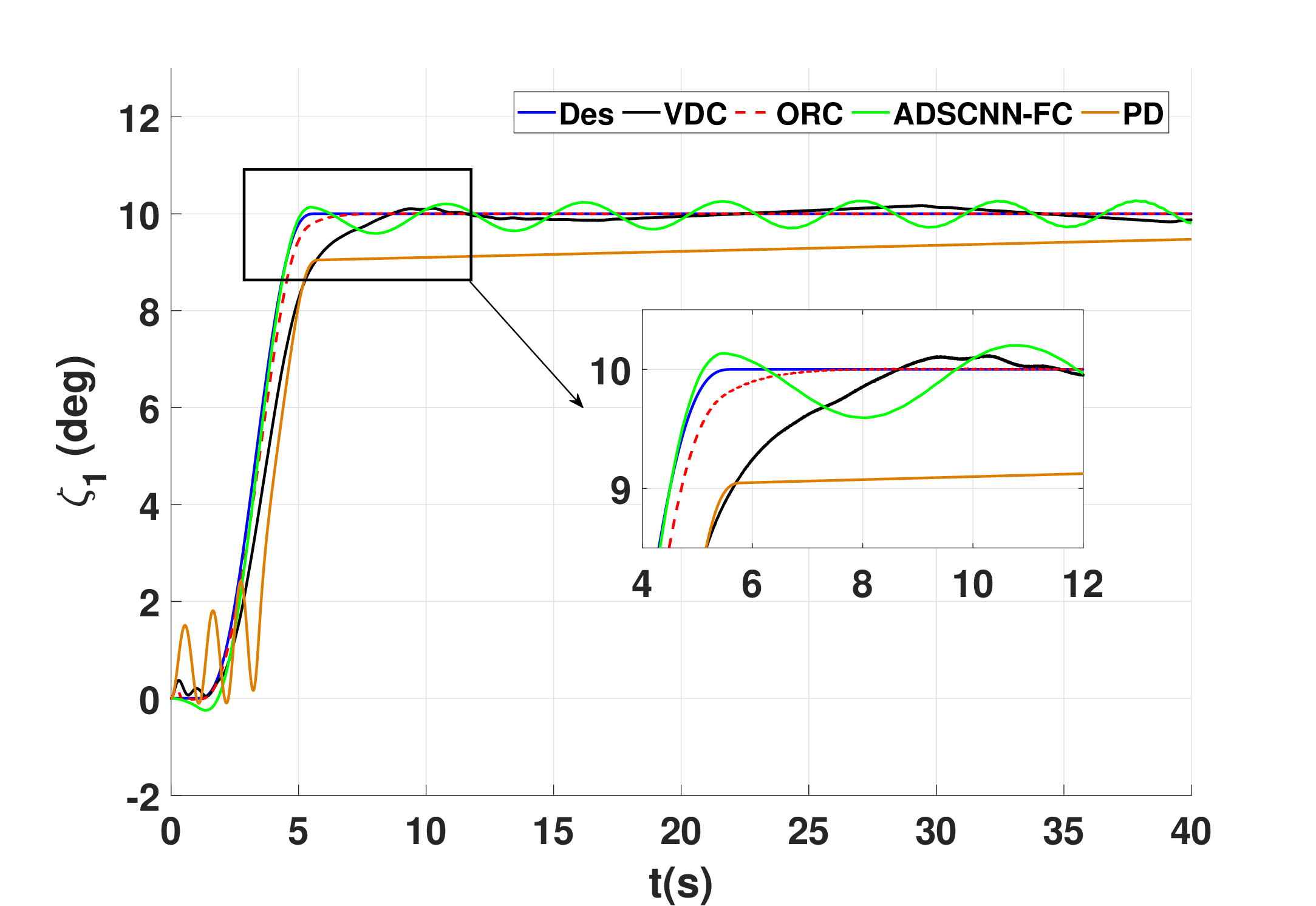}
      \centering
      \label{comp 1}}
      \hfil
      \subfloat[]{\includegraphics[width = 0.45\textwidth]{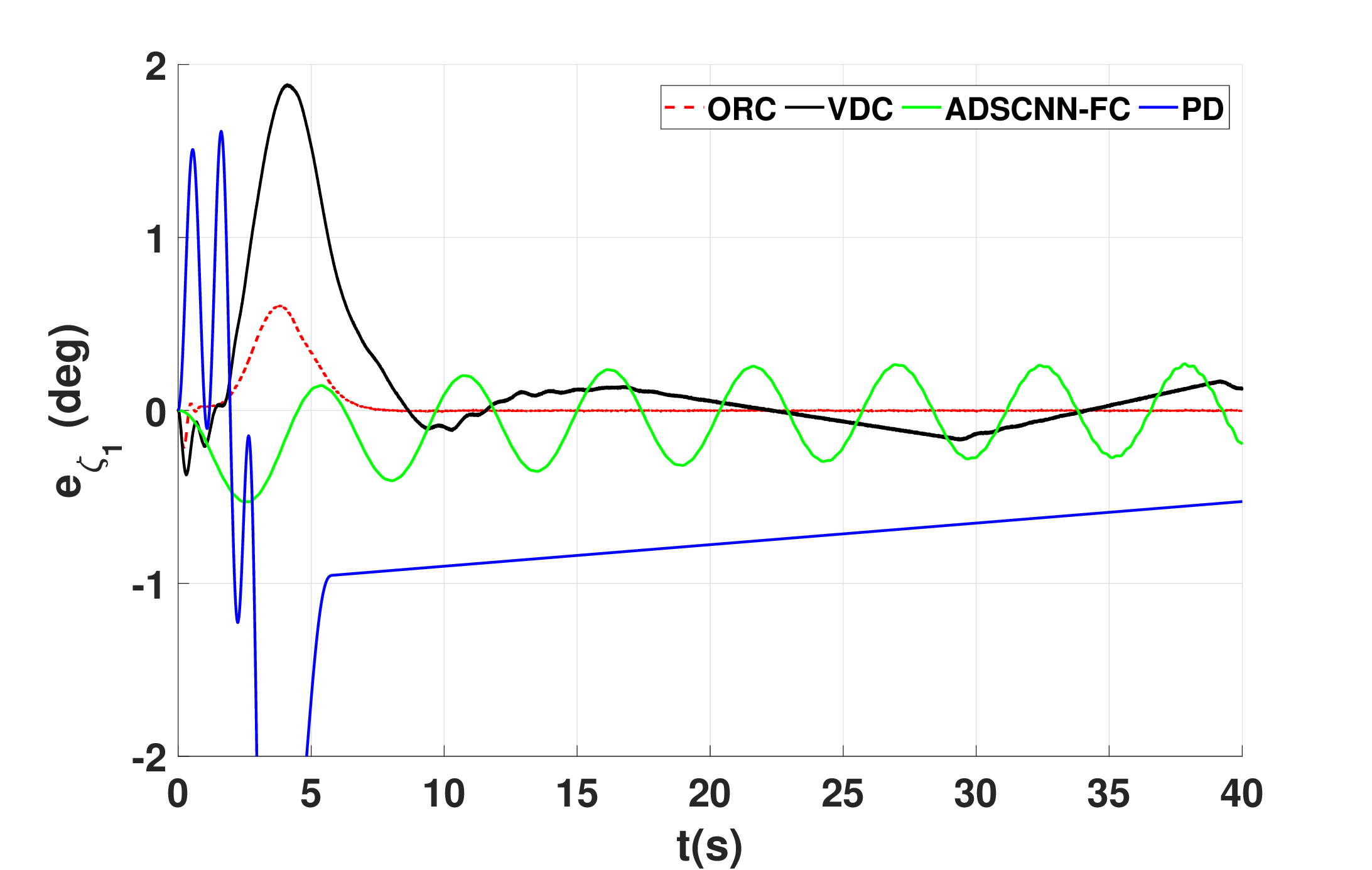}
      \centering
      \label{comp 2}}
      \caption{ a) The simulation result of tracking of a desired trajectory with four different controllers for the base joint, b) time history of the corresponding errors.}
      \label{Controller comparison}
   \end{figure}

In the simulation, which is performed by MATLAB \& SIMULINK, the deadzone and backlash constraints are considered to evaluate the performance of each controller in the presence of such compound constraints, as well as model uncertainties. The deadzone and backlash parameters are set to $b_r = -b_l = 0.2$, $m_d = 1$, $B_r = -B_l = 0.2$, $k_b = 1$. All the control gains are tuned to achieve the best results for each controller, and the 10 nodes in total are utilized for RBFNNs in the proposed method and ADSC-FC-NN. Fig. \ref{Controller comparison} demonstrates the result of the simulation. The desired trajectory in Fig. \ref{comp 1} helps to analyze both the transient and steady-state performances of each controller. As displayed in the figure, the ADSC-FC-NN controller has much better transient performance than the original VDC and PD. However, both the VDC and ADSC-FC-NN could not converge to the desired point, and both oscillated around the set point, with errors of 0.17 and 0.25 degrees, respectively, due to the compound input nonlinearities. This shows that the RBFNN in ADSC-FC-NN could not perfectly estimate the uncertainty in the actuator. In contrast, the proposed method not only improved the weakness of the original VDC in the transient part, but it also ultimately converged to the desired point and perfectly tackled the effect of input nonlinearities. Fig. \ref{comp 2} compares the errors of each controller. It can be concluded that using novel DRBFNNs for the estimation of actuator uncertainties, which estimation of the non-parametrizable term in (\ref{final DB error}) is part of it, based on the states of the actuator, perfectly accomplished the control objective. Table \ref{Compare table} additionally provides a quantitative performance evaluation of each controller by comparing the maximum-absolute of tracking error $|e_{max}|$, root-mean-square (RMS) error (RMSE) $e_{rms}$, and RMS of computed voltage $u_{rms}$. According to Table \ref{Compare table}, the maximum error value of the ADSC-FC-NN is better than the proposed method, while the designed controller has a much lower RMS error and almost zero steady-state error with lower voltage usage.

\begin{table}[h]
\centering
\caption{Simulation performance evaluation}
\begin{tabular}{P{30pt} P{15pt} P{50pt} P{20pt} P{20pt}}
\hline
Controller& 
PD& 
ADSC-NN-FC \cite{yang2022neural}& 
VDC&
ORC
\\ [0.5ex]
\hline \hline
$|e|_{max} (^{\circ}) $& 
4.57& 
$0.53  $&
1.88&
0.6\\
\hline
$e_{rms} (^{\circ}) $&
1.04&
$0.22$&
0.46&
0.12\\
\hline
$u_{rms} (v)$&
0.15&
$0.74 $&
0.34&
0.44\\
\hline
\end{tabular}
\label{Compare table}
\end{table}

\section{Experimental Results}
\begin{figure}[t]
\centering
\includegraphics[width=.5\textwidth]{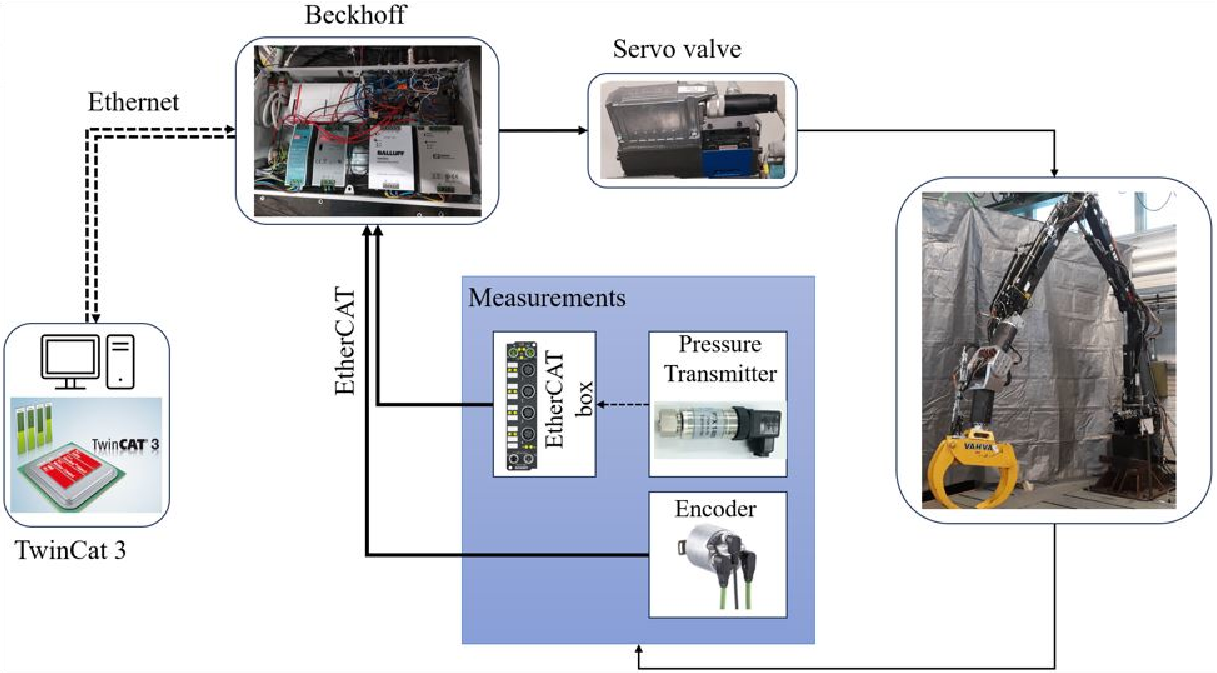}
\caption{Experimental set-up} \label{fig: set up}
\end{figure}

In this section, the performance of the proposed controller is evaluated in real-world scenarios. Fig. \ref{fig: set up} demonstrates the experimental set-up for performance evaluation of the controller. In the following, the hardware components and properties are provided:
\begin{itemize}
    \item Beckhoff and TwinCat 3 interface with a sample time of 1 ms
    \item Bosch Rexroth NG6 size servo solenoid valve with 12 l/min at $\Delta P = 3.5 MPa$ per notch for object 1, 100 l/min at $\Delta P = 3.5 MPa$ per notch for object 2, and 40 l/min at $\Delta P = 3.5 MPa$ per notch for object 3.
    \item Sick afS60 (18-bit) absolute encoders for all joint angle measurements.
    \item Eckart E3150-360 for first and third RHAs in object 3 and Eckart E3150-180 for second RHA ( see Fig. \ref{fig: object 3}).
    \item Druck PTX1400 and Unik 5000 pressure transmitters (range 25 MPa) for pressure measurements.
    \item EP3174-0002 EtherCAT box for connecting pressure sensors to Beckhoff.
\end{itemize}
In order to implement the designed controller, only encoder and pressure sensor data are utilized. Using the pressure sensor to compute the piston force in (\ref{fp}) for automation of heavy-duty manipulators is a more practical solution since retrofitting the pressure sensors is much easier than retrofitting the piston load cell. In order to perform numerical differentiation for $\frac{d}{dt}({^A}V_r)$ in (\ref{Fr des final}) and $\Dot{f}_{pr}$ and $\Dot{x}_{p}$ in (\ref{ufr}), the finite difference method introduced in \cite{harrison1995generalized} is used with a low-pass filter. A smooth fifth-order trajectory generator \cite{reza2010theory} is utilized to produce a smooth trajectory between the set points for a given execution time, $t_f$. The (\ref{TS pose vel}) and (\ref{JS ang vel}) are used to compute the joint's desired values. 

{To ensure safe and effective gain tuning for the 6-DoF manipulator, we employed a manual approach guided by safety precautions. Initially, the manipulator was moved to a specific position using joystick commands, and the controller was activated with low gains to verify feedforward term functionality and system stability. Gains were then incrementally increased to improve positional accuracy while maintaining stability. Finally, during trajectory execution, the gains were fine-tuned in real-time to achieve satisfactory performance without compromising the safety of the system.}

\begin{figure}[h]
      \centering
      \subfloat[]{\includegraphics[width = 0.2\textwidth]{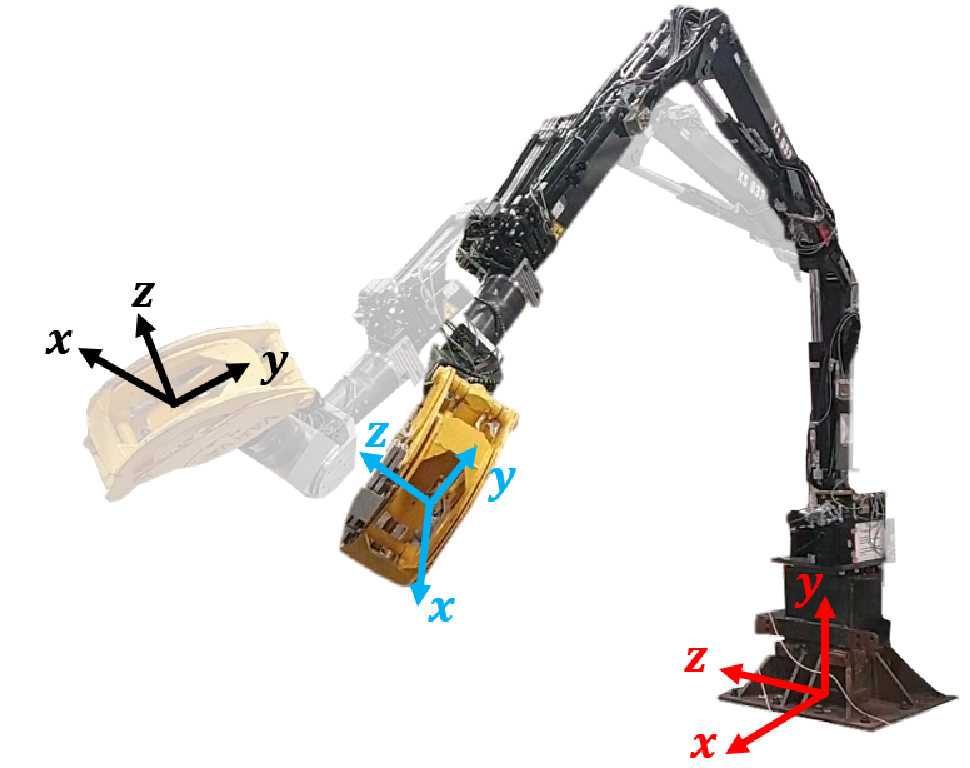}
      \centering
      \label{H1}}
      \hfil
      \subfloat[]{\includegraphics[width = 0.2\textwidth]{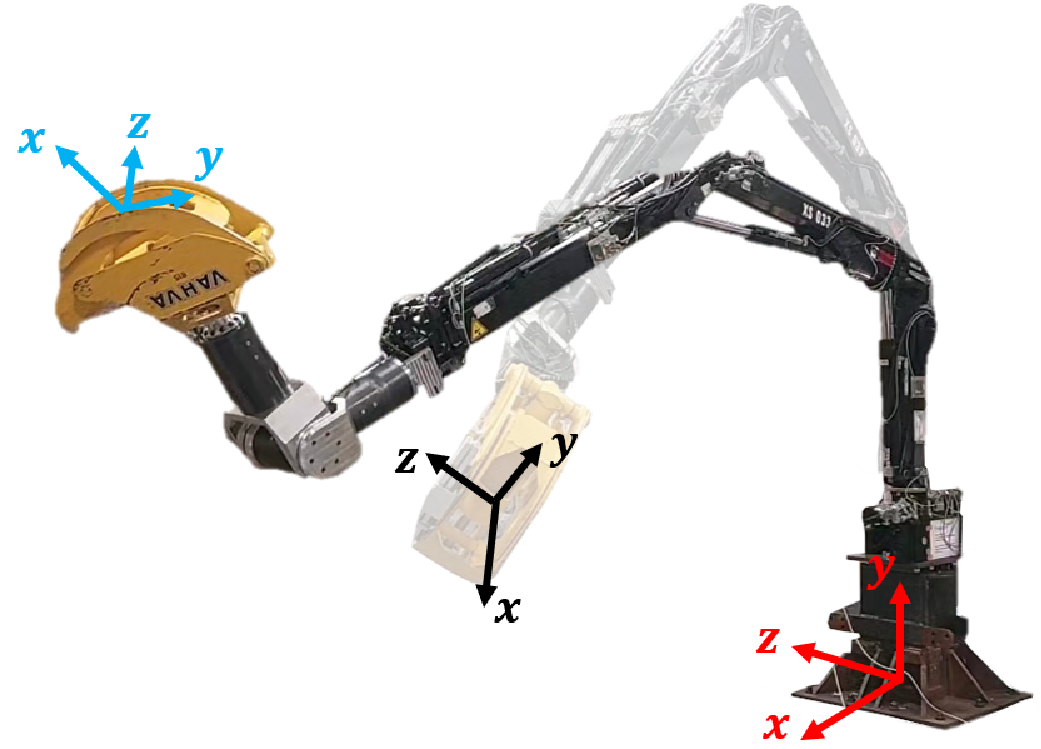}
      \centering
      \label{H2}}
      \subfloat[]{\includegraphics[width = 0.2\textwidth]{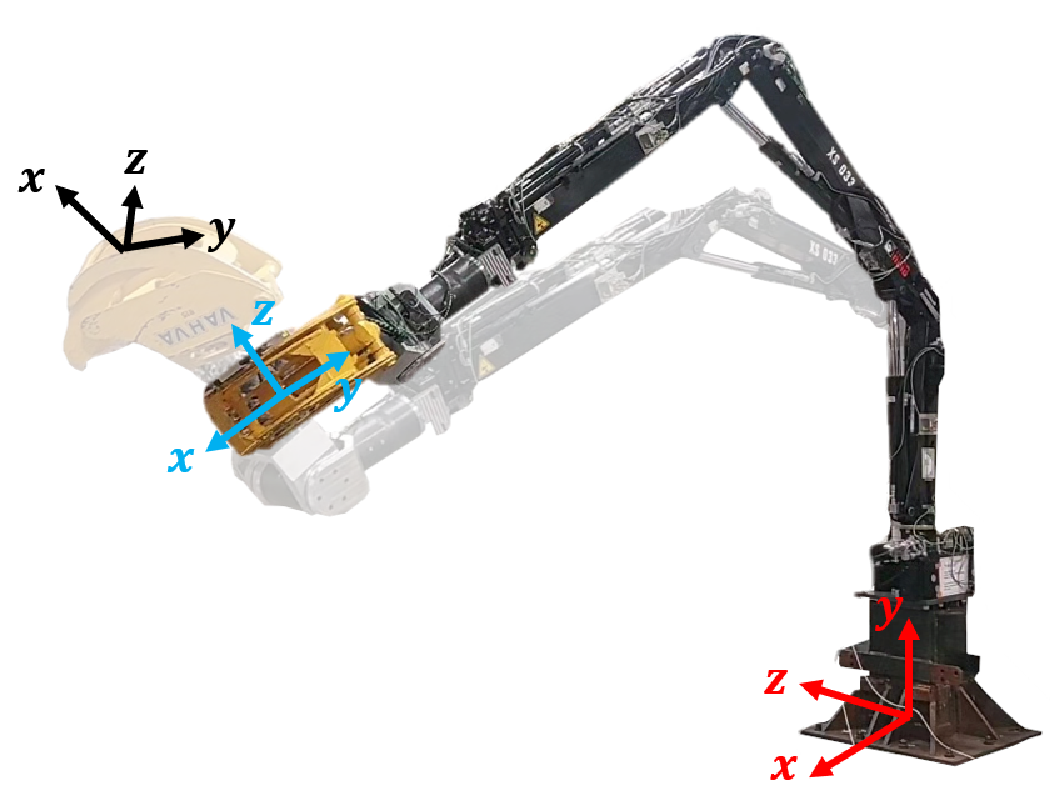}
      \centering
      \label{H3}}
      \hfil
      \subfloat[]{\includegraphics[width = 0.2\textwidth]{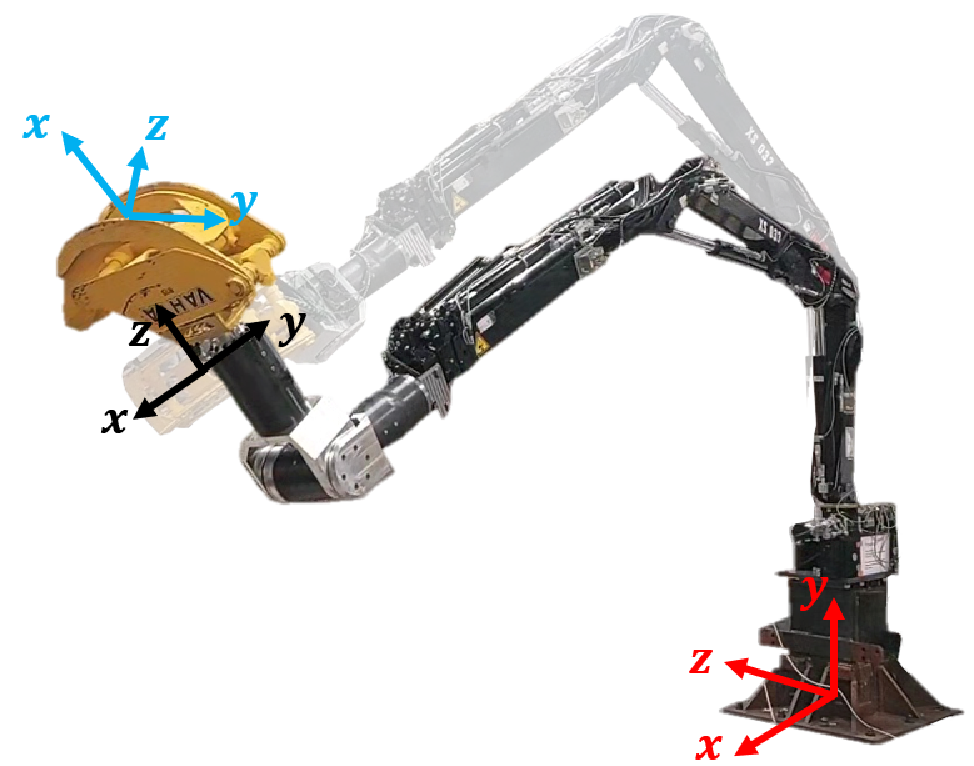}
      \centering
      \label{H4}}
      \hfil
      \subfloat[]{\includegraphics[width = 0.2\textwidth]{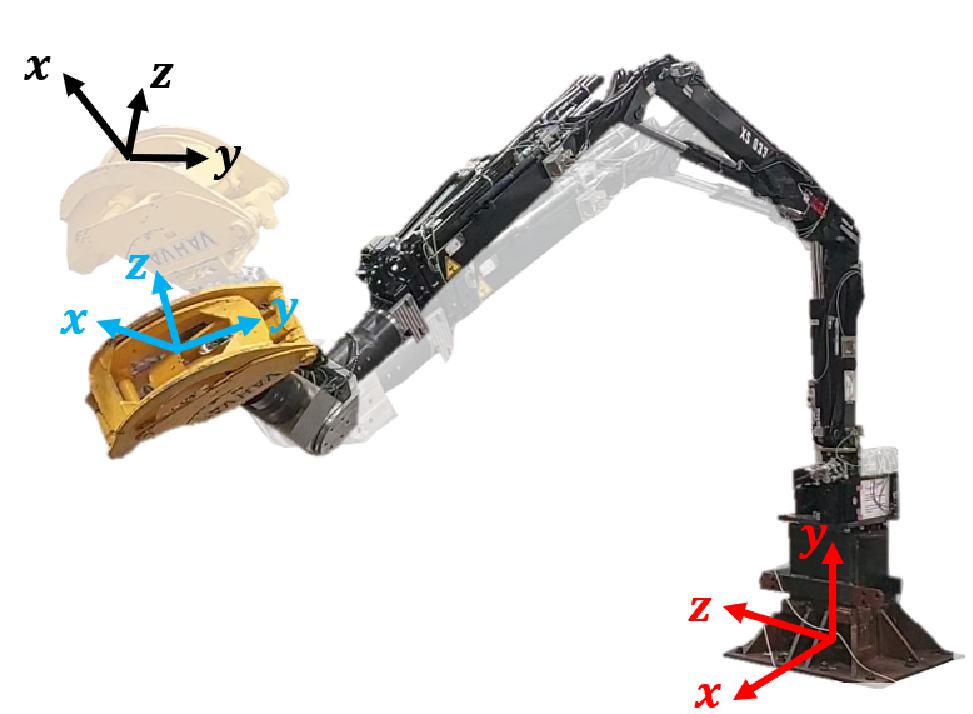}
      \centering
      \label{H5}}
      \caption{ Set-point reaching postures of the robot. Each image shows the transition from one set point to another. The end-effector is moving in the range of 4-5 meters reach.}
      \label{Motion Image}
   \end{figure}

\begin{table}[b]
\centering
\caption{Control parameters for experiment}
\label{table params}
\setlength{\tabcolsep}{10pt}
\begin{tabular}{ P{15pt} P{50pt} P{15pt} P{50pt} }
\hline
Symbol&  Value&  Symbol&  Value \\ [0.5ex]
\hline \hline
$\lambda $& 
3& 
$\Bar{\lambda}_{xj}  $&
5\\
\hline
$\sigma_{1,2} $&
12&
$\sigma_3 $&
18\\
\hline
$k_{xp} $&
0.02&
$k_{xj} $&
0.015\\
\hline
$k_{xwi} $&
0.08&
$k_{fp} $&
$0.01 \cdot 10^{-8}$\\
\hline
$k_{fj} $&
$2 \cdot 10^{-8}$ &
$k_{fwi} $&
$1 \cdot 10^{-8}$\\
\hline
$K_{A} $&
$50 \cdot \boldsymbol{I}$ &
$\gamma $&
$500$\\
\hline
${^A}\Gamma $&
$350$ &
${^A}\pi $&
$20$\\
\hline
$\delta_{a(.)} $&
$1.5 \cdot k_{x(.)}k_{f(.)}$ &
$\Bar{\delta}_{a(.)} $&
$0.5 \cdot k_{x(.)}k_{f(.)}$\\
\hline
\end{tabular}
\label{tab}
\end{table}

\subsection{Motion Performance Analysis}
In this section, the experimental results of implementing the proposed controller on an industrial 6-DoF HHM are provided. The control goal is to achieve five given set points in Cartesian space with the largest magnitude of $0.8\, m$, $1\, m$, and $1\, m$ in $x$, $y$, and $z$ directions, respectively, in different orientations. The Euler angles with the $XYZ$ convention are used to represent the orientation of the end-effector with respect to the base, and the orientation error is computed based on the rotation matrix. Fig. \ref{Motion Image} demonstrates the motion of the robot. 

The result of the proposed method is compared to that of the original VDC controller to evaluate the control performance more accurately. Table \ref{table params} provides the control gains utilized in experiments. For the RBFNNs, the Gaussian activation function is defined as $\Psi(\chi) = exp([-(\chi-c_{\mathbf{j}})^T(\chi-c_{\mathbf{j}})/(b_{\mathbf{j}}^2)])$, with \(c_{\mathbf{j}}\) and \(b_{\mathbf{j}}\) denoting the center and width of the neural cell in \(\mathbf{j}\)th unit, with $\mathbf{j} = \Bar{n}_A$ for rigid body subsystem, and $\mathbf{j} = \Bar{n}_a$ for actuator subsystem. In this study the value for \(c_{\mathbf{j}}\) is randomly selected in \([-1,1]\times[-1,1]\times[-1,1]\) and \([-1,1]\times[-1,1]\times[-1,1]\times[-1,1]\) for \(\chi_A\) and \(\chi_a\), respectively,  with \(b_{\Bar{n}_a} = 0.5\) and \(b_{\Bar{n}_A} = 5\) with the aggregation of 270 nodes. Additionally, prior to inputting the vectors into RBFNN, we normalize all input features to the range of \([-1,1]\). This step ensures that all input dimensions are treated uniformly and that the RBFNNs can effectively cover the input space without the risk of incompatibility.

\begin{figure}[h]
\centering
\includegraphics[width=.5\textwidth]{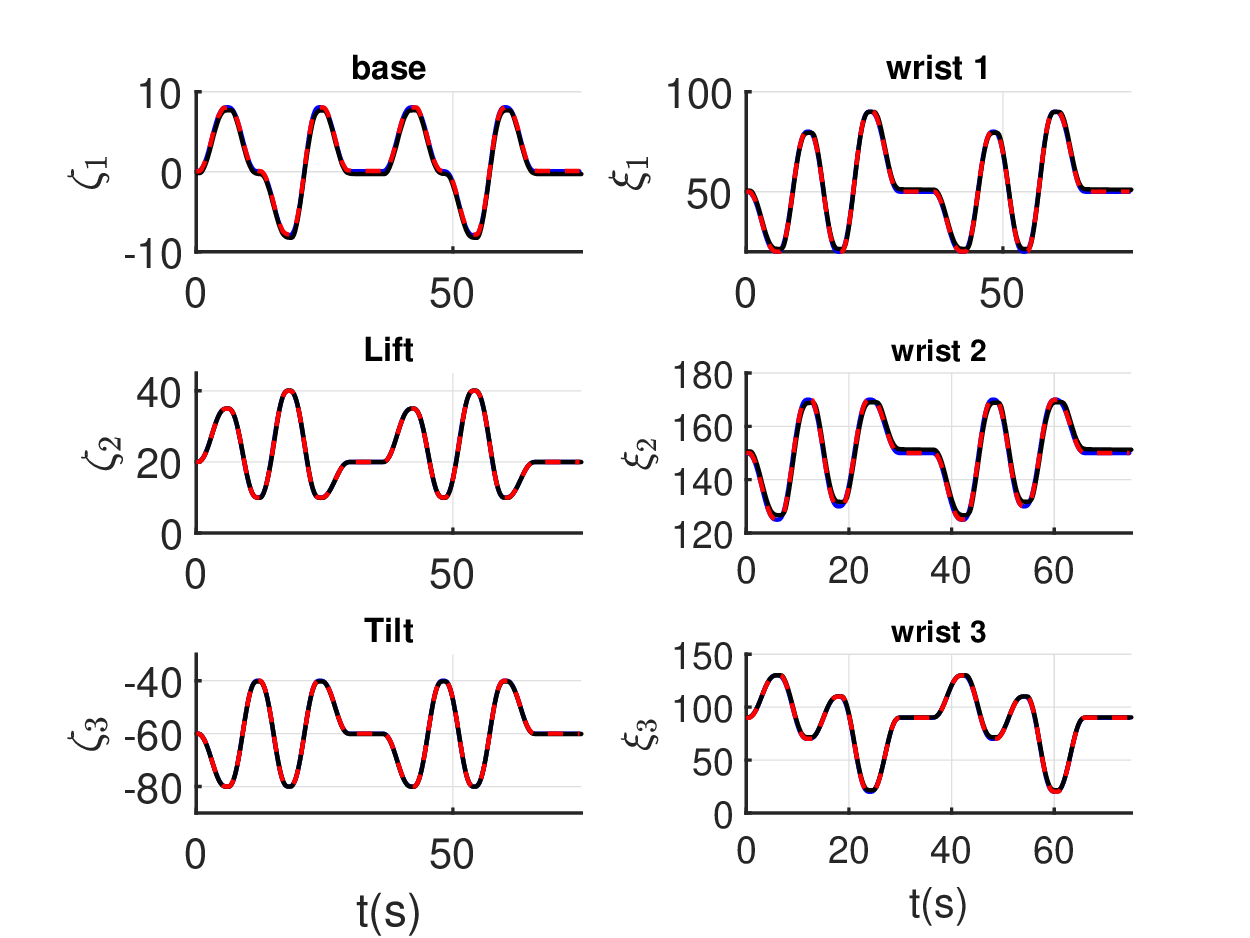}
\caption{Trajectory tracking of all six joints. The blue solid line is the desired trajectory, the dashed red line is with ORC, and the black solid line is with VDC. The y-axis is in degrees.} \label{fig: full tracking}
\end{figure}

\begin{figure}[h]
\centering
\includegraphics[width=.5\textwidth]{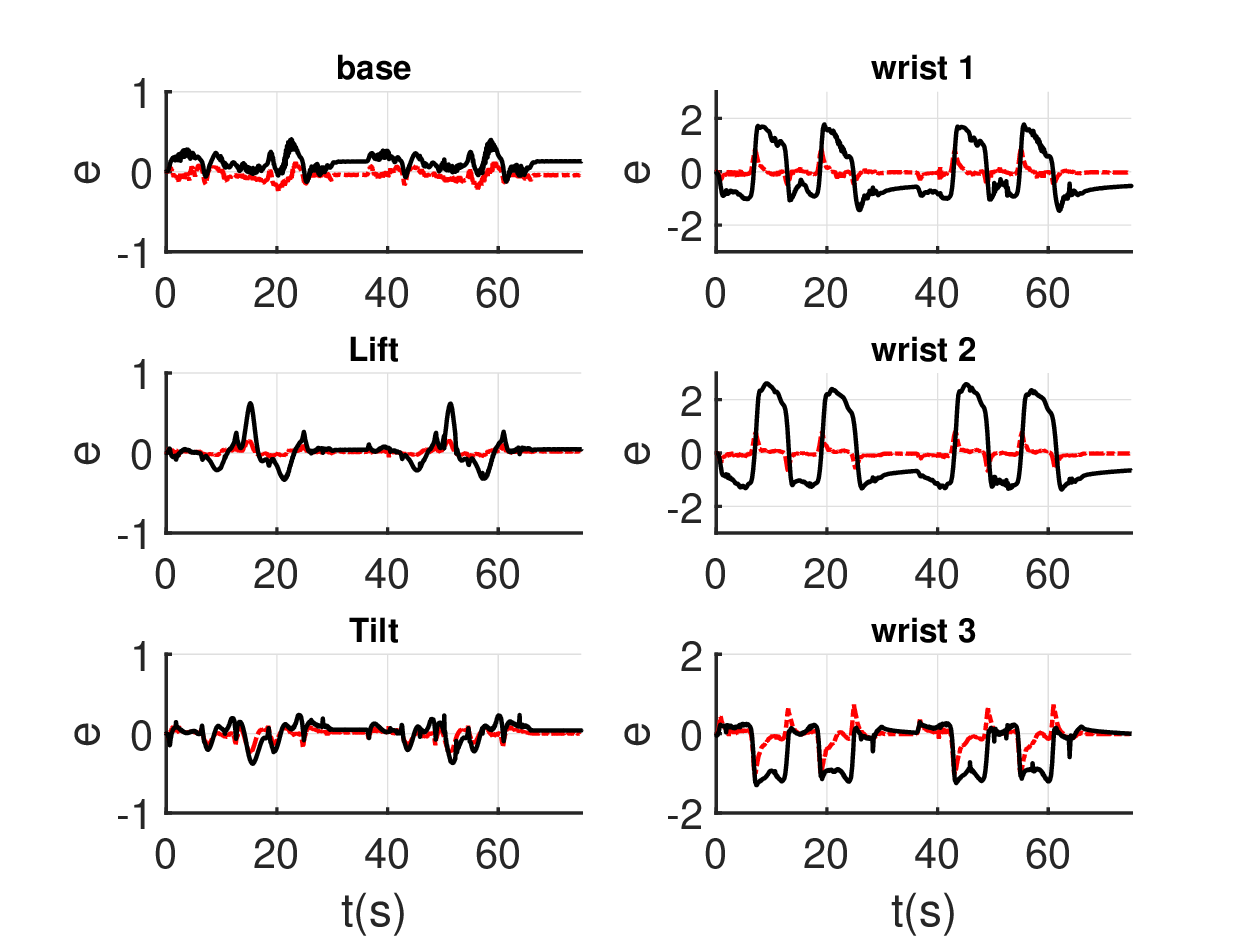}
\caption{Trajectory tracking error of each joint. The dashed red line is with ORC, and the black solid line is with VDC. The y-axis is in degrees.} \label{fig: full tracking error}
\end{figure}

\begin{figure}[b]
\centering
\includegraphics[width=.5\textwidth]{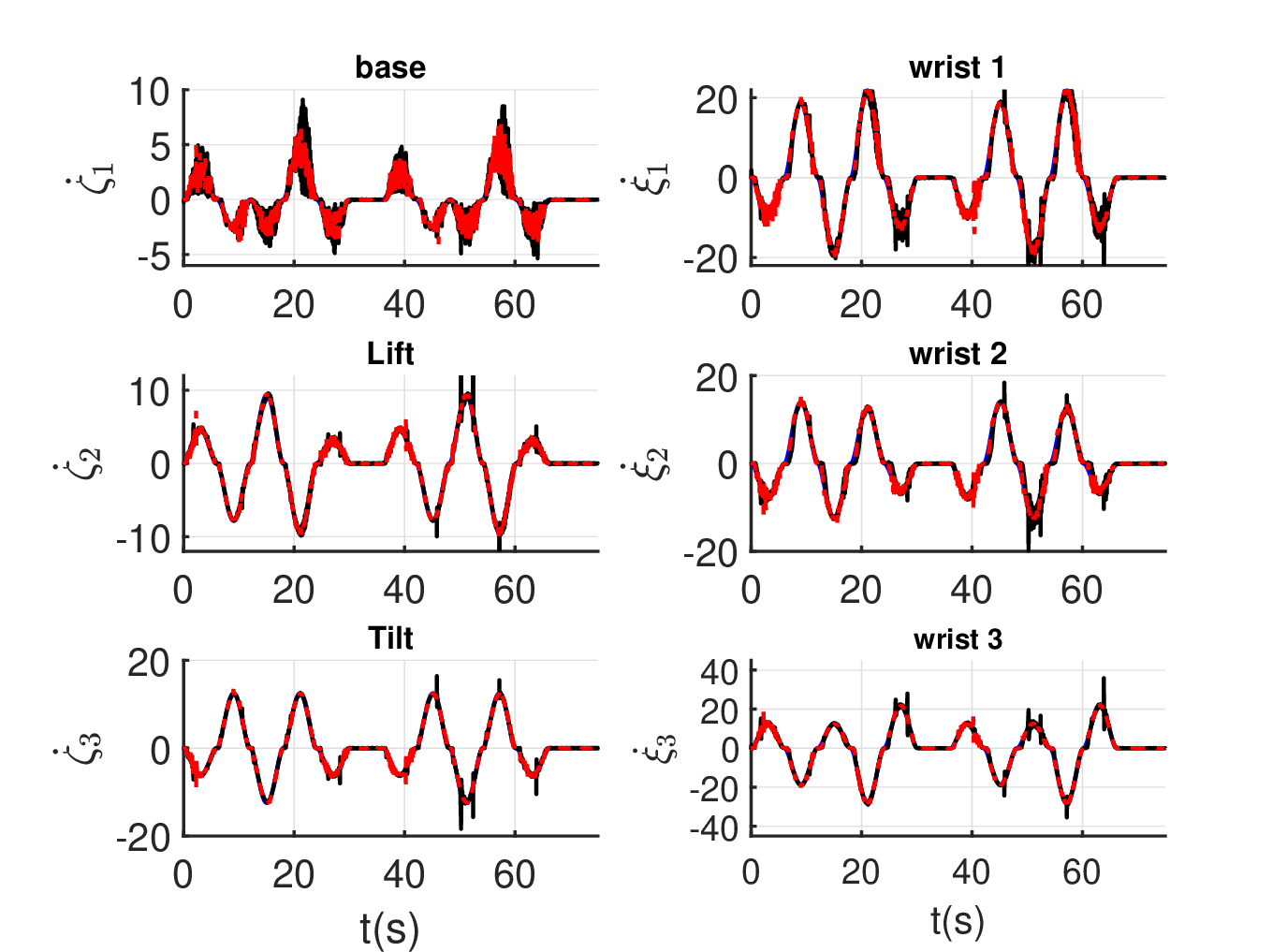}
\caption{Angular velocities of all joints. The blue solid line is the desired trajectory, the dashed red line is with ORC, and the black solid line is with VDC. The y-axis is in degrees/sec.} \label{fig: full angular velocity}
\end{figure}

In Fig. \ref{fig: full tracking}, the performance of the proposed method and the original VDC in tracking a desired trajectory is compared. As demonstrated, both controllers performed excellently in following the desired trajectory. However, from Fig. \ref{fig: full tracking error}, it can be concluded that the proposed controller achieved a much better result by handling the uncertainties in the model and considering the input nonlinearities. As mentioned previously, the base and wrist actuators (objects 1 and 3) suffer from compound input nonlinearities. Such nonliniearities along with the model uncertainties lead to the steady-state error of 0.13, -0.6, -0.7, and 0.07 degree in $\zeta_1$, $\xi_1$, $\xi_2$, and $\xi_3$, respectively, using VDC controller. By employing the presented controller, the errors are reduced to -0.04, -0.01, -0.02, and -0.01 degrees, respectively. In addition to the steady state errors, the proposed method achieved a considerably lower transient error, showing the perfect performance of the designed method in tackling unknown uncertainties and compound input nonlinearities. Fig. \ref{fig: full angular velocity}, on the other hand, depicts the angular velocity of each joint along with the desired values, displaying that the ORC ended up with much smoother tracking without jumps in velocities. 

\begin{figure}[t]
\centering
\includegraphics[width=.52\textwidth]{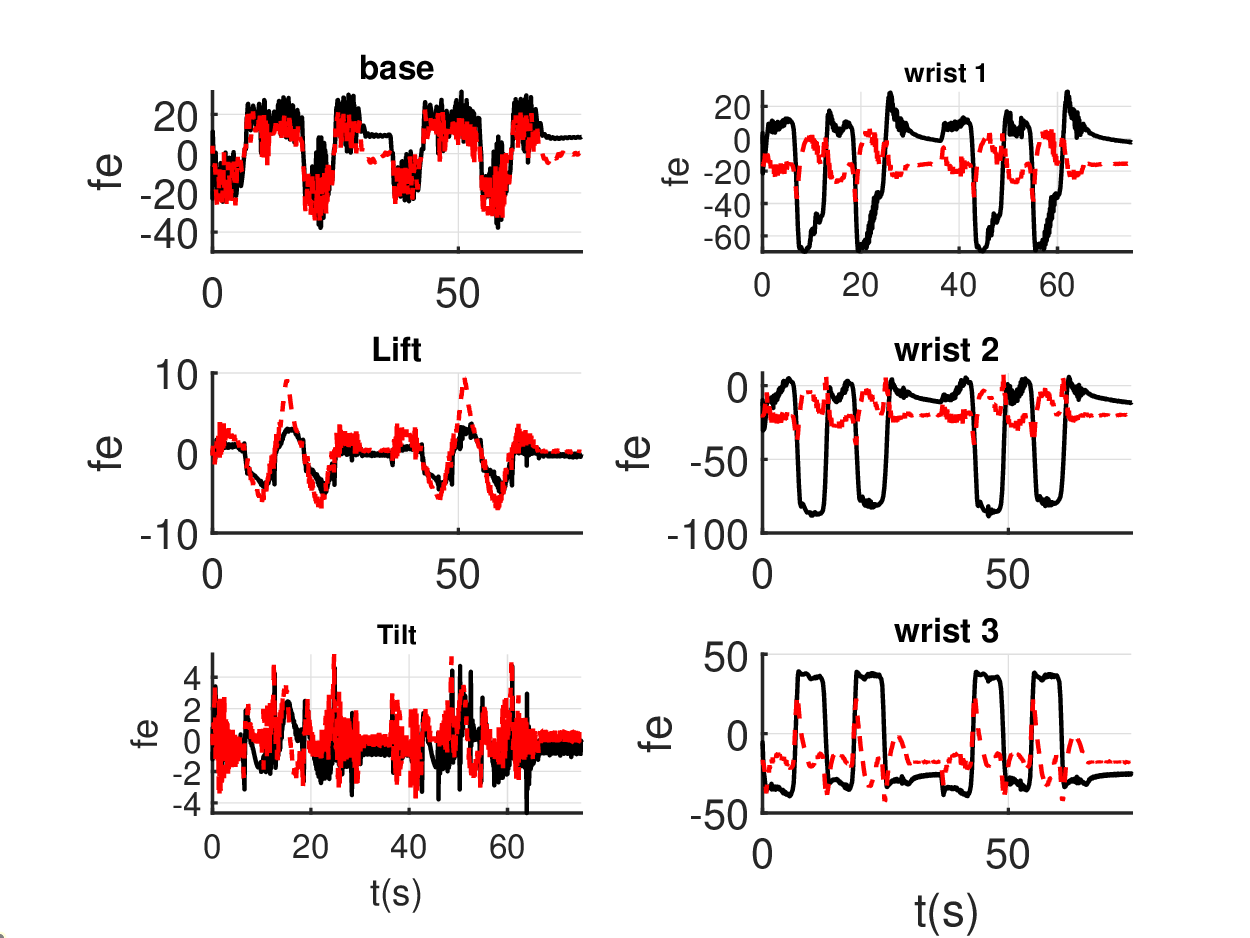}
\caption{force errors of each actuator with the implemented controllers. The dashed red line is with ORC, and the black solid line is with VDC. The y-axis is in kN.} \label{fig: force error}
\end{figure}

\begin{figure}[b]
\centering
\includegraphics[width=.5\textwidth]{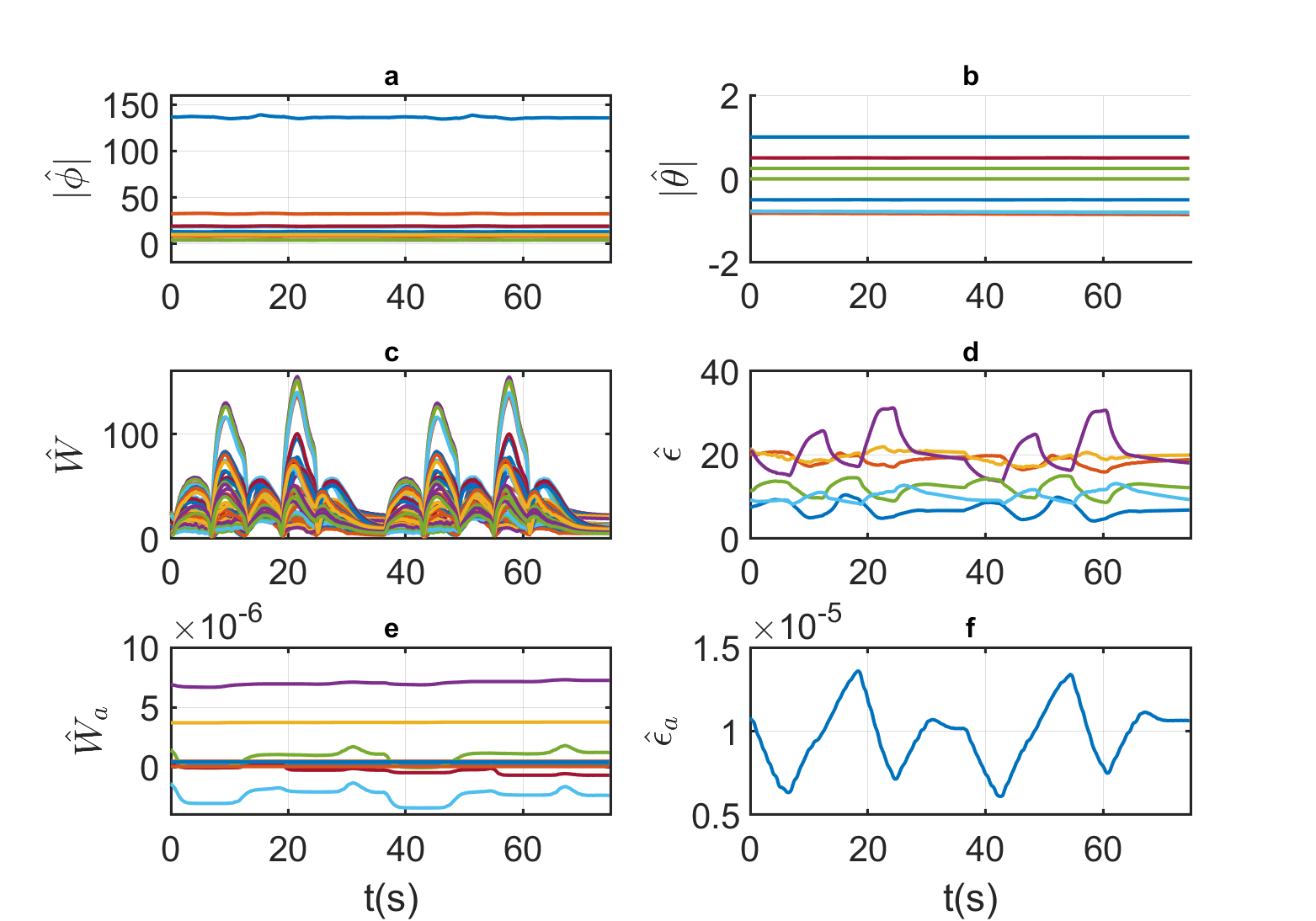}
\caption{Time history of estimated parameters, in which, each line displays the components of the corresponding vectors. a) norm of estimated inertial parameters, b) estimation of deadzone-backlash parameters, c) time history of updated weights of RBFNNs for the rigid body, d) estimation of bias error, e) weight estimation of RBFNNs for actuators, f) bias error estimation of actuators.} \label{fig: adaptation}
\end{figure}

The lower force error $(\Tilde{f} = f_{cr}-f_c)$, shows the better performance of the hydraulic actuator in generating the required forces. Such a good performance of the actuator can be dependent on friction and model uncertainties. The experimental results of the actuator force tracking error are demonstrated in Fig. \ref{fig: force error}, from which can be seen that the actuator with ORC had a much lower force error than VDC, displaying the impact of contributions of this study. Fig. \ref{fig: adaptation}, also, shows the time history of the estimated states. Fig. \ref{fig: adaptation}a shows the norm of all the estimated inertial parameter vectors of the rigid body in (\ref{Fr des final}). Fig. \ref{fig: adaptation}b displays the estimated adaptive inverse deadzone-backlash parameters. Fig. \ref{fig: adaptation}c and \ref{fig: adaptation}d demonstrate the norm of estimation of weights and approximation errors of the DRBFNNs in rigid body subsystems. Finally, Fig. \ref{fig: adaptation}e and \ref{fig: adaptation}f, show the norm of estimated weights and approximation error in the actuator subsystems. All the presented figures show the boundedness of all signals and the good performance of the proposed method, validating the theoretical results. Table \ref{table compare JC} summarizes the comparison between the original VDC and the proposed method. It can be seen that for a given roughly the same power source, the proposed approach achieved considerably lower tracking errors in comparison to the original VDC \cite{zhu2010virtual, hejrati2022decentralized}. The generated actuator forces, also, with ORC have a lower difference with the required value in contrast to the original VDC.

In order to check the sensitivity of the proposed controller to gain, three different experiments were performed with $70\%$, $90\%$, and $100\%$ of the tuned gains. The RMSE of all joints was computed to compare the results. The total RMSE of joint tracking for  $70\%$, $90\%$, and $100\%$ are 0.455, 0.27, and 0.157, respectively. These results show that by setting different gains to the controller, it will remain stable, and only the tracking performance will be reduced. Therefore, the controller is robust to the changes in gains.

\begin{table}[t]
    \centering
    \renewcommand{\arraystretch}{1.5}
    \caption{Performance evaluation}
    \label{table compare JC}
    \begin{tabular}{P{1.2cm} |P{.8cm} |P{.4cm} P{.5cm} P{.5cm} P{.5cm} P{.5cm} P{.5cm}}
        \hline
        \multicolumn{2}{c|}{Controllers} & {B} & {L} & {T} & {W1} & {W2} & {W3} \\[0.5ex] 
        \hline \hline
        \multirow{2}{*}{$e_{rms} (^{\circ}) $} & VDC & 0.14 & 0.15  & 0.12 &  0.94 &  1.38 &  0.59     \\ 
        \cline{2-8}
        & ORC & 0.07 & 0.03  & 0.06 &  0.18 &  0.19 &  0.25    \\
        \hline
        \multirow{2}{*}{$|e|_{max} (^{\circ})$} & VDC & 0.4 & 0.62  & 0.38 &  1.77 &  2.6 &  1.3     \\
        \cline{2-8}
        & ORC & 0.23 & 0.16  & 0.24 &  0.89 &  0.8 &  1.03    \\
        \hline
        \multirow{2}{*}{$\Tilde{f}_{rms} (kN)$} & VDC & 15.3 & 2  & 1.13 &  31.53 &  44.82 &  31.5     \\ 
        \cline{2-8}
        & ORC & 13.3 & 3.1  & 1.2 &  16.74 &  19.4 &  18.9    \\
        \hline
        \multirow{2}{*}{$u_{rms} (v)$} & VDC & 0.7 & 0.6  & 0.55 &  0.53 &  0.4 &  0.62     \\ 
        \cline{2-8}
        & ORC & 0.51 & 0.56  & 0.55 &  0.51 &  0.41 &  0.62    \\
        \hline
    \end{tabular}
\end{table}

\begin{figure}[b]
\centering
\includegraphics[width=.5\textwidth]{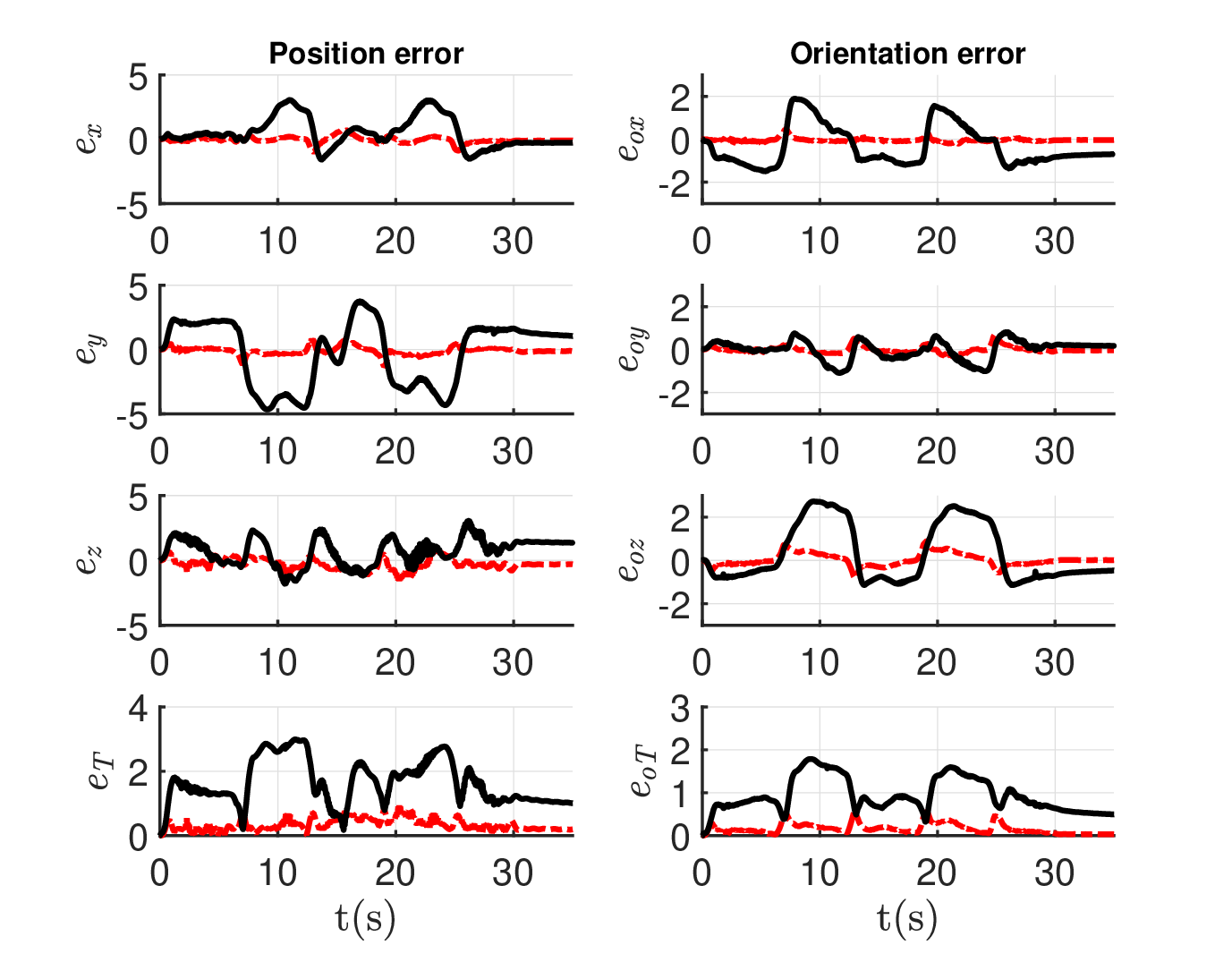}
\caption{Position and orientation tracking error of the end-effector in the $x-y-z$ plane. The solid black line is the tracking error with VDC and the dashed red line is with ORC. The position and orientation errors are presented in $cm$ and $deg$, respectively. } \label{fig: cartz error}
\end{figure}

To examine the tracking performance of the controller in Cartesian space, the experimental results of the path following are provided in Fig. \ref{fig: cartz error} for both the proposed method and the original VDC.  As depicted in Fig. \ref{fig: cartz error}, the proposed method displayed better performance than the original VDC. The RMSE in the $x$, $y$, and $z$ directions with VDC are $12.5\, mm$, $25\, mm$, and $12.5\, mm$, respectively, and with the proposed method, they are $2.5\, mm$, $3.2\, mm$, and $5.2\, mm$, respectively, with maximum end-effector velocities of $0.35\, m/s$, $0.43\, m/s$, and $0.55\, m/s$ in $x$, $y$, and $z$ directions, respectively. The steady-state error in the z-direction with VDC is $14\, mm$ because the motion in the z-direction is mostly generated by the base joint, which is suffering from backlash. Considering the backlash in our proposed method, the steady-state error in the z-direction is reduced to $3\, mm$. The maximum value of the Cartesian space motion error $e_T$, demonstrated in Fig. \ref{fig: cartz error}, with VDC, is $3\, cm$, and with the proposed method is $0.9\, cm$, with the maximum Cartesian velocity of $0.62\, m/s$. Therefore, the performance index in (\ref{rho index}) for VDC and the proposed method is $0.048$ and $0.015$, respectively. It can also be seen from Fig. \ref{fig: cartz error} that by tackling the backlash and model uncertainties in the wrist, the proposed method achieved a much better orientation error. Maximum value of orientation error $e_{oT}$ with VDC and the proposed method is $1.78$ and $0.55$ degrees, respectively. All these results demonstrate a significant improvement in tracking performance.

{
So far, the provided experimental results have thoroughly evaluated the position tracking accuracy of the proposed controller in both joint and Cartesian spaces. The following key insights demonstrate the exceptional performance and impact of the method:}

\begin{itemize}
    \item {The steady-state errors in both the base and wrist joints, specifically in \mbox{\(\zeta_1\), \(\xi_1\), \(\xi_2\)}, and \mbox{\(\xi_3\)}, which were previously suffering from backlash nonlinearity, have shown substantial improvement. The errors have been reduced by an outstanding {69.23\%}, {98.33\%}, {97.14\%}, and {85.71\%}, respectively. This demonstrates the effectiveness of the controller in overcoming complex nonlinearities.}
    \item {For each individual joint (from the base to the wrist), a significant improvement in maximum tracking error has been achieved, with reductions of {42.5\%}, {74.19\%}, {36.84\%}, {49.71\%}, {69.23\%}, and {20.76\%}. These reductions are a testament to the robustness of the controller and its precision despite of unknown uncertainties and model complexities.}
    \item {These remarkable improvements in joint space directly translate into a {70\%} reduction in the total end-effector position tracking error, and a {69.1\%} reduction in the total end-effector orientation error, as shown in Fig. \mbox{\ref{fig: cartz error}}. This is a dramatic improvement in the overall system accuracy, enabling precision operations in complex environments.}
    \item {The sensitivity of the controller to different gains are evaluated, revealing that with 70\%, 90\%, and 100\% of the gains applied, the total RMSE of all joints are 0.455 degrees, 0.27 degrees, and 0.157 degrees, respectively. These results highlight the controller’s robustness, demonstrating that while performance slightly decreases with changes in gain, the system maintains its overall reliability and accuracy.}
    \item {All these performance improvements have been achieved while maintaining the same level of control effort, highlighting the efficiency and practicality of the proposed controller. This ensures that the system can operate at a high performance without requiring excessive computational resources or additional power consumption (almost 70\% of CPU were used to ensure real-time computation).}
\end{itemize}

{
The performance index defined in \mbox{(\ref{rho index})} is a valid and insightful indicator of the controller's performance. Unlike conventional metrics, this index incorporates both the velocity of the system and the tracking error, offering a more comprehensive perspective on system behavior. Achieving low tracking errors at lower velocities is relatively straightforward; however, increased system velocity excites higher frequencies of the unmodeled dynamics, significantly challenging the controller's ability to maintain precise motion tracking. The \mbox{\(\rho\)} index is uniquely suited to evaluate the performance under these conditions, as it effectively captures the velocity-dependent accuracy of the system. A critical aspect of \mbox{\(\rho\)} is its ability to explain the velocity at which the tracking error is achieved. Lower \mbox{\(\rho\)} values consistently represent superior tracking accuracy at a given velocity. However, to ensure fairness and relevance in comparisons, the DoFs of the system must be accounted for. An increase in system DoFs inherently imposes greater demands on the controller, and thus, direct comparisons of \mbox{\(\rho\)} values across methods without normalization can lead to misleading conclusions. It was also reported in\mbox{\cite{koivumaki2015stability}} that the established \mbox{\(\rho\)} with two DoF was smaller than with three DoF, indicating the effect of nonlinearity in control performance.}

\begin{table}[b]
\centering
\caption{Performance index evaluation of Cartesian space}
\label{table rho}
\setlength{\tabcolsep}{10pt}
\begin{tabular}{ P{70pt} P{15pt} P{40pt} P{40pt} }
\hline
Study&  $\rho$ &  EE-DoF &  $\rho$/EE-DoF \\ [0.5ex]
\hline \hline
This study &
\textbf{0.015}&
\textbf{6}&
\textbf{0.0025}\\
Koivumaki 2019 \cite{koivumaki2019energy}& 
0.0057&
2&
0.0029\\
Kalmari 2015 & 
0.12&
3&
0.04\\
Zhu 2005 & 
0.015&
3&
0.005\\
Tsukamoto 2002 & 
0.126&
6&
0.021\\
Egelan 1987 & 
0.038&
3&
0.012\\
\hline
\end{tabular}
\label{tab}
\end{table}

{To address this, this study normalizes the performance index relative to the controlled DoFs in the Cartesian space for the end-effector (EE-DoF). For example, the maximum EE-DoF is six, encompassing three DoFs for position and three for orientation. Table \mbox{\ref{table rho}} presents the computed \mbox{\(\rho\)} values from this study alongside those reported in \mbox{\cite{mattila2017survey}}, including their normalization based on EE-DoF.

The results in Table \mbox{\ref{table rho}} clearly highlight the superiority of the proposed method. Even when compared to approaches with fewer EE-DoFs, which inherently benefit from less complexity, the proposed method demonstrates a better normalized performance index. This underscores the importance of normalization in fairly assessing controller performance. Remarkably, the proposed method achieves precise motion tracking for a 6-DoF HHM, despite facing significant challenges such as unknown model uncertainties, input nonlinearities, and system complexities.}

Considering all the experimental results provided in this section, along with the comparison to the original VDC approach and other existing methods, the validity of the theoretical results is verified. As shown, tackling unknown model and actuator uncertainties with RBFNNs and handling compound input nonlinearities with an adaptive deadzone-backlash controller significantly improved the tracking performance, both in transient and steady-state. As it was mathematically proved in Theorem \ref{thm: total} and experimentally shown in this section, the designed controller ensured SGUUB stability for the system and converged the tracking error to a substantially small neighborhood of origin.

\begin{figure*}[t]
      \centering
      \subfloat[]{\includegraphics[width = 0.195\textwidth]{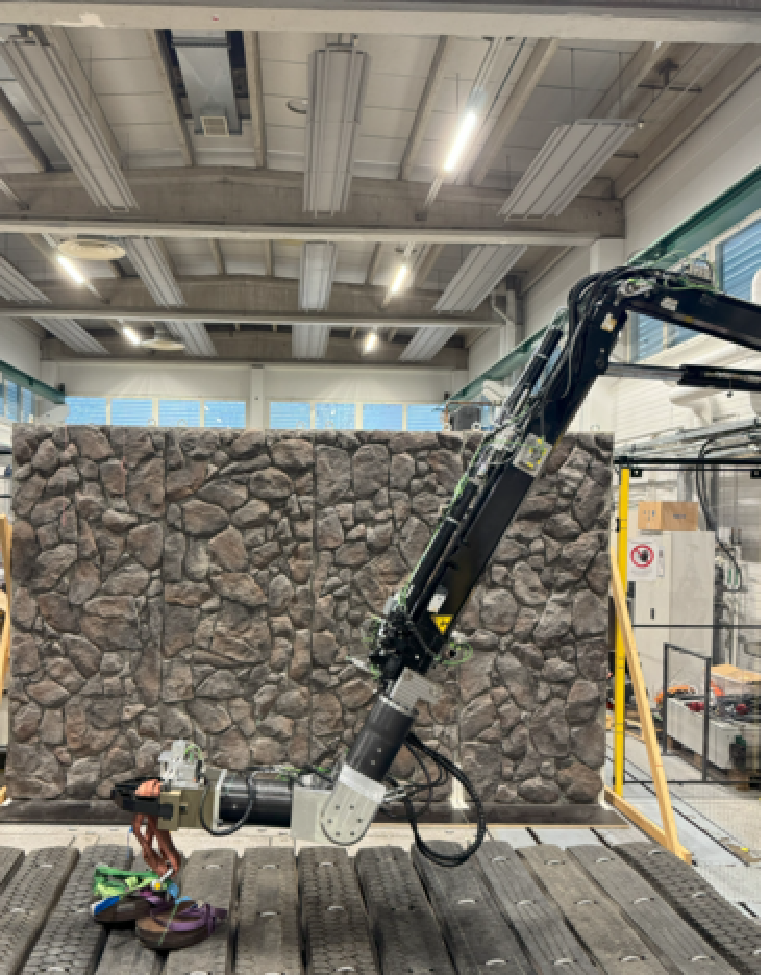}}
      \centering
      \label{D1}
      \subfloat[]{\includegraphics[width = 0.195\textwidth]{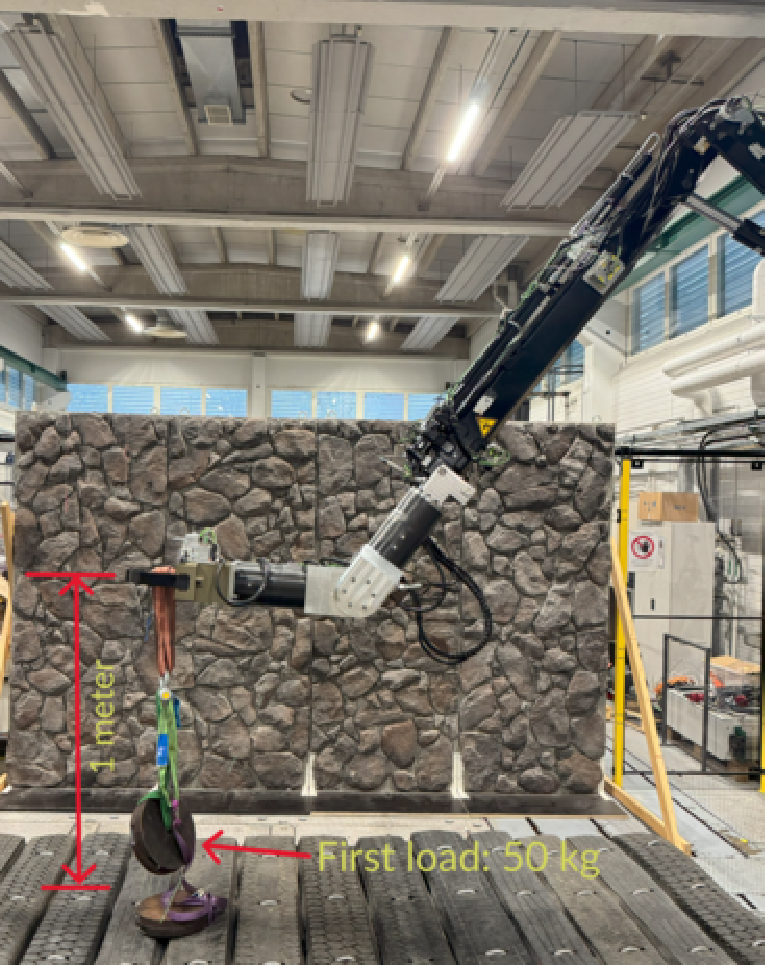}}
      \centering
      \label{D2}
      \subfloat[]{\includegraphics[width = 0.195\textwidth]{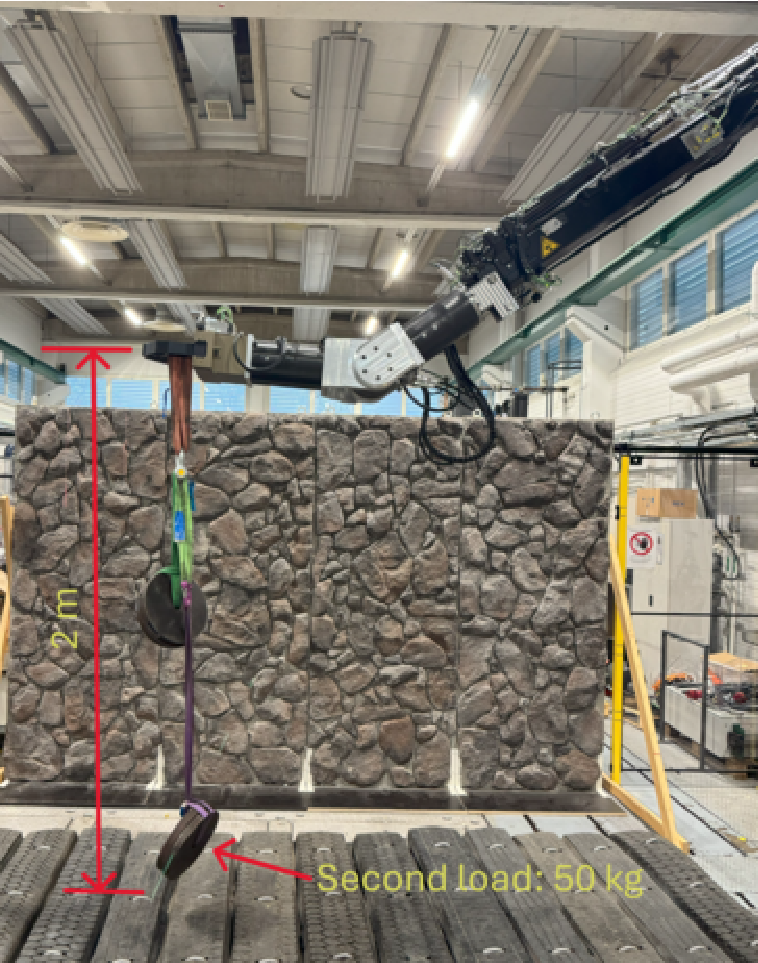}}
      \centering
      \label{D3}
      \subfloat[]{\includegraphics[width = 0.195\textwidth]{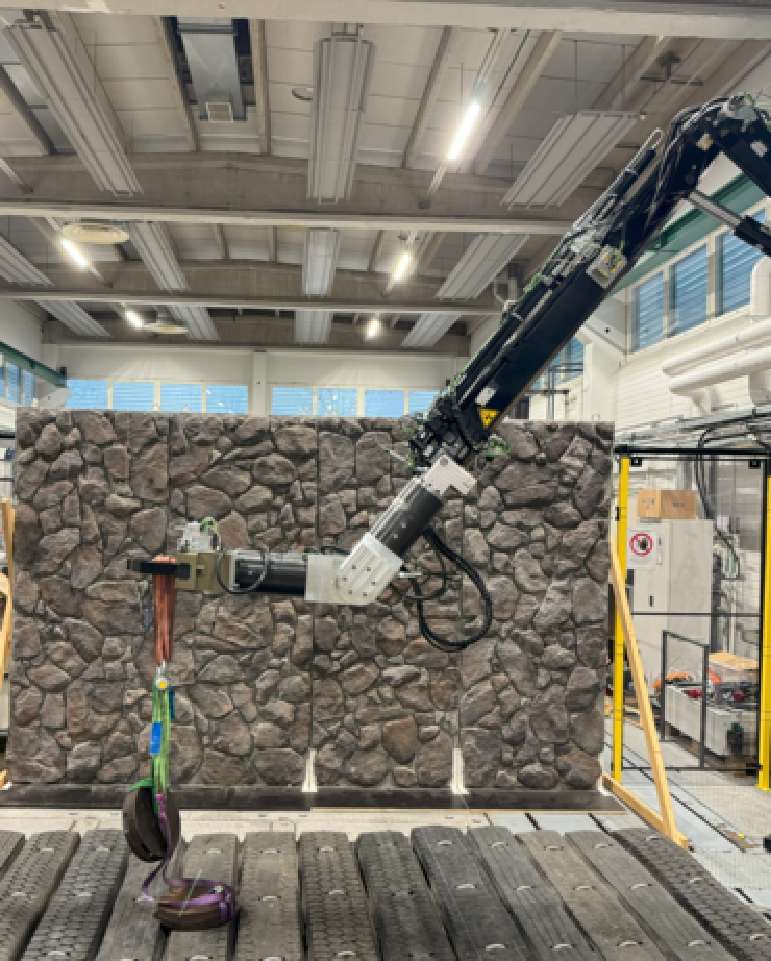}}
      \centering
      \label{D4}
      \subfloat[]{\includegraphics[width = 0.195\textwidth]{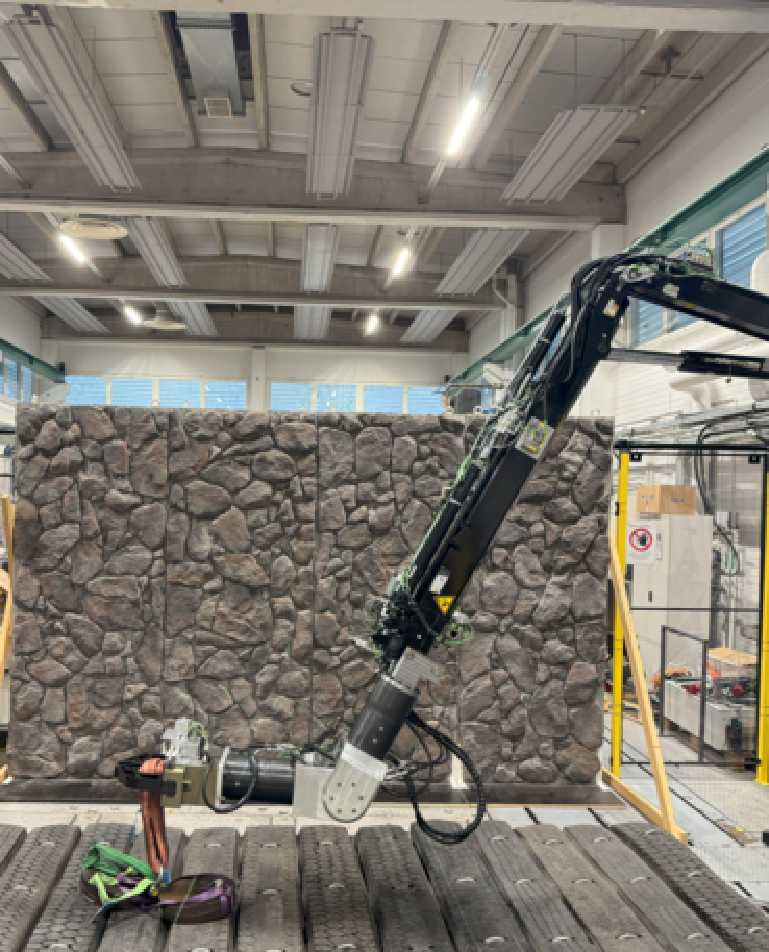}}
      \centering
      \label{DD5}
      \caption{ {Performance analysis in the presence of external disturbances and uncertainties. Each image shows the transition from one set point to another each of which is one meter.}}
      \label{Disturbance}
   \end{figure*}
   
\subsection{{Disturbance Rejection Performance Analysis}}
\begin{figure}[t]
\centering
{
\includegraphics[width=.48\textwidth]{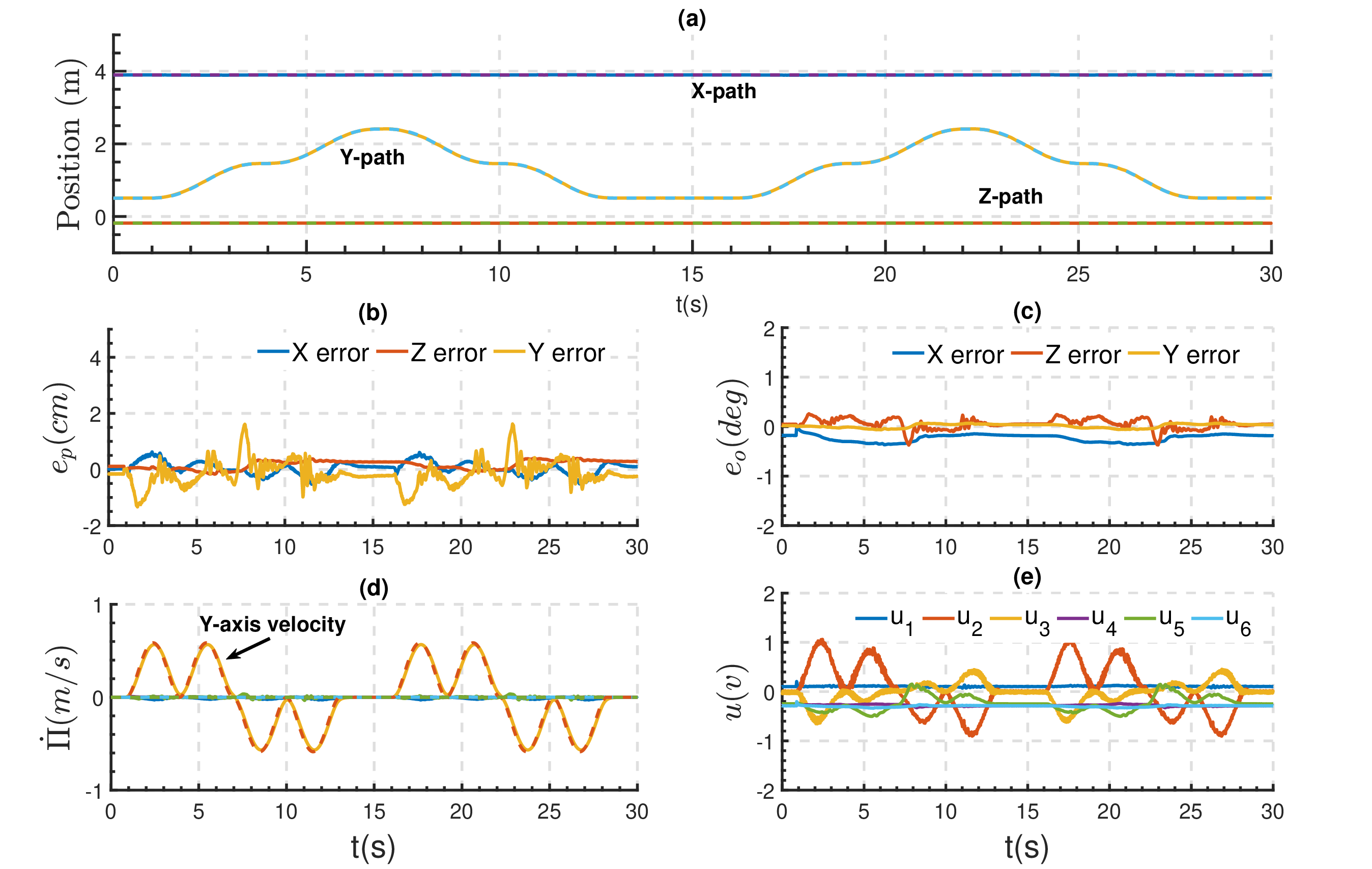}}
\caption{{Disturbance-rejection performance of the proposed ORC controller with an end-effector velocity of approximately 0.6 m/s. Dashed lines are desired values, while solid lines are actual values.} } \label{Dt3s}
\end{figure}

{In this section, we evaluate the robustness of the proposed controller against unknown external disturbances and uncertainties. To achieve this, we attach a two-meter-long rope to the end-effector of the HHM while it operates in full-pose control mode, as shown in Fig. \mbox{\ref{Disturbance}}. Along this rope, we secure two 50 kg weights, each positioned one meter apart. The experiment begins with the end-effector following a predefined trajectory in the y-direction. At the midpoint of this motion, the weights are suddenly added, creating an unexpected disturbance that the controller must compensate for in real time. The sequence of steps are demonstrated in Fig. \mbox{\ref{Disturbance}a-e}. Initially, this additional mass acts as an unknown disturbance, introducing oscillations due to its swinging motion during the lift. This dynamic interference negatively impacts the controller's performance, temporarily altering the system dynamics and requiring rapid compensation. However, as the motion continues, the disturbance becomes an unknown uncertainty, requiring the controller to adapt and maintain precise tracking performance. This setup effectively simulates real-world scenarios where the system encounters unforeseen dynamic changes. The experiments are conducted at three different end-effector velocities: 0.6 m/s, 0.7 m/s, and 0.8 m/s along the direction of motion. This variation enables a more comprehensive evaluation of the proposed controller's ability to handle external disturbances across practical working velocities. 

As shown in Fig. \mbox{\ref{Dt3s}a}, the proposed controller successfully maintains precise path tracking in the y-direction despite external disturbances, while also regulating motion in the x- and z-directions with high accuracy. Additionally, Figs. \mbox{\ref{Dt3s}b} and c present the position and orientation tracking errors, respectively. The y-direction error, which is most affected by external disturbances, remains within 2 cm, with minor peaks caused by disturbance interference. Despite this challenge, the controller demonstrates strong robustness while maintaining excellent tracking performance. Moreover, the errors in the x- and z-directions are minimal, highlighting the controller's ability to effectively isolate disturbances and maintain stability across all motion axes. The total RMSE of position and orientation error were 3.24 mm and 0.15 degree for this experiment. Fig. \mbox{\ref{Dt3s}d} showcases the velocity tracking performance, where the proposed ORC method precisely follows the desired velocity trajectory with RMSE of 0.02 cm/s, ensuring smooth motion of the heavy manipulator. Finally, Fig. \mbox{\ref{Dt3s}e} presents the voltage command applied to the manipulator's valves, revealing a smooth control signal. This further underscores the practicality and real-world applicability of the proposed controller, proving its effectiveness in handling dynamic disturbances while achieving high-performance motion control.}

\begin{figure}[h]
\centering
{
\includegraphics[width=.48\textwidth]{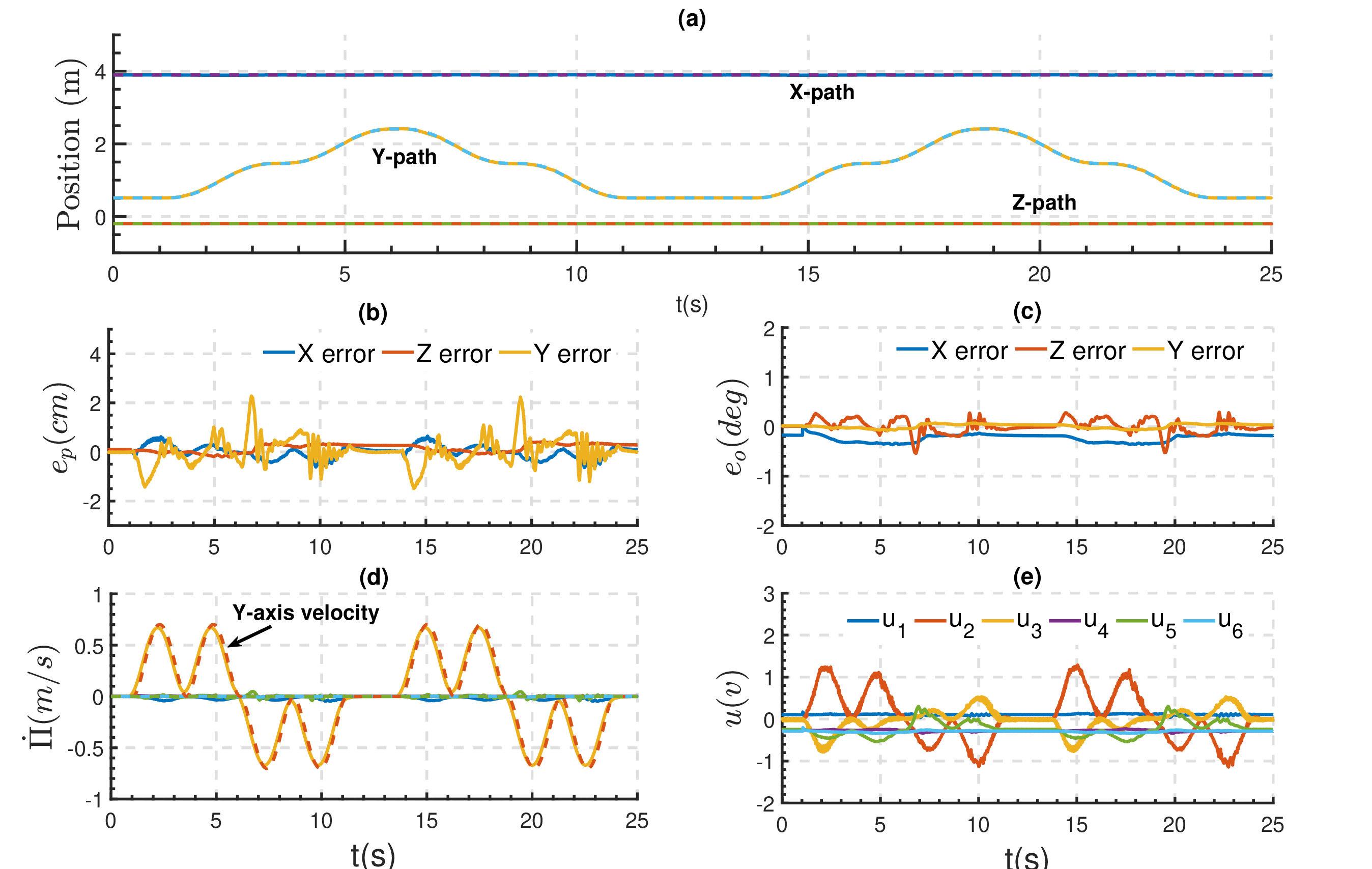}}
\caption{{Disturbance-rejection performance of the proposed ORC controller with an end-effector velocity of approximately 0.7 m/s. Dashed lines are desired values, while solid lines are actual values.}} \label{D2_5s}
\end{figure}

\begin{figure}[h]
\centering
{
\includegraphics[width=.48\textwidth]{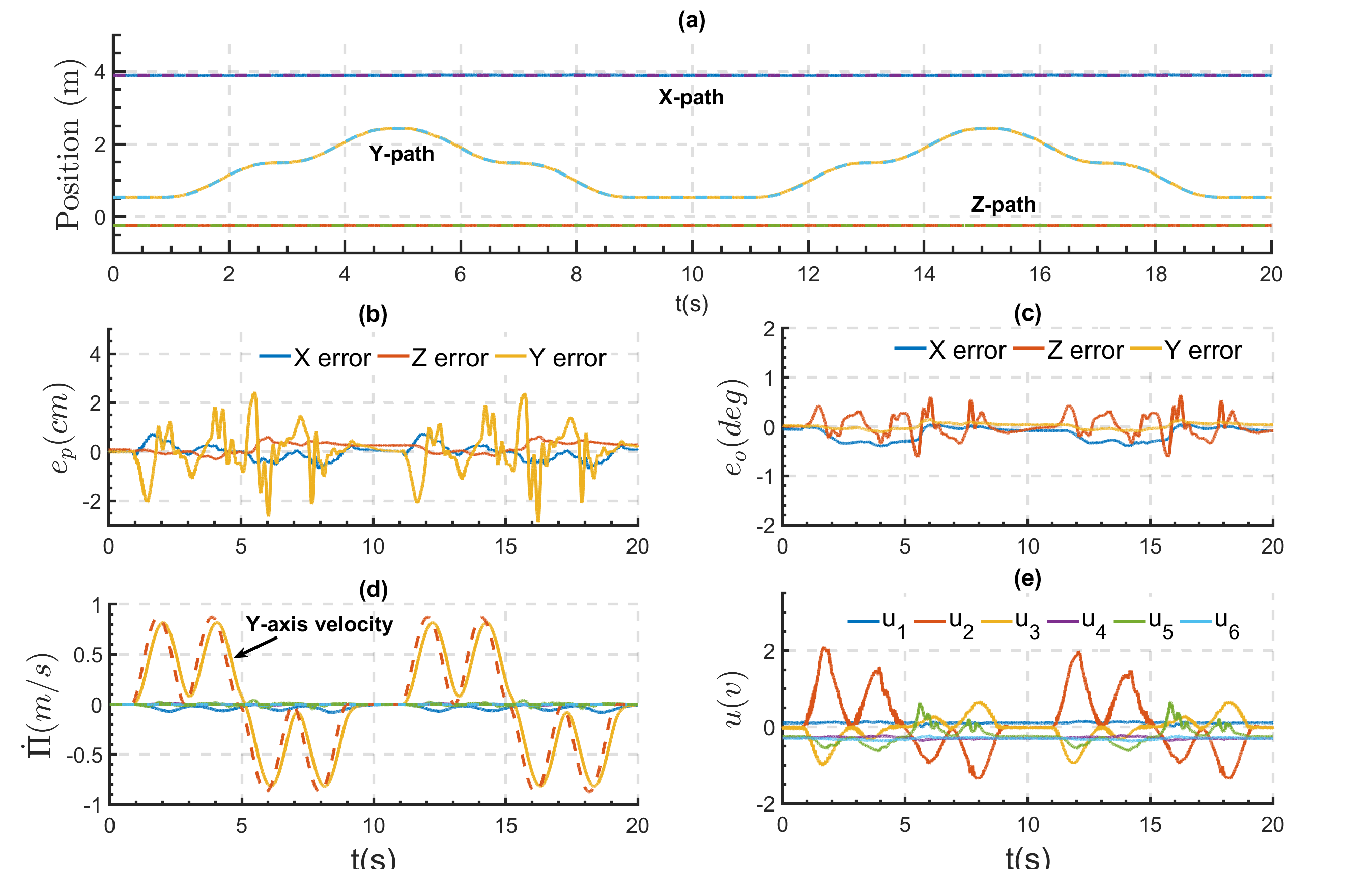}}
\caption{{Disturbance-rejection performance of the proposed ORC controller with an end-effector velocity of approximately 0.8 m/s. Dashed lines are desired values, while solid lines are actual values.}} \label{D2s}
\end{figure}

{Figs. \mbox{\ref{D2_5s}} and \mbox{\ref{D2s}} present the experimental results for end-effector velocities of 0.7 m/s and 0.8 m/s, respectively. These results provide a striking demonstration of the controller’s ability to handle extreme dynamic conditions. As expected, increasing the end-effector velocity amplifies the effect of external disturbances. However, the proposed controller not only stabilized the system despite the highly nonlinear and unpredictable behavior induced by the swinging loads but also achieved low tracking errors, with peak deviations contained around just 2 cm. The overall RMSE for position and orientation tracking at 0.7 m/s were 3.6 mm and 0.15 degrees, respectively, while at 0.8 m/s, they remained at 5.12 mm and 0.17 degrees. These results highlight the unprecedented accuracy and robustness of the controller, especially considering the extreme challenges posed by such high-velocity motion in the direction of gravity for a heavy manipulator with almost 5 meters workspace. The controller’s ability to maintain such superior tracking accuracy under these conditions is a testament to its performance, making it an feasible solution for precision motion control in real-world, high-impact applications.}

\subsection{Discussion}
{The proposed control method has been successfully implemented on a real-world 6-DoF HHM using a Beckhoff industrial PC with a 1ms sampling time, achieving efficient real-time performance with approximately \mbox{70\%} CPU utilization. While the hybrid nature of the controller, combining model-based (VDC) and non-model-based (RBFNNs) schemes, ensures high precision, increasing the system’s DoF could lead to higher computational demands and potentially limit real-time implementability on less capable hardware. Additionally, higher DoF systems would require more controller gains, making the tuning process more arduous. However, the controller's robustness to gain variations, which primarily impacts tracking accuracy rather than stability, helps mitigate this challenge. These considerations highlight common challenges for high-precision control systems, yet the demonstrated performance of the proposed method validates its practical applicability.}

\section{Conclusion}
In this study, an orchestrated robust controller is designed to achieve a precise and robust performance. For this purpose, VDC was employed as the baseline controller. VDC decomposes the entire complex system into subsystems of the rigid body and actuators by means of VCP. Then, the controller design and stability analysis are accomplished at the subsystem level and expanded to the entire system by means of VPFs and virtual stability. A novel way of incorporating RBFNNs and adaptive inverse deadzone-backlash controllers into each subsystem resulted in the orchestrated robust controller approach. Considering all unknown model and actuator uncertainties, compound input nonlinearities, and unknown disturbances, the SGUUB stability is achieved for the first time in the context of VDC. All theoretical results are validated via extensive simulation and experimental results concerning the highly nonlinear, six DoF HHM. The simulation results demonstrated that the proposed controller outperformed the original VDC, PD, and the state-of-the-art method in the presence of compound input nonlinearities. Moreover, the experimental results demonstrate that the proposed method tackles unknown model and actuator uncertainties and reduces the impact of compound input nonlinearities in real-world applications, resulting in considerably lower transient and steady-state tracking errors and lower actuator force errors in comparison to the original VDC for a given power. Additionally, the proposed method ended up with a precise tracking performance in Cartesian space in comparison to the original VDC. The robustness of the scheme, moreover, is examined in presence of unknown external disturbances and its excellent performance is demonstrated. All the analyses, including sensitivity to control gain, tracking performance in joint space and Cartesian space, comparison to the state-of-the-art controller, performance evaluation in the sense of $\rho$ value, disturbance rejection performance, and implementation on the real-world generic manipulator, display the universality of the proposed method in theory and applications.

{Potential future works could focus on extending the proposed control algorithm to tasks involving both free motion and interaction with the environment, such as pick-and-place operations. This could include the design of an impedance control framework to improve performance in scenarios requiring environmental contact. Additionally, exploring techniques to enhance energy efficiency would make the approach more suitable for mobile platforms. Another promising direction would be the integration of intelligent task planning algorithms, such as reinforcement learning, to enable high-level decision-making and broaden the applicability of the proposed control method in industrial settings.}

\appendix
\subsection{Proof of Theorem \ref{thm: rigid body}}
    Subtracting (\ref{Fr des final}) from (\ref{eqn: tot force 2}) and recalling (\ref{Delta def}), we have:
    \begin{equation}
    \begin{split}
        {^{ A} F_r^*} - {^{ A} F^*} &= M_{ A} \dfrac{\rm d}{\mathrm{d}t} \left( {^{ A}V_r}-{^{ A}V} \right) + C_{ A} \left( {^{ A}{\omega}}  \right) ( {^{ A}V_r}-{^{ A}V})\\
        &+ K_{ A} \, \left({^{ A} {  V}_r} - {^{ A} { V}} \right)+ Y_{A}\Tilde{\phi}_{ A}+\, {^{ A} }\Tilde{W}^T\Psi(\chi_{ A}) +\, {^{ A}}\Tilde{\varepsilon}.
        \end{split}
        \label{Fr-F}
    \end{equation}
    Now, by defining the non-negative accompanying function in the sense of Definition \ref{Def: 4} as:
    \begin{equation}
    \begin{split}
	{\nu}_1 &= \sum_{ A \in \Upsilon} \dfrac{1}{2} \, \left( {^{ A}{ V}_r} - {^{ A}{ V}}  \right)^T \, {\rm M_{\rm A}} \, \left( {^{ A}{ V}_r} - {^{ A}{ V}}  \right)\\
    &+\, \sum_{ A \in \Upsilon} \gamma \mathcal{D}_F(\mathcal{L}_{A}\rVert \hat{\mathcal{L}}_{A}) + \frac{1}{2}tr({^{ A}\Tilde{W}^T}\,{^{ A}\Gamma^{-1}}{^{ A}\Tilde{W}}) \\
        &+ \frac{1}{2\,{^{ A}}\pi} {^{ A}\Tilde{\varepsilon}^T}\,{^{ A}\Tilde{\varepsilon}}.
    \end{split}
	\label{eqn: v function for RB}
    \end{equation}
    where $\mathcal{D}_F(\mathcal{L}_{A}\rVert \hat{\mathcal{L}}_{A})$ is in the sense of Lemma \ref{Lemma: 3}, ${^{ A}\Gamma} \in \mathbb{R}^{\Bar{n}_A\times \Bar{n}_A}$ is a positive-definite matrix, and ${^{ A}}\pi \in \mathbb{R}$ is a positive constant gain. Consider $\mathcal{D}_F(\mathcal{L}_{A}\rVert \hat{\mathcal{L}}_{A}) \leq tr(\Tilde{\mathcal{L}}_{A}\,\Tilde{\mathcal{L}}_{A})$ With $ ||\Tilde{\mathcal{L}}_{A}||_F^2= tr(\Tilde{\mathcal{L}}_{A}\,\Tilde{\mathcal{L}}_{A})$, where $||.||_F$ is the Frobenius norm of a matrix. It can be shown for ${\nu}_1$ that:
    \begin{equation}
        \begin{split}
            {\nu}_1 &\leq \mu_1 ( \sum_{ A \in \Upsilon} \left({^{ A}{ V}_r} - {^{ A}{ V}}\right)^T\, \left( {^{ A}{ V}_r} - {^{ A}{ V}}\right) \\
            &+ tr({^{ A}\Tilde{W}}^T{^{ A}\Tilde{W}}) + {^{ A}\Tilde{\varepsilon}^T}{^{ A}\Tilde{\varepsilon}}+tr(\Tilde{\mathcal{L}}_{A}\,\Tilde{\mathcal{L}}_{A}))
        \end{split}
        \label{max v2}
    \end{equation}
    with $\mu_1 = \max\left\lbrace \gamma, \lbrace \lambda_{max}(M),0.5\left(\lambda_{max}(\Gamma)\right),0.5{^{ A}}\pi^{-1} \rbrace_{ A \in  \Upsilon} \right\rbrace$, where $\lambda_{max}(.)$ is the maximum eigenvalue of the matrix.
    Taking the time derivative of (\ref{eqn: v function for RB}) with using (\ref{Fr-F}), defining $^{ A}\mathbf{s} =\, ^{ A}Y^T\,(^{ A}{ V}_r-\,^{ A}{ V})$, and $^{ A}\Tilde{\phi}^T\,{^{ A}\mathbf{s}} = tr(\,{^{ A}\Tilde{\mathcal{L}}}\, {^{ A}\mathcal{I}(\mathbf{s})})$ (\(^{ A}\mathcal{I}(\mathbf{s})\) defined in Appendix \ref{Apendix D}), exploiting $(^{ A}{ V}_r-\,^{ A}{ V})^T\,{^{ A} }\Tilde{W}^T\Psi(\chi_{ A}) = tr((^{ A}{ V}_r-\,^{ A}{ V})\,\Psi^T(\chi_{ A})\,{^{ A} }\Tilde{W})$, and replacing from (\ref{L adapt})-(\ref{eps adapt}), one can achieve:
    \begin{equation}
        \begin{split}
            \Dot{{\nu}}_1 &= \, \sum_{ A \in \Upsilon} \left( {^{ A}{ V}_r} - {^{ A}{ V}}  \right)^T \, ({^{ A} F_r^*} - {^{ A} F^*}) \\
            &-  \, \left( {^{ A}{ V}_r} - {^{ A}{ V}}  \right)^T K_{ A} \, \left({^{ A} {  V}_r} - {^{ A} { V}} \right)\\
        &-\gamma_0tr(\hat{\mathcal{L}}_{A}\,\Tilde{\mathcal{L}}_{A})- {^{ A}}\tau_0tr(\,{^{ A}\hat{W}^T}{^{ A}\Tilde{W}}) - {^{ A}}\pi_0{{^{ A}}\Tilde{\varepsilon}^T} {^{ A}}\hat{\varepsilon}.
        \end{split}
        \label{Dv2 3}
    \end{equation}
    
    The first term in (\ref{Dv2 3}) is the VPF of the corresponding frame $\left\lbrace  A \right\rbrace$. In order to expand the VPFs, we need to use the Definition \ref{Def: VPF} for each frame in $\left\lbrace  \Upsilon \right\rbrace$. Starting from $\left\lbrace  P_1 \right\rbrace$ and recalling (\ref{eqn: BV trans}), (\ref{eqn: Pillar Force}), and (\ref{eqn: Pillar Force req}), we have:
    \begin{equation}
    \begin{split}
        \Dot{{\nu}}_{{ P_1}} &= \left( {^{ P_1}{ V}_r} - {^{ P_1}{ V}}  \right)^T \, ({^{ P_1} F_r^*} - {^{ P_1} F^*}) \\
        &= \left( {^{ P_1}{ V}_r} - {^{ P_1}{ V}}  \right)^T \, ({^{ P_1} F_r} - {^{ P_1} F})\\
        &+\left( {^{ P_1}{ V}_r} - {^{ P_1}{ V}}  \right)^T\,{^{{P_1}}{U}_{B_{c1}}}  \, ({^{{B_{c1}}} F_r} - {^{{B_{c1}}} F})\\
        &= p_{{ P_1}}-\left( {^{{P_1}}{U}_{B_{c1}}^T} \left({^{ P_1}{ V}_r} - {^{ P_1}{ V}}  \right)\right)^T  \, ({^{{B_{c1}}} F_r} - {^{{B_{c1}}} F})\\
        &= p_{{ P_1}}- \left({^{ B_{c1}}{ V}_r} - {^{ B_{c1}}{ V}}  \right)^T  \, ({^{{B_{c1}}} F_r} - {^{{B_{c1}}} F})
    \end{split}
    \label{P1 vpf 1}
    \end{equation}
    By replacing from (\ref{eqn: driven cc force req}) and (\ref{eqn: driven cc force}) for $j = 1$ and taking the same procedure in (\ref{P1 vpf 1}), we have:
        \begin{equation}
            \begin{split}
                 \Dot{{\nu}}_{{ P_1}} &= p_{{ P_1}}- \left({^{ B_{c1}}{ V}_r} - {^{ B_{c1}}{ V}}  \right)^T  \, \left({^{{B_{01}}} F_r} - {^{{B_{01}}} F}\right)\\
        &+\left({^{ B_{c1}}{ V}_r} - {^{ B_{c1}}{ V}}  \right)^T\,{^{{B}_{01}}{U}_{B_{11}}}  \, \left({^{{B_{11}}} F_r} - {^{{B_{11}}} F}\right)\\
        &-\,\left({^{ B_{c1}}{ V}_r} - {^{ B_{c1}}{ V}}  \right)^T  \,{^{{B}_{01}}{U}_{B_{11}}} \, \left({^{{B_{11}}} F_r} - {^{{B_{11}}} F}\right)\\
        &+\,\left({^{ B_{c1}}{ V}_r} - {^{ B_{c1}}{ V}}  \right)^T  \,{^{{B}_{01}}{U}_{B_{c2}}} \, \left({^{{B_{c2}}} F_r} - {^{{B_{c2}}} F}\right)\\
        &-\left({^{ B_{c1}}{ V}_r} - {^{ B_{c1}}{ V}}  \right)^T  \,{^{{B}_{01}}{U}_{P_{11}}} \, \left({^{{P_{11}}} F_r} - {^{{P_{11}}} F}\right)\\
        &-\,\left({^{ B_{c1}}{ V}_r} - {^{ B_{c1}}{ V}}  \right)^T  \,{^{{B}_{21}}{U}_{B_{31}}} \, \left({^{{B_{31}}} F_r} - {^{{B_{31}}} F}\right)\\
        &+\left({^{ B_{c1}}{ V}_r} - {^{ B_{c1}}{ V}}  \right)^T  \,{^{{B}_{21}}{U}_{B_{41}}} \, \left({^{{B_{41}}} F_r} - {^{{B_{41}}} F}\right)\\
        &-\,\left({^{ B_{c1}}{ V}_r} - {^{ B_{c1}}{ V}}  \right)^T  \,{^{{B}_{21}}{U}_{B_{41}}} \, \left({^{{B_{41}}} F_r} - {^{{B_{41}}} F}\right)\\
        &+\,\left({^{ B_{c1}}{ V}_r} - {^{ B_{c1}}{ V}}  \right)^T  \,{^{{B}_{21}}{U}_{P_{11}}} \, \left({^{{P_{11}}} F_r} - {^{{P_{11}}} F}\right)\\
        &-\,\left({^{ B_{c1}}{ V}_r} - {^{ B_{c1}}{ V}}  \right)^T  \,{^{{B}_{01}}{U}_{B_{c2}}} \, \left({^{{B_{c2}}} F_r} - {^{{B_{c2}}} F}\right)\\
        & = p_{{ P_1}}-p_{{ B_{01}}}+p_{{ B_{11}}}-p_{{ B_{11}}}+p_{{ B_{C2}}}-p_{{ P_{11}}}\\
        &-p_{{ B_{21}}}+p_{{ B_{41}}}-p_{{ B_{41}}}+p_{{ P_{11}}}-p_{{ B_{C2}}}\\
        &=p_{{ P_1}}-p_{{ B_{01}}}-p_{{ B_{21}}}.
            \end{split}
            \label{P1 vpf 2}
        \end{equation}
    In the same sense, we can write for all other frames in $\left\lbrace  \Upsilon \right\rbrace$:
    \begin{equation}
    \begin{split}
        \Dot{{\nu}}_{{ P_{p2}}} &=  p_{{ P_{p2}}}
        \end{split}
        \label{Pp2 vpf}
    \end{equation}
    \begin{equation}
    \begin{split}
        \Dot{{\nu}}_{{ B_{0j}}} 
        &=  p_{{ B_{0j}}}-p_{{ B_{1j}}}
        \end{split}
        \label{B0j vpf}
    \end{equation}
    \begin{equation}
    \begin{split}
        \Dot{{\nu}}_{{ B_{1j}}} 
        &=  p_{{ B_{1j}}}-p_{{ B_{c,j+1}}}+p_{{ P_{1j}}}
        \end{split}
        \label{B1j vpf}
    \end{equation}
    \begin{equation}
    \begin{split}
        \Dot{{\nu}}_{{ B_{3j}}} 
        &=  p_{{ B_{2j}}}-p_{{ B_{4j}}}+( f_{cjr} - f_{cj})(\Dot{x}_{pjr}-\Dot{x}_{pj})
        \end{split}
        \label{B3j vpf}
    \end{equation}
    \begin{equation}
    \begin{split}
        \Dot{{\nu}}_{{ B_{4j}}} 
        &=  p_{{ B_{4j}}}-p_{{ P_{1j}}}
        \end{split}
        \label{B4j vpf}
    \end{equation}
    \begin{equation}
    \begin{split}
        \Dot{{\nu}}_{{ G_{i}}} 
        &=  p_{{ G_{i}}}-p_{{ E_{i+1}}}.
        \end{split}
        \label{Gi vpf}
    \end{equation}
    On the other hand, in (\ref{Dv2 3}), by recalling Definition \ref{Lemma: 2}, using Trace operator property and the fact that $\hat{\mathcal{L}}_{A}$ is a positive-definite and symmetric matrix a long with $\hat{\mathcal{L}}_{A} = \Tilde{\mathcal{L}}_{A} + \mathcal{L}_{A}$, we have: $-tr(\hat{\mathcal{L}}_{A}\,\Tilde{\mathcal{L}}_{A}) = -tr(\Tilde{\mathcal{L}}_{A}\,\Tilde{\mathcal{L}}_{A})-tr(\mathcal{L}_{A}\,\Tilde{\mathcal{L}}_{A})$. Additionally, it can be shown that 
        $-tr(\mathcal{L}_{A}\,\Tilde{\mathcal{L}}_{A}) \leq 0.5\left(tr(\Tilde{\mathcal{L}}_{A}\,\Tilde{\mathcal{L}}_{A})+tr(\mathcal{L}_{A}\,\mathcal{L}_{A})\right)$. Therefore, we can write that $-tr(\hat{\mathcal{L}}_{A}\,\Tilde{\mathcal{L}}_{A}) \leq -0.5\,tr(\Tilde{\mathcal{L}}_{A}\,\Tilde{\mathcal{L}}_{A})+0.5\,tr(\mathcal{L}_{A}\,\mathcal{L}_{A})$. In the same way, we can write that $-{{^{ A}}\Tilde{\varepsilon}^T} {^{ A}}\hat{\varepsilon} \leq -0.5\,{{^{ A}}\Tilde{\varepsilon}^T}{{^{ A}}\Tilde{\varepsilon}}+\,0.5\,{{^{ A}}\varepsilon^T}{{^{ A}}\varepsilon}$, and $-tr(\,{^{ A}\hat{W}^T}{^{ A}\Tilde{W}})\leq -0.5\,tr(\,{^{ A}\Tilde{W}^T}{^{ A}\Tilde{W}})+\,0.5\,tr(\,{^{ A}W}^T{^{ A}W})$. Therefore, by replacing (\ref{P1 vpf 2})-(\ref{Gi vpf}) in (\ref{Dv2 3}), and considering $p_{B_{c2}} = p_{B_{02}}+p_{B_{22}}$ and $p_{B_{c3}} = p_{E_{1}}$ with using (\ref{max v2}), we have:
    \begin{equation}
        \begin{split}
             \Dot{{\nu}}_1 & \leq -\mu {\nu}_1 + \mu_{0}+ VPFs
        \end{split}
        \label{Dv2 5}
    \end{equation}
    where $VPFs = p_{P_1}+p_{{ P_{p2}}}+p_{G_{1}}+p_{G_{2}}+p_{G_{3}}-p_{E_{1}}-p_{E_{2}}-p_{E_{3}}-p_{E_{4}}+\sum_{j=1,2}( f_{cjr} - f_{cj})(\Dot{x}_{pjr}-\Dot{x}_{pj})$. Additionally, $\mu = \mu_2/\mu_1$, where $\mu_2 = \min \left\lbrace \gamma_0, \lbrace \lambda_{min}(^{A}K),0.5{^{ A}}\tau_0,0.5{^{ A}}\pi_{0}^{-1}\rbrace_{ A \in  \Upsilon} \right\rbrace$, and $\mu_0 = \sum_{ A \in \Upsilon} 0.5\gamma_0tr(\mathcal{L}_{A}\,\mathcal{L}_{A})+0.5{^{ A}}\tau_0tr(\,{{^{ A}}W^T}{^{ A}W})+0.5{^{ A}}\pi_0{{^{ A}}\varepsilon^T} {^{ A}}\varepsilon$.

\subsection{Proof of Theorem \ref{thm: actuator}}
    By subtracting (\ref{ufr}) from (\ref{eqn: fp with uf}), adding $(\Dot{f}_{pr}-\Dot{f}_{pr})/\beta$ to the right side, using (\ref{eqn: uf def}), (\ref{final DB error}), and (\ref{eqn: uf def req}), we have:
    
    \begin{equation}
    \begin{split}
        &\frac{1}{\beta}(\Dot{f}_{pr}-\Dot{f}_{p}) = -Y_d \, \Tilde{\theta}_d - Y_v \Tilde{\theta}_v-k_f(f_{pr}-f_p)\\
        &-k_x(\Dot{x}_{pr}-\Dot{x}_p)-\Tilde{W}_a^T\Psi(\chi_a) -\,\Tilde{\varepsilon}_a - \Tilde{\theta}^T\,\eta.
        \label{Dfr and Df}
    \end{split}
    \end{equation}
    Define the non-negative accompanying function as:
    \begin{equation*}
    \begin{split}
        \nu_a &= \left( f_{ppr} - f_{pp} \right)^2/(2 \, \beta \, k_{xp})+\sum_{j=1,2}\left( f_{pjr} - f_{pj} \right)^2/(2 \, \beta \, k_{xj})\\
        &+\sum_{i=1,2,3}\left( f_{pwir} - f_{pwi} \right)^2/(2 \, \beta \, k_{xwi})+ \frac{1}{2\gamma_{dp}}\left(\Tilde{\theta}_{dp}^T\,\Tilde{\theta}_{dp}\right)\\
        &+ \sum_{j=1,2}\frac{1}{2\gamma_{dj}}\left(\Tilde{\theta}_{dj}^T\, \Tilde{\theta}_{dj}\right)+\sum_{i=1,2,3}\frac{1}{2\gamma_{dwi}}\left(\Tilde{\theta}_{dwi}^T\,\Tilde{\theta}_{dwi}\right)\\
        & + \frac{1}{2\gamma_{vp}}\left(\Tilde{\theta}_{vp}^T\,\Tilde{\theta}_{vp}\right)+ \sum_{j=1,2}\frac{1}{2\gamma_{vj}}\left(\Tilde{\theta}_{vj}^T\,\Tilde{\theta}_{vj}\right)\\
        &+\sum_{i=1,2,3}\frac{1}{2\gamma_{vwi}}\left(\Tilde{\theta}_{vwi}^T\,\Tilde{\theta}_{vwi}\right)+ \sum_{j=1,2}\frac{1}{2\gamma_{fj}}\left(\Tilde{\theta}_{fj}^T\,\Tilde{\theta}_{fj}\right)\\
        & + \frac{1}{2\gamma_{fp}}\left(\Tilde{\theta}_{fp}^T\,\Tilde{\theta}_{fp}\right)+\sum_{i=1,2,3}\frac{1}{2\gamma_{fwi}}\left(\Tilde{\theta}_{fwi}^T\,\Tilde{\theta}_{fwi}\right)\\
        \end{split}
    \end{equation*}
    \begin{equation}
    \begin{split}
        &+\frac{1}{2\delta_{p}}\left(\Tilde{W}_{ap}^T\,\Tilde{W}_{ap}\right)+ \sum_{j=1,2}\frac{1}{2\delta_{j}}\left(\Tilde{W}_{aj}^T\,\Tilde{W}_{aj}\right)\\
        &+\sum_{i=1,2,3}\frac{1}{2\delta_{wi}}\left(\Tilde{W}_{awi}^T\,\Tilde{W}_{awi}\right)+\sum_{i=1,2,3}\frac{1}{2\Bar{\delta}_{wi}}\left(\Tilde{\varepsilon}_{awi}\right)^2\\
        &+ \frac{1}{2\Bar{\delta}_{p}}\left(\Tilde{\varepsilon}_{ap}\right)^2+ \sum_{j=1,2}\frac{1}{2\Bar{\delta}_{j}}\left(\Tilde{\varepsilon}_{aj}\right)^2+ \frac{1}{2\delta_{p\theta}}\left(\Tilde{\theta}_{p}^T\Tilde{\theta}_{p}\right)\\
        &+ \sum_{j=1,2}\frac{1}{2\delta}_{j\theta}\left(\Tilde{\theta}_{j}^T\,\Tilde{\theta}_{j}\right)+\sum_{i=1,2,3}\frac{1}{2\delta}_{wi\theta}\left(\Tilde{\theta}_{awi}^T\,\Tilde{\theta}_{awi}\right)
    \end{split}
        \label{nu_a 1}
    \end{equation}
    with $k_{x(.)}$ being a positive constant. By computing (\ref{Dfr and Df}) for each linear hydraulic actuator, choosing the adaptation functions as in (\ref{hat theta f})- (\ref{hat vareps d}), (\ref{fpr}), (\ref{fp1}), and (\ref{friction lin-in-par}), and substituting in the time derivative of (\ref{nu_a 1}), we have:
    \begin{equation}
    \begin{split}
        \Dot{\nu}_{a} &= -\frac{k_{fp}}{k_{xp}}\,( f_{ppr} - f_{pp})^2-\sum_{j=1,2}\frac{k_{fj}}{k_{xj}}( f_{pjr} - f_{pj})^2\\
        &-\sum_{i=1,2,3}\frac{k_{fwi}}{k_{xwi}}( f_{pwir} - f_{pwi})^2\\
        &-( f_{cpr} - f_{cp})(\Dot{x}_{ppr}-\Dot{x}_{pp})-\sum_{j=1,2}( f_{cjr} - f_{cj})(\Dot{x}_{pjr}-\Dot{x}_{pj})\\
        &-\sum_{i=1,2,3}( f_{cwir} - f_{cwi})(\Dot{x}_{pwir}-\Dot{x}_{pwi})\\
        &- \gamma_{dp0}\Tilde{\theta}_{dp}^T\,\hat{\theta}_{dp}-\sum_{j=1,2}\gamma_{dj0}\,\Tilde{\theta}_{dj}^T\hat{\theta}_{dj}- \sum_{i=1,2,3}\gamma_{dwi0}\,\Tilde{\theta}_{dwi}^T\hat{\theta}_{dwi}\\
        &- \gamma_{vp0}\Tilde{\theta}_{vp}^T\hat{\theta}_{vp}-\sum_{j=1,2}\gamma_{vj0}\Tilde{\theta}_{vj}^T\hat{\theta}_{vj}\\
        &- \sum_{i=1,2,3}\gamma_{vwi0}\Tilde{\theta}_{vwi}^T\hat{\theta}_{vwi}- {\gamma_{fp0}}\Tilde{\theta}_{fp}^T\hat{\theta}_{fp}-\sum_{j=1,2}{\gamma_{fj0}}\Tilde{\theta}_{fj}^T\hat{\theta}_{fj}\\
        &- \sum_{i=1,2,3}{\gamma_{fwi0}}\Tilde{\theta}_{fwi}^T\hat{\theta}_{fwi}\\
        &- {\delta_{p0}}\Tilde{W}_{ap}^T\hat{W}_{ap}-\sum_{j=1,2}{\delta_{j0}}\Tilde{W}_{aj}^T\hat{W}_{aj}- \sum_{i=1,2,3}{\delta_{wi0}}\Tilde{W}_{awi}^T\hat{W}_{awi}\\
        &-{\Bar{\delta}_{p0}}\Tilde{\varepsilon}_{ap}\hat{\varepsilon}_{ap}-\sum_{j=1,2}{\Bar{\delta}_{j0}}\Tilde{\varepsilon}_{aj}\hat{\varepsilon}_{aj}\\
        &- \sum_{i=1,2,3}{\Bar{\delta}_{wi0}}\Tilde{\varepsilon}_{awi}\hat{\varepsilon}_{awi}
        - {\delta_{p\theta0}}\Tilde{\theta}_{p}^T\hat{\theta}_{p}-\sum_{j=1,2}{\delta_{j\theta0}}\Tilde{\theta}_{j}^T\hat{\theta}_{j}\\
        &- \sum_{i=1,2,3}{\delta_{wi\theta0}}\Tilde{\theta}_{wi}^T\hat{\theta}_{wi}
        \end{split}
        \label{Dnu_a2 2}
    \end{equation}
    By considering (\ref{Dxp}), (\ref{eqn: P1_Vr}), and (\ref{base piston force req}), one can rewrite the VPFs in (\ref{Dnu_a2 2}) as:
    \begin{equation}
        \begin{split}
            \Dot{\nu}_{a{{ P_{1}}}} &= -(\Dot{x}_{ppr}-\Dot{x}_{pp})( f_{cpr} - f_{cp}) \\
            & = -\left(\frac{1}{r_p}y_\tau(\Dot{x}_{ppr}-\Dot{x}_{pp})\right)^T\left({^{ P_{1}} F_r} - {^{ P_{1}} F}\right)\\
            &-\left(x_f(\Dot{x}_{ppr}-\Dot{x}_{pp})\right)^T\left({^{ P_{p2}} F_r} - {^{ P_{p2}} F}\right)\\
            & = -\left({^{ P_1} V_r}-{^{ P_1} V}\right)^T\left({^{ P_{1}} F_r} - {^{ P_{1}} F}\right)\\
            &-\left({^{ P_{p2}} V_r}-{^{ P_{p2}} V}\right)^T\left({^{ P_{p2}} F_r} - {^{ P_{p2}} F}\right)\\
            &+\left({^{ P_{p1}}{U_{P_{p2}}^T}}\left({^{ P_{p1}} V_r}-{^{ P_{p1}} V}\right)\right)^T\left({^{ P_{p1}} F_r} - {^{ P_{p1}} F}\right)\\
            & = -p_{P_{1}}-p_{P_{p2}}+p_{P_{p1}}
        \end{split}
        \label{Dnu_a P1}
    \end{equation}
    and,
    \begin{equation}
        \begin{split}
            \Dot{\nu}_{a{{ G_{i}}}} &= -( f_{cwir} - f_{cwi})(\Dot{x}_{pwir}-\Dot{x}_{pwi}) = -p_{G_{i}}+p_{E_{i}}.
        \end{split}
        \label{Dnu_a Gi}
    \end{equation}
    On the other hand, we know $\hat{\theta}_{(\varkappa)} = \Tilde{\theta}_{(\varkappa)}+{\theta}_{(\varkappa)}$. Then we can write $-\Tilde{\theta}_{(\varkappa)}^T\,\hat{\theta}_{(\varkappa)} = -\Tilde{\theta}_{(\varkappa)}^T\,\Tilde{\theta}_{(\varkappa)}-\Tilde{\theta}_{(\varkappa)}^T\,{\theta}_{(\varkappa)}$. Moreover, $-\Tilde{\theta}_{(\varkappa)}^T\,{\theta}_{(\varkappa)} \leq 0.5\left(\Tilde{\theta}_{(\varkappa)}^T\Tilde{\theta}_{(\varkappa)}+{\theta}_{(\varkappa)}^T{\theta}_{(\varkappa)}\right)$. Therefore, we have $-\Tilde{\theta}_{(\varkappa)}^T\,\hat{\theta}_{(\varkappa)} \leq -0.5\Tilde{\theta}_{(\varkappa)}^T\,\Tilde{\theta}+0.5{\theta}_{(\varkappa)}^T{\theta}_{(\varkappa)}$. By doing the same for all the same terms in (\ref{Dnu_a2 2}) and substituting (\ref{Dnu_a P1}) and (\ref{Dnu_a Gi}), one can obtain:
    \begin{equation*}
    \begin{split}
        \Dot{\nu}_{a} &\leq -\frac{k_{fp}}{k_{xp}}\,( f_{ppr} - f_{pp})^2-\sum_{j=1,2}\frac{k_{fj}}{k_{xj}}( f_{pjr} - f_{pj})^2\\
        &-\sum_{i=1,2,3}\frac{k_{fwi}}{k_{xwi}}( f_{pwir} - f_{pwi})^2+\, VPFs\\
        &- 0.5\gamma_{dp0}\Tilde{\theta}_{dp}^T\,\Tilde{\theta}_{dp} -0.5\sum_{j=1,2}\gamma_{dj0}\,\Tilde{\theta}_{dj}^T\Tilde{\theta}_{dj}\\
        &- 0.5\sum_{i=1,2,3}\gamma_{dwi0}\,\Tilde{\theta}_{dwi}^T\Tilde{\theta}_{dwi}-0.5 \gamma_{vp0}\Tilde{\theta}_{vp}^T\Tilde{\theta}_{vp}\\
        &-0.5\sum_{j=1,2}\gamma_{vj0}\Tilde{\theta}_{vj}^T\Tilde{\theta}_{vj}- 0.5\sum_{i=1,2,3}\gamma_{vwi0}\Tilde{\theta}_{vwi}^T\Tilde{\theta}_{vwi}\\
        &- 0.5{\gamma_{fp0}}\Tilde{\theta}_{fp}^T\Tilde{\theta}_{fp}- 0.5\sum_{i=1,2,3}{\gamma_{fwi0}}\Tilde{\theta}_{fwi}^T\Tilde{\theta}_{fwi}\\
        &-0.5\sum_{j=1,2}{\gamma_{fj0}}\Tilde{\theta}_{fj}^T\Tilde{\theta}_{fj}-0.5 \sum_{i=1,2,3}{\delta_{wi0}}\Tilde{W}_{awi}^T\Tilde{W}_{awi}\\
        &-0.5 {\delta_{p0}}\Tilde{W}_{ap}^T\Tilde{W}_{ap}-0.5\sum_{j=1,2}{\delta_{j0}}\Tilde{W}_{aj}^T\Tilde{W}_{aj}\\
        &-0.5{\Bar{\delta}_{p0}}\Tilde{\varepsilon}_{ap}\Tilde{\varepsilon}_{ap}-0.5\sum_{j=1,2}{\Bar{\delta}_{j0}}\Tilde{\varepsilon}_{aj}\Tilde{\varepsilon}_{aj}\\
        &- 0.5\sum_{i=1,2,3}{\Bar{\delta}_{wi0}}\Tilde{\varepsilon}_{awi}\Tilde{\varepsilon}_{awi}
        - 0.5{\delta_{p\theta0}}\Tilde{\theta}_{p}^T\Tilde{\theta}_{p}\\
        &-0.5\sum_{j=1,2}{\delta_{j\theta0}}\Tilde{\theta}_{j}^T\Tilde{\theta}_{j}- 0.5\sum_{i=1,2,3}{\delta_{wi\theta0}}\Tilde{\theta}_{wi}^T\Tilde{\theta}_{wi}\\
        &+0.5\gamma_{dp0}\theta_{dp}^T\,\theta_{dp} +0.5\sum_{j=1,2}\gamma_{dj0}\,\theta_{dj}^T\theta_{dj}\\
        &+ 0.5\sum_{i=1,2,3}\gamma_{dwi0}\,\theta_{dwi}^T\theta_{dwi}+0.5 \gamma_{vp0}\theta_{vp}^T\theta_{vp}\\
        &+0.5\sum_{j=1,2}\gamma_{vj0}\theta_{vj}^T\theta_{vj}+0.5\sum_{j=1,2}{\Bar{\delta}_{j0}}{\varepsilon}_{aj}{\varepsilon}_{aj}\\
        &+ 0.5\sum_{i=1,2,3}\gamma_{vwi0}\theta_{vwi}^T\theta_{vwi}+ 0.5{\gamma_{fp0}}\theta_{fp}^T\theta_{fp}\\
        &+0.5\sum_{j=1,2}{\gamma_{fj0}}\theta_{fj}^T\theta_{fj}+ 0.5\sum_{i=1,2,3}{\gamma_{fwi0}}\theta_{fwi}^T\theta_{fwi}\\
        &+0.5 {\delta_{p0}}{W}_{ap}^T{W}_{ap}+0.5\sum_{j=1,2}{\delta_{j0}}{W}_{aj}^T{W}_{aj}\\
        &+0.5 \sum_{i=1,2,3}{\delta_{wi0}}{W}_{awi}^T{W}_{awi}+0.5{\Bar{\delta}_{p0}}{\varepsilon}_{ap}{\varepsilon}_{ap}\\
        &+ 0.5\sum_{i=1,2,3}{\Bar{\delta}_{wi0}}{\varepsilon}_{awi}{\varepsilon}_{awi}
        + 0.5{\delta_{p\theta0}}\theta_{p}^T\theta_{p}\\
        \end{split}
    \end{equation*}
    \begin{equation}
    \begin{split}
        &+0.5\sum_{j=1,2}{\delta_{j\theta0}}\Tilde{\theta}_{j}^T\theta_{j}+ 0.5\sum_{i=1,2,3}{\delta_{wi\theta0}}\theta_{wi}^T\theta_{wi}\\
        &\leq -\mu_{a}\nu_{a2}+\mu_{a0}+VPFs.
        \end{split}
        \label{Dnu_a2 4}
    \end{equation}
     where $ VPFs = -p_{P_{1}}-p_{P_{p2}}+p_{P_{p1}}-\sum_{j=1,2}( f_{cjr} - f_{cj})(\Dot{x}_{pjr}-\Dot{x}_{pj})+\sum_{i=1,2,3}(-p_{G_{i}}+p_{E_{i}})$. In the same sense in (\ref{max v2}) and (\ref{Dv2 5}) we have $\mu_{a} = \mu_{1a}/\mu_{2a}$, where $\mu_{1a} = \min ( \frac{k_{fp}}{k_{xp}}, \frac{k_{fj}}{k_{xj}}, \frac{k_{fwi}}{k_{xwi}}, 0.5\gamma_{dp0},0.5\gamma_{dj0}, 0.5\gamma_{dwi0},0.5\gamma_{vp0}, 0.5\gamma_{vj0}\newline,0.5\gamma_{vwi0},0.5{\gamma_{fp0}}, 0.5{\gamma_{fj0}}, 0.5{\gamma_{fwi0}},0.5 {\delta_{p0}},0.5{\delta_{j0}}\newline, 0.5{\delta_{wi0}},0.5 {\Bar{\delta}_{p0}},0.5{\Bar{\delta}_{j0}}, 0.5{\Bar{\delta}_{wi0}},0.5 {\delta_{p\theta0}},0.5{\delta_{j\theta0}}, 0.5{\delta_{wi\theta0}})$, while $\mu_{2a} = \max (\frac{1}{2 \, \beta \, k_{xp}},\frac{1}{2 \, \beta \, k_{xj}},\frac{1}{2 \, \beta \, k_{xwi}},{\frac{1}{2 \, \gamma_{dp}}},\frac{1}{2 \, \gamma_{dj}},\frac{1}{2 \,\gamma_{dwi}}\newline,{\frac{1}{2 \, \gamma_{vp}}},\frac{1}{2 \, \gamma_{vj}},\frac{1}{2 \,\gamma_{vwi}},{\frac{1}{2 \, \gamma_{fp}}},\frac{1}{2 \, \gamma_{fj}},{\frac{1}{2 \,\gamma_{fwi}}},\frac{1}{2\delta_{p}},\frac{1}{2\delta_{j}},\frac{1}{2\delta_{wi}},\frac{1}{2\Bar{\delta}_{p}},\frac{1}{2\Bar{\delta}_{j}}\newline,\frac{1}{2\Bar{\delta}_{wi}},\frac{1}{2\delta_{p\theta}},\frac{1}{2\delta}_{j\theta}, \frac{1}{2\delta}_{wi\theta})$, and $\mu_{a0} = (0.5\gamma_{dp0}\theta_{dp}^T\,\theta_{dp} +0.5\sum_{j=1,2}\gamma_{dj0}\,\theta_{dj}^T\theta_{dj}+ 0.5\sum_{i=1,2,3}\gamma_{dwi0}\,\theta_{dwi}^T\theta_{dwi}+0.5\gamma_{vp0}\theta_{vp}^T\theta_{vp}+0.5\sum_{j=1,2}\gamma_{vj0}\theta_{vj}^T\theta_{vj}+ 0.5\sum_{i=1,2,3}\gamma_{vwi0}\theta_{vwi}^T\theta_{vwi}+ 0.5{\gamma_{fp0}}\theta_{fp}^T\theta_{fp}+0.5\sum_{j=1,2}{\gamma_{fj0}}\theta_{fj}^T\theta_{fj}+ 0.5\sum_{i=1,2,3}{\gamma_{fwi0}}\theta_{fwi}^T\theta_{fwi})+0.5 {\delta_{p0}}{W}_{ap}^T{W}_{ap}+0.5\sum_{j=1,2}{\delta_{j0}}{W}_{aj}^T{W}_{aj}+0.5 \sum_{i=1,2,3}{\delta_{wi0}}{W}_{awi}^T{W}_{awi}+0.5{\Bar{\delta}_{p0}}{\varepsilon}_{ap}{\varepsilon}+{ap}+0.5\sum_{j=1,2}{\Bar{\delta}_{j0}}{\varepsilon}_{aj}{\varepsilon}_{aj}+ 0.5\sum_{i=1,2,3}{\Bar{\delta}_{wi0}}{\varepsilon}_{awi}{\varepsilon}_{awi}+ 0.5{\delta_{p\theta0}}\theta_{p}^T\theta_{p}+0.5\sum_{j=1,2}{\delta_{j\theta0}}\Tilde{\theta}_{j}^T\theta_{j}+ 0.5\sum_{i=1,2,3}{\delta_{wi\theta0}}\theta_{wi}^T\theta_{wi}$.

\subsection{Proof of Theorem \ref{thm: total}}
    Define the non-negative accompanying function as below:
    \begin{equation}
        \nu = \nu_1+\nu_a
        \label{total nu}
    \end{equation}
    Taking the time derivative of (\ref{total nu}), one can obtain:
    \begin{equation}
        \begin{split}
            \Dot{\nu} &\leq -\mu {\nu}_1 + \mu_{0}+p_{P_{1}}+p_{P_{p2}}+p_{G_{1}}+p_{G_{2}}+p_{G_{3}}\\
            &-p_{E_{1}}-p_{E_{2}}-p_{E_{3}}-p_{E_{4}}+\sum_{j=1,2}( f_{cjr} - f_{cj})(\Dot{x}_{pjr}-\Dot{x}_{pj})\\
        &-\mu_{a}\nu_{a}+\mu_{a0}-p_{P_{1}}-p_{P_{p2}}+p_{E_{1}}+p_{E_{2}}+p_{E_{3}}\\
        &-\sum_{j=1,2}( f_{cjr} - f_{cj})(\Dot{x}_{pjr}-\Dot{x}_{pj})-p_{G_{1}}-p_{G_{2}}-p_{G_{3}}\\
        &\leq - \Bar{\mu}\nu+\Bar{\mu}_0+p_{P_{1}}-p_{E_{4}}
        \end{split}
        \label{final nu}
    \end{equation}
    with $\Bar{\mu} = \left(\min\lbrace \mu, \mu_a\rbrace\right)/\left(\max\lbrace \mu_1, \mu_{2a}\rbrace\right)$ and $\Bar{\mu}_0 = \mu_{0}+\mu_{a0}$ being positive. Moreover, since there is no contact with the environment and the base is fixed, $p_{E_{4}} = p_{P_{1}} = 0$. Now, multiplying both sides by $e^{\Bar{\mu}t}$, (\ref{final nu}) becomes:
    \begin{equation}
        \frac{d}{dt}(\nu(t)\,e^{\Bar{\mu}t})\leq\, \Bar{\mu}_0\,e^{\Bar{\mu}t}.
        \label{exp nu}
    \end{equation}
    Integrating (\ref{exp nu}) over $[0,t]$, one can obtain:
    \begin{equation}
        \nu(t)\,\leq \, \nu(0)\,e^{-\Bar{\mu}t}+\, \frac{\Bar{\mu}_0}{\Bar{\mu}}\,(1-e^{-\Bar{\mu}t})\, \leq\, \nu(0)+ \frac{\Bar{\mu}_0}{\Bar{\mu}}.
        \label{ nu(t)}
    \end{equation}
    Considering (\ref{eqn: v function for RB}) and (\ref{total nu}), we can derive for the velocity error $\dfrac{1}{2}( {^{ A}{ V}_r} - {^{ A}{ V}})^T \, { M_{ A}} \, ( {^{ A}{ V}_r} - {^{ A}{ V}})\, \leq \nu(t)\, \leq \, \nu(0)+ \frac{\Bar{\mu}_0}{\Bar{\mu}}$ which results in $||( {^{ A}{ V}_r} - {^{ A}{ V}})||\leq \sqrt{2\frac{\nu(0)+ \frac{\Bar{\mu}_0}{\Bar{\mu}}}{\lambda_{min}(M)}}$. Therefore, it can be concluded that $\lim_{t \rightarrow \infty} ||( {^{ A}{ V}_r} - {^{ A}{ V}})|| = \sqrt{\frac{2\Bar{\mu}_0}{\Bar{\mu}\,{\lambda_{min}(M)}}}$. It can be seen that by fine-tuning the control gains in $\Bar{\mu}$, the compact set can be very small near zero. In the same way, we can achieve $\lim_{t \rightarrow \infty} ||(f_{p(.)r}-f_{p(.)}||)\, \leq\, \sqrt{\frac{2\beta k_{x(.)}\Bar{\mu}_0}{\Bar{\mu}}}$ (with $(.)$ indicating all joints), which ensures the semi-global boundedness of the force tracking error in the presence of all the uncertainties. We can derive the same conclusion for the update law of the RBFNNs weights as $\lim_{t \rightarrow \infty} ||\Tilde{W}_{a(.)}||)\, \leq\, \sqrt{\frac{2\delta{(.)}\Bar{\mu}_0}{\Bar{\mu}}}$. For all other errors, the compact set can be achieved as well.

\subsection{Unique Matrix Definition}
\label{Apendix D}
The unique symmetric and  positive-definite matrix in (\ref{L adapt}) can be defined as,
\begin{equation*}
    ^A\mathcal{I}(\mathbf{s}) = \begin{bmatrix}
    \mathbf{s}_6+\mathbf{s}_7 & -0.5\mathbf{s}_8 & -0.5\mathbf{s}_{10} & 0.5\mathbf{s}_2\\
    -0.5\mathbf{s}_8 & \mathbf{s}_5+\mathbf{s}_7 & -0.5\mathbf{s}_9 & 0.5\mathbf{s}_3 \\
    -0.5\mathbf{s}_{10} & -0.5\mathbf{s}_9 & \mathbf{s}_5+\mathbf{s}_6 & 0.5\mathbf{s}_4 \\
    0.5\mathbf{s}_2 & 0.5\mathbf{s}_3 & 0.5\mathbf{s}_4 & \mathbf{s}_1
    \end{bmatrix}.
\end{equation*}

\bibliographystyle{IEEEtran}
\bibliography{mybib}

\begin{thebibliography}{10}
\providecommand{\url}[1]{#1}
\csname url@samestyle\endcsname
\providecommand{\newblock}{\relax}
\providecommand{\bibinfo}[2]{#2}
\providecommand{\BIBentrySTDinterwordspacing}{\spaceskip=0pt\relax}
\providecommand{\BIBentryALTinterwordstretchfactor}{4}
\providecommand{\BIBentryALTinterwordspacing}{\spaceskip=\fontdimen2\font plus
\BIBentryALTinterwordstretchfactor\fontdimen3\font minus \fontdimen4\font\relax}
\providecommand{\BIBforeignlanguage}[2]{{%
\expandafter\ifx\csname l@#1\endcsname\relax
\typeout{** WARNING: IEEEtran.bst: No hyphenation pattern has been}%
\typeout{** loaded for the language `#1'. Using the pattern for}%
\typeout{** the default language instead.}%
\else
\language=\csname l@#1\endcsname
\fi
#2}}
\providecommand{\BIBdecl}{\relax}
\BIBdecl

\bibitem{chen2023accurate}
J.~Chen, X.~Du, L.~Lyu, Z.~Fei, and X.-M. Sun, ``Accurate finite-time motion control of hydraulic actuators with event-triggered input,'' \emph{IEEE Transactions on Automation Science and Engineering}, 2023.

\bibitem{ABB8700}
ABB, ``8th generation of heavy payload robot,'' [Online]. Available: \url{https://search.abb.com/library/Download.aspx?DocumentID=3HAC052852-001&LanguageCode=en&DocumentPartId=&Action=Launch/}, accessed: November 29, 2023.

\bibitem{Sandvik}
S.~MineStories, ``The automine journey,'' [Online]. Available: \url{https://solidground.sandvik/automine-journey/}, accessed: January 1, 2025.

\bibitem{li2024energy}
L.~Li, M.~Cheng, R.~Ding, and B.~Xu, ``Energy efficiency improvement of hydraulic manipulator through flow-optimal redundancy resolution with path following,'' \emph{IEEE Transactions on Automation Science and Engineering}, 2024.

\bibitem{truong2023backstepping}
H.~V.~A. Truong, S.~Nam, S.~Kim, Y.~Kim, and W.~K. Chung, ``Backstepping-sliding-mode-based neural network control for electro-hydraulic actuator subject to completely unknown system dynamics,'' \emph{IEEE Transactions on Automation Science and Engineering}, 2023.

\bibitem{USExcavator}
K.~Gribbins, ``The ultimate mini ex overview: A comprehensive analysis of the 2021 compact excavator market,'' [Online]. Available: \url{https://compactequip.com/excavators/the-ultimate-mini-ex-overview-a-comprehensive-analysis-of-the-2021/-compact-excavator-market/}, accessed: September 27, 2023.

\bibitem{saidi2016robotics}
K.~S. Saidi, T.~Bock, and C.~Georgoulas, ``Robotics in construction,'' in \emph{Springer handbook of robotics}.\hskip 1em plus 0.5em minus 0.4em\relax Springer, 2016, pp. 1493--1520.

\bibitem{mattila2017survey}
J.~Mattila, J.~Koivum{\"a}ki, D.~G. Caldwell, and C.~Semini, ``A survey on control of hydraulic robotic manipulators with projection to future trends,'' \emph{iEeE/ASME Transactions on Mechatronics}, vol.~22, no.~2, pp. 669--680, 2017.

\bibitem{petrovic2022mathematical}
G.~R. Petrovi{\'c} and J.~Mattila, ``Mathematical modelling and virtual decomposition control of heavy-duty parallel--serial hydraulic manipulators,'' \emph{Mechanism and Machine Theory}, vol. 170, p. 104680, 2022.

\bibitem{bonchis2002experimental}
A.~Bonchis, P.~I. Corke, and D.~C. Rye, ``Experimental evaluation of position control methods for hydraulic systems,'' \emph{IEEE Transactions on Control Systems Technology}, vol.~10, no.~6, pp. 876--882, 2002.

\bibitem{bech2013experimental}
M.~M. Bech, T.~O. Andersen, H.~C. Pedersen, and L.~Schmidt, ``Experimental evaluation of control strategies for hydraulic servo robot,'' in \emph{2013 IEEE International Conference on Mechatronics and Automation}.\hskip 1em plus 0.5em minus 0.4em\relax IEEE, 2013, pp. 342--347.

\bibitem{bu2000observer}
F.~Bu and B.~Yao, ``Observer based coordinated adaptive robust control of robot manipulators driven by single-rod hydraulic actuators,'' in \emph{Proceedings 2000 ICRA. Millennium Conference. IEEE International Conference on Robotics and Automation. Symposia Proceedings (Cat. No. 00CH37065)}, vol.~3.\hskip 1em plus 0.5em minus 0.4em\relax IEEE, 2000, pp. 3034--3039.

\bibitem{bu2001desired}
F.~Bu and B.~Yuo, ``Desired compensation adaptive robust control of single-rod electro-hydraulic actuator,'' in \emph{Proceedings of the 2001 American Control Conference.(Cat. No. 01CH37148)}, vol.~5.\hskip 1em plus 0.5em minus 0.4em\relax IEEE, 2001, pp. 3926--3931.

\bibitem{zhu2005adaptive}
W.-H. Zhu and J.-C. Piedboeuf, ``Adaptive output force tracking control of hydraulic cylinders with applications to robot manipulators,'' \emph{Journal of Dynamic Systems, Measurement, and Control}, vol. 127, no.~2, pp. 206--2017, 2005.

\bibitem{zhu2010virtual}
W.-H. Zhu, \emph{Virtual decomposition control: toward hyper degrees of freedom robots}.\hskip 1em plus 0.5em minus 0.4em\relax Springer Science \& Business Media, 2010, vol.~60.

\bibitem{koivumaki2013automation}
J.~Koivum{\"a}ki and J.~Mattila, ``The automation of multi degree of freedom hydraulic crane by using virtual decomposition control,'' in \emph{2013 IEEE/ASME International Conference on Advanced Intelligent Mechatronics}.\hskip 1em plus 0.5em minus 0.4em\relax IEEE, 2013, pp. 912--919.

\bibitem{koivumaki2019energy}
J.~Koivum{\"a}ki, W.-H. Zhu, and J.~Mattila, ``Energy-efficient and high-precision control of hydraulic robots,'' \emph{Control Engineering Practice}, vol.~85, pp. 176--193, 2019.

\bibitem{liang2023adaptive}
X.~Liang, Z.~Yao, W.~Deng, and J.~Yao, ``Adaptive control of n-link hydraulic manipulators with gravity and friction identification,'' \emph{Nonlinear Dynamics}, pp. 1--17, 2023.

\bibitem{kim2019discrete}
J.~Kim, M.~Jin, W.~Choi, and J.~Lee, ``Discrete time delay control for hydraulic excavator motion control with terminal sliding mode control,'' \emph{Mechatronics}, vol.~60, pp. 15--25, 2019.

\bibitem{yang2022neural}
X.~Yang, W.~Deng, and J.~Yao, ``Neural adaptive dynamic surface asymptotic tracking control of hydraulic manipulators with guaranteed transient performance,'' \emph{IEEE Transactions on Neural Networks and Learning Systems}, 2022.

\bibitem{deng2022neural}
W.~Deng, H.~Zhou, J.~Zhou, and J.~Yao, ``Neural network-based adaptive asymptotic prescribed performance tracking control of hydraulic manipulators,'' \emph{IEEE transactions on systems, man, and cybernetics: systems}, vol.~53, no.~1, pp. 285--295, 2022.

\bibitem{liang2024adaptive}
X.~Liang, Z.~Yao, W.~Deng, and J.~Yao, ``Adaptive neural network finite-time tracking control for uncertain hydraulic manipulators,'' \emph{IEEE/ASME Transactions on Mechatronics}, 2024.

\bibitem{yao2024model}
Z.~Yao, X.~Liang, S.~Wang, and J.~Yao, ``Model-data hybrid driven control of hydraulic euler--lagrange systems,'' \emph{IEEE/ASME Transactions on Mechatronics}, 2024.

\bibitem{tao1995continuous}
G.~Tao and P.~V. Kokotovic, ``Continuous-time adaptive control of systems with unknown backlash,'' \emph{IEEE Transactions on Automatic Control}, vol.~40, no.~6, pp. 1083--1087, 1995.

\bibitem{recker1991adaptive}
D.~Recker, P.~Kokotovic, D.~Rhode, and J.~Winkelman, ``Adaptive nonlinear control of systems containing a deadzone,'' in \emph{[1991] Proceedings of the 30th IEEE Conference on Decision and Control}.\hskip 1em plus 0.5em minus 0.4em\relax IEEE, 1991, pp. 2111--2115.

\bibitem{zhao2022deterministic}
Z.~Zhao, W.~He, F.~Zhang, C.~Wang, and K.-S. Hong, ``Deterministic learning from adaptive neural network control for a 2-dof helicopter system with unknown backlash and model uncertainty,'' \emph{IEEE Transactions on Industrial Electronics}, vol.~70, no.~9, pp. 9379--9389, 2022.

\bibitem{zhao2023neural}
X.~Zhao, Z.~Liu, and Q.~Zhu, ``Neural network-based adaptive controller design for robotic manipulator subject to varying loads and unknown dead-zone,'' \emph{Neurocomputing}, vol. 546, p. 126293, 2023.

\bibitem{li2021valve}
L.~Li, Z.~Lin, Y.~Jiang, C.~Yu, and J.~Yao, ``Valve deadzone/backlash compensation for lifting motion control of hydraulic manipulators,'' \emph{Machines}, vol.~9, no.~3, p.~57, 2021.

\bibitem{lampinen2019model}
S.~Lampinen, J.~Koivum{\"a}ki, J.~Mattila, and J.~Niemi, ``Model-based control of a pressure-compensated directional valve with significant dead-zone,'' in \emph{Fluid Power Systems Technology}, vol. 59339.\hskip 1em plus 0.5em minus 0.4em\relax American Society of Mechanical Engineers, 2019, p. V001T01A028.

\bibitem{kang2020almost}
S.~Kang, R.~Nagamune, and H.~Yan, ``Almost disturbance decoupling force control for the electro-hydraulic load simulator with mechanical backlash,'' \emph{Mechanical Systems and Signal Processing}, vol. 135, p. 106400, 2020.

\bibitem{mustalahti2018nonlinear}
P.~Mustalahti and J.~Mattila, ``Nonlinear model-based controller design for a hydraulic rack and pinion gear actuator,'' in \emph{Fluid Power Systems Technology}, vol. 51968.\hskip 1em plus 0.5em minus 0.4em\relax American Society of Mechanical Engineers, 2018, p. V001T01A020.

\bibitem{mustalahti2019nonlinear}
P.Mustalahti and J.~Mattila, ``Nonlinear model-based control design for a hydraulically actuated spherical wrist,'' in \emph{Fluid Power Systems Technology}, vol. 59339.\hskip 1em plus 0.5em minus 0.4em\relax American Society of Mechanical Engineers, 2019, p. V001T01A027.

\bibitem{linan2012controller}
M.~d. C.~R. Li{\~n}{\'a}n and W.~P. Heath, ``Controller structure for plants with combined saturation and deadzone/backlash,'' in \emph{2012 IEEE International Conference on Control Applications}.\hskip 1em plus 0.5em minus 0.4em\relax IEEE, 2012, pp. 1394--1399.

\bibitem{yecsildirek1995feedback}
A.~Ye{\c{s}}ildirek and F.~L. Lewis, ``Feedback linearization using neural networks,'' \emph{Automatica}, vol.~31, no.~11, pp. 1659--1664, 1995.

\bibitem{xu2024dynamic}
N.~Xu, X.~Liu, Y.~Li, G.~Zong, X.~Zhao, and H.~Wang, ``Dynamic event-triggered control for a class of uncertain strict-feedback systems via an improved adaptive neural networks backstepping approach,'' \emph{IEEE Transactions on Automation Science and Engineering}, 2024.

\bibitem{liu2018adaptive}
Y.-J. Liu, Q.~Zeng, L.~Liu, and S.~Tong, ``An adaptive neural network controller for active suspension systems with hydraulic actuator,'' \emph{IEEE Transactions on Systems, Man, and Cybernetics: Systems}, vol.~50, no.~12, pp. 5351--5360, 2018.

\bibitem{sunderhauf2018limits}
N.~S{\"u}nderhauf, O.~Brock, W.~Scheirer, R.~Hadsell, D.~Fox, J.~Leitner, B.~Upcroft, P.~Abbeel, W.~Burgard, M.~Milford \emph{et~al.}, ``The limits and potentials of deep learning for robotics,'' \emph{The International journal of robotics research}, vol.~37, no. 4-5, pp. 405--420, 2018.

\bibitem{lampinen2021force}
S.~Lampinen, J.~Koivum{\"a}ki, W.-H. Zhu, and J.~Mattila, ``Force-sensor-less bilateral teleoperation control of dissimilar master--slave system with arbitrary scaling,'' \emph{IEEE Transactions on Control Systems Technology}, vol.~30, no.~3, pp. 1037--1051, 2021.

\bibitem{10210088}
M.~Hejrati and J.~Mattila, ``Nonlinear subsystem-based adaptive impedance control of physical human-robot-environment interaction in contact-rich tasks,'' \emph{IEEE Robotics and Automation Letters}, vol.~8, no.~10, pp. 6083--6090, 2023.

\bibitem{koivumaki2015stability}
J.~Koivum{\"a}ki and J.~Mattila, ``Stability-guaranteed force-sensorless contact force/motion control of heavy-duty hydraulic manipulators,'' \emph{IEEE Transactions on Robotics}, vol.~31, no.~4, pp. 918--935, 2015.

\bibitem{hejrati2022decentralized}
M.~Hejrati and J.~Mattila, ``Decentralized nonlinear control of redundant upper limb exoskeleton with natural adaptation law,'' in \emph{2022 IEEE-RAS 21st International Conference on Humanoid Robots (Humanoids)}.\hskip 1em plus 0.5em minus 0.4em\relax IEEE, 2022, pp. 269--276.

\bibitem{chen2010robust}
M.~Chen, S.~S. Ge, and B.~V.~E. How, ``Robust adaptive neural network control for a class of uncertain mimo nonlinear systems with input nonlinearities,'' \emph{IEEE Transactions on Neural Networks}, vol.~21, no.~5, pp. 796--812, 2010.

\bibitem{lee2018natural}
T.~Lee, J.~Kwon, and F.~C. Park, ``A natural adaptive control law for robot manipulators,'' in \emph{2018 IEEE/RSJ International Conference on Intelligent Robots and Systems (IROS)}.\hskip 1em plus 0.5em minus 0.4em\relax IEEE, 2018, pp. 1--9.

\bibitem{harrison1995generalized}
A.~Harrison and D.~Stoten, ``Generalized finite difference methods for optimal estimation of derivatives in real-time control problems,'' \emph{Proceedings of the institution of mechanical engineers, part I: journal of systems and control engineering}, vol. 209, no.~2, pp. 67--78, 1995.

\bibitem{reza2010theory}
R.~N. Jazar, \emph{Theory of Applied Robotics: Kinematics, Dynamics, and Control}.\hskip 1em plus 0.5em minus 0.4em\relax Springer., 2010.

\end{thebibliography}

\begin{IEEEbiography}[{\includegraphics[width=1in,height=1.25in,clip,keepaspectratio]{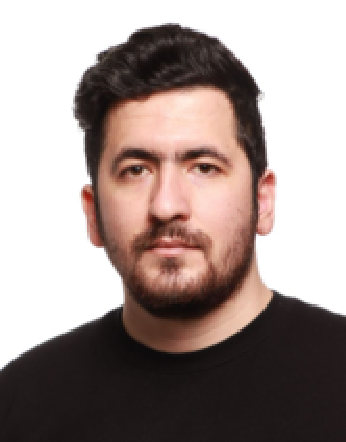}}]{Mahdi Hejrati} received his M.Sc. degree in 2021 from Sharif University of Technology (SUT), Tehran, Iran. He is currently a PhD student at the unit of Automation Technology and Mechanical Engineering, Tampere University, Tampere, Finland. His research interests include nonlinear model-based control, physical human-robot interaction, and human-inspired control.
\end{IEEEbiography}

\begin{IEEEbiography}[{\includegraphics[width=1in,height=1.25in,clip,keepaspectratio]{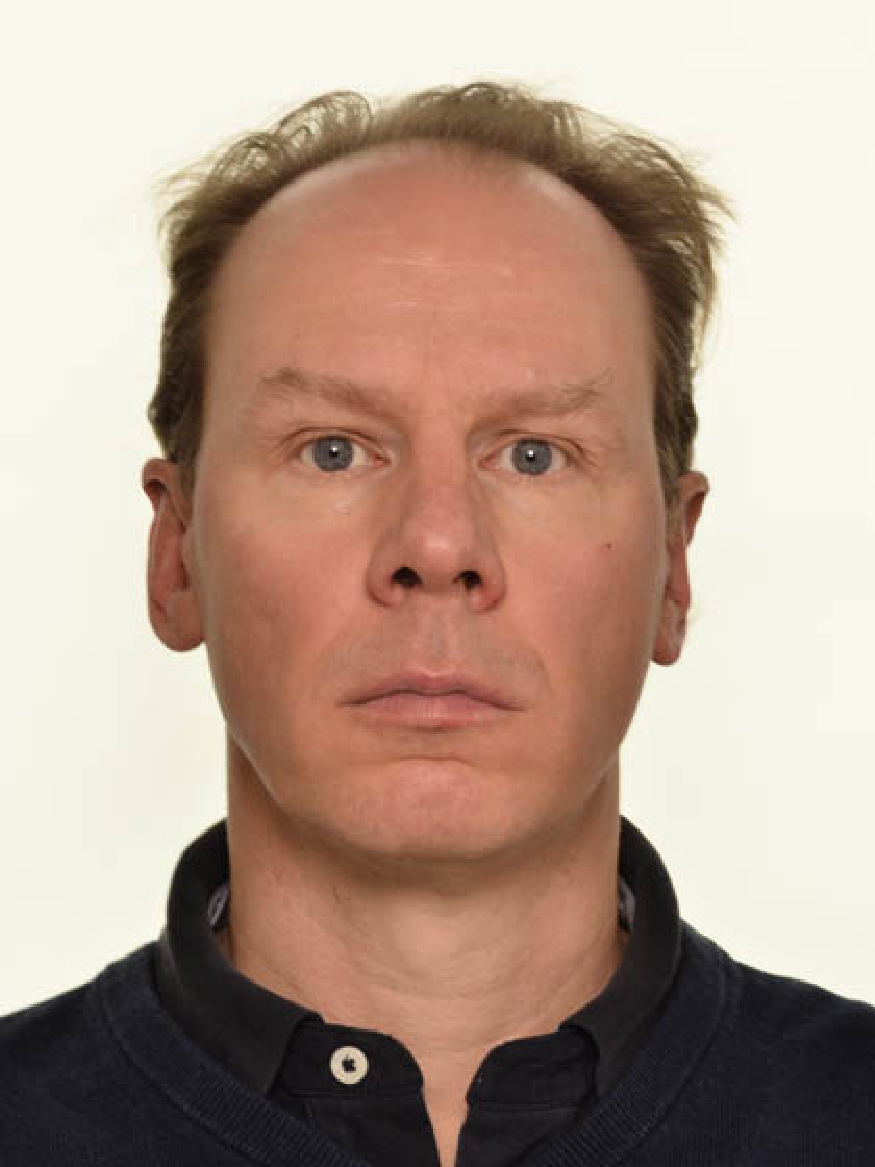}}]{Jouni Mattila}
Dr. Tech. received his M.Sc. (Eng.) in 1995 and Dr. Tech. in 2000, both from Tampere University of Technology (TUT), Tampere, Finland. He is currently a Professor of machine automation with the unit of Automation Technology and Mechanical Engineering, Tampere University. His research interests include machine automation, nonlinear model-based control of robotic manipulators and energy-efficient control of heavy-duty mobile manipulators.
\end{IEEEbiography}

\end{document}